\documentclass[preprint,12pt]{elsarticle}
\usepackage{graphicx}
\usepackage{hyperref, xcolor}
\usepackage{todonotes} 
\usepackage{booktabs}  
\usepackage{caption}   
\usepackage{subcaption} 
\usepackage{geometry}  
\usepackage{multirow}  
\usepackage{makecell} 
\usepackage{longtable}  
\usepackage{array, ragged2e}      
\usepackage{lscape}     
\usepackage{tabularx}   
\usepackage{amsmath, amssymb, amsthm} 
\usepackage{float}
\usepackage{subcaption}
\usepackage{pgfplots}
\usepgfplotslibrary{statistics}
\usetikzlibrary{decorations.pathreplacing}
\usepgfplotslibrary{groupplots}
\pgfplotsset{compat=1.18}
\usepackage{verbatim}
\usepackage{afterpage}
\graphicspath{ {./figures/} }

\usepackage{framed}
\usepackage{mdframed}

\theoremstyle{definition}
\newtheorem{definition}{Definition}
\newtheorem{remark}{Remark}
\newtheorem{metric}{Metric}

\newcolumntype{L}[1]{>{\raggedright\let\newline\\\arraybackslash\hspace{0pt}}m{#1}}
\newcolumntype{P}[1]{>{\raggedright\arraybackslash}p{#1}}
\newcolumntype{Y}{>{\raggedright\arraybackslash}X}

\begin{document}

\begin{frontmatter}


\cortext[cor1]{Corresponding author}

\title{Evaluation metrics for temporal preservation in synthetic longitudinal patient data}

\author[label1,label2]{Katariina Perkonoja\corref{cor1}} 
\author[label2]{Parisa Movahedi}
\author[label2]{Antti Airola}
\author[label1,label3]{Kari Auranen}
\author[label1]{Joni Virta}

\affiliation[label1]{organization={Department of Mathematics and Statistics, University of Turku},
            city={Turku},
            country={Finland}}

\affiliation[label2]{organization={Department of Computing, University of Turku},
            city={Turku},
            country={Finland}}

\affiliation[label3]{organization={Department of Clinical Medicine, University of Turku},
            city={Turku},
            country={Finland}}

\begin{abstract}
This study introduces a set of metrics for evaluating temporal preservation in synthetic longitudinal patient data, defined as artificially generated data that mimic real patients’ repeated measurements over time. The proposed metrics assess how synthetic data reproduces key temporal characteristics, categorized into marginal, covariance, individual-level and measurement structures. We show that strong marginal-level resemblance may conceal distortions in covariance and disruptions in individual-level trajectories. Temporal preservation is influenced by factors such as original data quality, measurement frequency, and preprocessing strategies, including binning, variable encoding and precision. Variables with sparse or highly irregular measurement times provide limited information for learning temporal dependencies, resulting in reduced resemblance between the synthetic and original data. No single metric adequately captures temporal preservation; instead, a multidimensional evaluation across all characteristics provides a more comprehensive assessment of synthetic data quality. Overall, the proposed metrics clarify how and why temporal structures are preserved or degraded, enabling more reliable evaluation and improvement of generative models and supporting the creation of temporally realistic synthetic longitudinal patient data.
\end{abstract}







\begin{highlights}
\item Novel metrics are proposed to assess temporal preservation in synthetic longitudinal data
\item The metrics cover mean, covariance, subject-level, and measurement structures
\item Multi-metric evaluation gives clearer insight into synthetic data quality
\item Assessing only mean-level resemblance can mask loss of temporal dependencies
\item Data quality, balance, and preprocessing strongly affect temporal preservation
\end{highlights}

\begin{keyword}
Synthetic data generation \sep Longitudinal patient data \sep Temporal preservation \sep Resemblance evaluation \sep Fidelity metrics



\end{keyword}

\end{frontmatter}



\section{Introduction} \label{sec:intro}

Electronic health records (EHRs) serve as the primary repository for contemporary patient data, storing treatment-related information across various health information systems. EHRs include diverse data types such as clinical notes, laboratory results, vital signs, imaging, and signal data. Among the various types of EHR-derived data, longitudinal patient data (LPD), i.e., structured tabular data that combines both static variables (e.g., patient demographics) and time-varying variables (i.e., repeated within-patient measurements), holds a great potential by providing a time-sequenced view of patient health.

While primarily used for patient care, EHR-derived data are increasinly leveraged by hospitals and healthcare providers for \emph{secondary purposes}, such as research, development, and innovation, as well as education. This broader use of patient data has raised concerns about privacy and confidentiality, prompting stringent regulations. Consequently, there is growing interest in facilitating streamlined access to patient data. Synthetic data generation (SDG) has emerged as a strategy to enhance data accessibility and sharing while minimizing privacy risks.

The goal of SDG is to produce synthetic data that closely resembles the original identifiable information, enabling similar use cases without compromising privacy. The increasing attention to synthetic data is evident in the growing variety of SDG methods documented in recent literature reviews \cite{Perkonoja2025MethodsReview,Ghosheh2024ARecords,Hernandez2022,Lautrup2024SystematicData,Murtaza2023SyntheticDomain, Ibrahim2025GenerativeChallenges}. Moreover, the significance of synthetic data is increasingly recognized at the policy level: for example, within the emerging European Health Data Space framework, synthetic data is envisioned as a key enabler for secure secondary use of health data \cite{TowardsEuropeanHealthDataSpaceTEHDAS2025DraftData}. Yet, evaluating SDG methods in terms of both data quality and privacy protection remains a nontrivial challenge.

A common practice is to categorize the evaluation of synthetic data into three primary domains:  \textit{resemblance} (or fidelity) between synthetic and original data, \textit{utility} of synthetic data, and \textit{privacy} aspects \cite{Murtaza2023SyntheticDomain,Hernandez2022,Ghosheh2024ARecords,Perkonoja2025MethodsReview}. Additionally, \textit{fairness} or \textit{diversity}, referring to the extent to which subgroups in the original data are proportionally represented in the synthetic data, has been proposed as a fourth domain \cite{Alaa2022HowModels,Bhanot2021a,Pereira2024AssessmentData}. 

In general, resemblance refers to the alignment of variable distributions in synthetic data with those in the original dataset, while utility refers to the ability to produce comparable inferential or predictive results to those obtained from the original data. These concepts exhibit partial overlap, as a good resemblance should imply high utility. Nevertheless, synthetic data may demonstrate high utility even in cases of moderate resemblance, provided that essential properties on which the utility measure relies are retained in synthetic data. Privacy considerations, while an important aspect of synthetic data evaluation, are outside the scope of this work and are therefore not addressed further; interested readers are referred to \cite{Osorio-Marulanda2024PrivacyReview}.

In this work, we develop resemblance evaluation for synthetic longitudinal patient data, with particular emphasis on unbalanced LPD. Here, unbalanced data refers to longitudinal datasets in which the number or timing of measurements vary across patients \cite{Diggle2002MissingData}. Unbalanced data is common in secondary-use settings as individualized treatment schedules result in irregular measurement times. Evaluating resemblance in this context is particularly challenging, owing not only to varying measurement times across patients but also to incomplete data from missed visits, dropouts (subjects who do not continue the treatment), or loss to follow-up (subjects researchers can no longer reach). 

Our objective is to propose metrics for evaluating \textit{temporal preservation}, that is, the extent to which synthetic data reproduces the evolution of variables over time and the distribution of measurements (in both timing and frequency). These metrics capture multiple aspects of temporal behavior, including preservation of marginal and covariance structures, individual trajectories, and measurement patterns. Most of the proposed metrics are calculated as functions of time, while a few provide aggregate scores that summarize similarity across all time points. Despite its importance for analytical validity, temporal preservation is frequently overlooked: many SDG studies either ignore it or aggregate data over time, discarding the very dependencies that define LPD \cite{Perkonoja2025MethodsReview}. Existing evaluation frameworks also rarely include metrics that assess temporal preservation \cite{Hernadez2023SyntheticDimensions,Dankar2022AGenerators,Yan2022AModels,Vallevik2024CanHealthcare,Belgodere2024AuditingTrade-offs}. 

Our focus on temporal preservation highlights an overlooked dimension of resemblance evaluation in synthetic LPD. In this context, we make four main contributions:
\begin{itemize}
    \item \textbf{A survey of existing metrics}. We survey prior temporal evaluation metrics used in synthetic LPD research, identifying key limitations such as assumptions of balanced data, reliance on specific preprocessing or models, over-aggregation into single scores, and limited interpretability in practice.
     \item \textbf{A novel set of metrics for assessing temporal preservation}. Motivated by the limitations identified in our survey, we introduce robust and adaptable metrics suited for real-world LPD that capture diverse temporal patterns using non-parametric estimators such as kernel smoothing, enabling flexible analysis and intuitive visualization of discrepancies between real and synthetic data.
    \item \textbf{Open-source implementation}. We provide easy-to-use implementations of the proposed metrics in \textit{R}, released under an open-source license, to support transparency, reproducibility and practical adoption.
    \item \textbf{Empirical illustration on real-world LPD}. We demonstrate our proposed evaluation metrics using two synthetic datasets derived from MIMIC-III \cite{Johnson2016MIMIC-III1.4,Johnson2016MIMIC-IIIDatabase}, generated via HALO \cite{Theodorou2023SynthesizeModel} and Health Gym GAN \cite{Kuo2022TheAlgorithms}. These examples highlight practical insights and help guide the interpretation of the metrics and SDG method behavior.
\end{itemize}

This work is structured as follows. Section~\ref{sec:prelim} introduces the preliminaries, including the notation used throughout the work and the formalization of the concept of \textit{univariate temporal preservation}, which provides the basis for many of the proposed metrics. Section~\ref{sec:survey} reviews existing approaches to assessing temporal preservation in synthetic LPD. Section~\ref{sec:evalm} details the proposed evaluation metrics and Section~\ref{sec:illustr} illustrates the application of these metrics in practice. Section~\ref{sec:discuss} concludes with a discussion of findings and practical implications.

\section{Preliminaries} \label{sec:prelim}

This section introduces the notation and setup used throughout the paper, outlines the kernel smoothing techniques employed in the construction of the evaluation metrics, and defines the concept of univariate temporal preservation.

\subsection{Notations}\label{sec:notations}

Let $ i = 1, \dots, n $, index the subjects (individuals) and $j = 1, \ldots, p$, index variables that have been measured over time. For subject $i$ and variable $j$, let 
$$
\textbf{t}_{ij} = (t_{ij1}, \ldots, t_{ijn_{ij}})', \quad n_{ij} \geq 2,
$$
denote the vector of ordered measurement times. When the measurement times differ between subjects or variables, a dataset is referred to as \textit{unbalanced}, and when they are identical across all subjects and variables, the dataset is \textit{balanced}. This notion of balance refers to the design of the measurement schedule and should be distinguished both from missing data, which occurs when a scheduled measurement did not result to observed data (e.g., a patient missed the visit, the measurement device failed, or the result was lost), and from other time-indexed variables that evolve naturally over time, such as age \cite{Diggle2002MissingData}. In the context of secondary data use, however, it is often difficult to distinguish true missingness from structural irregularities in measurement schedules. Consequently, in this work we do not attempt to differentiate missingness from unbalancedness.

The corresponding observations for subject $i$ and variable $j$ are collected into the vector $\textbf{x}_{ij} = (x_{ij1}, \ldots, x_{ij n_{ij}})'$
where $x_{ijk}$ denotes the value of variable $j$ for subject $i$ at time $t_{ijk}$. Depending on the type of variable, the entries $x_{ijk}$ may be continuous or discrete. In the latter case, they take values in a finite space $\mathcal{C}_j = \{c_{j1}, \dots, c_{jL}\}$, which may differ across variables. For simplicity, we refer to these sets collectively as $\mathcal{C}$ when the distinction is not essential.

For each variable $j$, subject-specific observations are combined with their measurement times $\mathbf{t}_{ij}$ into a data matrix $\mathbf{X}_j$ of size $\sum_i n_{ij} \times 2$. This type of representation is commonly referred to as the \textit{long format}. An alternative is the \textit{wide format}, in which the rows consist of transposed subject-specific vectors.

For each variable $j$, we define the set of unique measurement times across all subjects as
$$
T_j = \bigcup_{i=1}^n \{\, t_{ijk} : k = 1,\dots,n_{ij} \,\}.
$$
$T_j$ thus collects all distinct time points at which variable $j$ was measured. We also define a \textit{measurement indicator matrix}
\begin{align}\label{def:measurement_indicator_matrix}
    \mathbf{I}_j = \bigl(I_{ijt}\bigr)_{i=1,\dots,n;\; t \in T_j},    
\end{align}
with entries
$I_{ijt} = \mathbf{1}\!\left\{ \exists\, k \;:\; t_{ijk} = t \right\}$,
where $\mathbf{1}\{\cdot\}$ denotes the indicator function. In other words, $I_{ijt} = 1$ if subject $i$ has at least one observed measurement of variable $j$ at time $t$, and $I_{ijt} = 0$ otherwise.

For each subject $i$, the dropout point is defined as the earliest time $t$ after which no further measurements were made:
$$
d_{ij} = \min \left\{\, t \in T_j \;\middle|\; I_{ijs} = 0 \;\; \forall s \in T_j, s > t \right\}.
$$
Throughout this work, a tilde  (e.g, $\tilde{x}_{ijk}$) denotes the synthetic counterpart of the corresponding original notion.

\subsection{Kernel smoothing}

Kernel smoothing offers a flexible tool for summarizing unbalanced longitudinal data \cite{Wand1995KernelSmoothing}. By using non-parametric smoothing, arbitrary temporal patterns can be captured and values can be interpolated at unobserved time points. A key component of this approach is the use of normalized kernel weights, which assign relevance to observed time points when estimating values at a target time $t$.

Since the time distributions of the original and synthetic data may differ, i.e., certain time points may only appear in one of them, we evaluate all metrics for a given variable $j$ on a common \emph{regular} time grid. 
$$
T_j^\ast = \bigl\{\, t_j^{\min} + m\,\Delta_j \;:\; 
m = 0,1,\dots, \big\lfloor (t_j^{\max}-t_j^{\min})/\Delta_j \big\rfloor \,\bigr\},
$$
where $t_j^{\min} = \min\bigl(T_j \cup \tilde T_j\bigr)$, $t_j^{\max} = \max\bigl(T_j \cup \tilde T_j\bigr)$, and $\Delta_j > 0$ is a fixed grid step (the unit of time for variable $j$). This grid choice defines where kernel-based estimates are evaluated, and alternative choices can be used depending on the aspects of temporal behavior one wishes to emphasize, for example by restricting evaluation to the original time range.

\begin{definition}[Normalized kernel weights] \label{def:kernel}
Let $K_h(t, t')$ denote a kernel function, giving higher weights to (measurement) time points $t'$ that are closer to $t \in T_j^\ast$.
For example, the Gaussian kernel is defined as
$$
K_h(t, t') = \frac{1}{h} \exp\left( -\frac{(t - t')^2}{2h^2} \right),
$$
where $h > 0$ is the bandwidth controlling the degree of smoothing. The normalized kernel weights for the $j$th variable at time $t \in T_j^\ast$ are
$$
\bar{\omega}_{ijk}(t) = \frac{K_h(t, t_{ijk})}{\sum_{i' = 1}^n \sum_{k' = 1}^{n_{i'j}} K_h(t, t_{i'jk'})}, \quad i = 1, \ldots, n, \quad k = 1, \ldots, n_{ij}.
$$
\end{definition}

The weights $\bar{\omega}_{ijk}(t)$ in Definition~\ref{def:kernel} represent how much influence each observed time point $t_{ijk}$ has when estimating or smoothing a value at the target time $t$. The bandwidth $h$ determines how quickly the weights decrease, effectively setting a window of nearby points that contribute significantly to the estimate. Normalizing ensures that all weights sum to one, letting us later use them to construct meaningful weighted averages at each time $t$.

Definition~\ref{def:cdf} defines a weighted empirical cumulative distribution function used by some of the proposed metrics.
\begin{definition}[Weighted empirical cumulative distribution]\label{def:cdf}
The weighted empirical cumulative distribution function of the $j$th variable at time $t \in T_j^\ast$ is defined as
$$
F_{jz}^{(K)}(t) = \sum_{i = 1}^n \sum_{k = 1}^{n_{ij}} \bar{\omega}_{ijk}(t) \, \mathbf{1}(x_{ijk} \leq z),
$$
where $\bar{\omega}_{ijk}(t)$ are the normalized kernel weights.    
\end{definition}

\begin{remark}[Choice of bandwdith $h$]
Kernel methods are often sensitive to the choice of bandwidth $h$, but in the present context this sensitivity is not a drawback. Since the aim is to assess whether real and synthetic data behave similarly, any discrepancy between the two, regardless of the value of $h$, can be interpreted as evidence of a difference. Thus, the bandwidth serves more as a lens that highlights differences across various smoothing scales than as a tuning parameter to be optimized. Exploring several values of $h$ may therefore provide a more nuanced comparison. Larger values of $h$ emphasize population-level trends, whereas smaller values highlight local fluctuations and subject-specific variation. In principle, $h$ could be chosen systematically, using a selection rule from the classical kernel smoothing literature, see \cite{Wand1995KernelSmoothing}. However, for simplicity, in this work $h$ was selected manually.

\end{remark}

\subsection{Univariate temporal preservation}

A central objective in this work is to evaluate how well synthetic data reproduces the temporal patterns observed in the original longitudinal data. Since variables may differ in their measurement schedules, frequencies, and temporal behaviors, it is natural to consider each variable separately. We refer to this aspect as \emph{univariate temporal preservation}. 

The key idea is to summarize the temporal evolution of each variable $j$ using a suitable metric $M_j(\cdot)$. By applying the same metric to both the original and synthetic datasets, we can assess whether the synthetic data preserves the temporal structure present in the original data. 
\begin{definition}[Univariate temporal preservation] \label{def:utp}
For each variable $j$, consider the value--time pairs
$$
\{(x_{ijk}, t_{ijk}) : i = 1,\dots,n;\; k = 1,\dots,n_{ij}\},
$$
and let $M(\cdot)$ denote a metric that summarizes the evolution of variable $j$ over time.  
Applying $M_j$ to the original and synthetic data yields
$$
M_j = M\big((x_{ijk}, t_{ijk})\big),  \qquad  \tilde{M}_j = M\big((\tilde{x}_{ijk}, \tilde{t}_{ijk})\big).
$$
We say that the synthetic data exhibit \textbf{univariate temporal preservation} for variable $j$ if $\tilde{M}_j \approx M_j$.
\end{definition}

\begin{remark}[Goodness of approximation]
The goodness of approximation in Definition~\ref{def:utp} is inherently data- and task-dependent. What constitutes sufficient preservation depends on the specific variable, the intended analyses, and the scale of variation in the data. In practice, similarity between $M_j$ and $\tilde{M}_j$ can be judged in different ways. One option is to quantify their difference numerically, e.g., by computing deviations relative to a tolerance threshold or by using statistics such as correlation coefficients. Another option is to evaluate similarity qualitatively, for example through different visualizations. In practice, these preservation metrics should be interpreted together with other resemblance and utility evaluations: a small discrepancy between $M_j$ and $\tilde{M}_j$ may not impair the analytical validity of the synthetic data, whereas a similar discrepancy in a different context could be critical. Conversely, a large difference between $M_j$ and $\tilde{M}_j$ is typically indicative of reduced utility. \end{remark}

\section{Survey of existing metrics} \label{sec:survey}

To identify gaps in existing approaches and to avoid duplicating prior efforts, we reviewed different metrics used to evaluate temporal preservation in the SDG literature. The metrics were identified from a recent review of methods for generating and evaluating synthetic longitudinal data \cite{Perkonoja2025MethodsReview}. We classified the identified approaches in seven distinct categories: marginal summary statistics, correlation statistics, transition probabilities, conditional dependency metrics, trend-residual decomposition, latent space representations and measurement time reconstruction error.

\textit{Marginal summary statistics}, applied in \cite{Zhang2021, Li2023GeneratingApplications,Haleem2023Deep-Learning-DrivenSynthesis,Achterberg2024OnRecords,Wang2023EnhancingWCGAN-GP, Theodorou2023SynthesizeModel,Nikolentzos2023SyntheticAutoencoders}, capture univariate temporal preservation of means, variances, proportions, and frequencies. While these statistics are useful for comparing original and synthetic data, prior implementations often assumed balanced datasets, imputed missing values, normalized variables, or aggregated metrics into single scores, limiting interpretability and applicability to unbalanced data. To address these limitations, we apply kernel smoothing (Sections~\ref{sec:mean_struc}, \ref{sec:cov_stuct}) to pool information across uneven measurement schedules. Marginal statistics have also been used to compare distributions of measurement times, but summarizing them with a single value obscures important distributional details. To overcome this, we introduce metrics that offer more intuitive and informative comparisons (Section~\ref{sec:metrics_struc}).

\textit{Correlation statistics} capture dependencies either within a subject’s consecutive measurements (autocorrelation) or between subjects at the same time point (cross-correlation). Prior works have used autocorrelation functions and Pearson pairwise correlation (PPC) \cite{Li2023GeneratingApplications,Isasa2024ComparativeSynthesis}. When such measures are reported as a single summary measure, however, they provide only a limited view of the dependency structure. Moreover, both approaches typically assume balanced data, and PPC further assumes linearity. To address these limitations, we provide a metric that accommodates unbalanced data, captures non-linear dependencies and provides intuitive visualizations that enhance understanding of the underlying correlation structure (Section~\ref{sec:cov_stuct}).

\textit{Transition probabilities}, used in \cite{Mosquera2023AData, Theodorou2023SynthesizeModel,Sun2023CollaborativeInference,Theodorou2024,Pang2024CEHR-GPT:Timelines, Nikolentzos2023SyntheticAutoencoders}, describe the sequential dependence between discrete states by quantifying the conditional likelihood of moving from one state to another. While intuitive, reducing the probabilities to secondary summaries loses important structure, and higher-order transitions are often sparse and difficult to interpret. Existing approaches also assume balanced data. We introduce a smoothed first-order transition probability metric that accommodates unbalanced data and can be visualized intuitively (Section~\ref{sec:cov_stuct}).

\textit{Conditional dependency metrics}, applied in \cite{Wang2022PromptEHR:Learning, Hashemi2023Time-seriesNetwork}, extend the idea of transition probabilities to a continuous setting, evaluating sequential dependence by conditioning each observation on its prior history. While these metrics are valid and informative, they require fitting a pre-chosen parametric model on the data, limiting their usefulness. To keep the approach fully non-parametric and assumption-free, we do not consider such metrics in this work.

\textit{Trend-residual decomposition}, utilized in \cite{Kuo2022TheAlgorithms,Kuo2023GeneratingHIV}, fits a subject-specific model where the fitted curve captures the overall trend and the residuals represent short-term deviations. These metrics suffer the same model selection issue as the conditional dependency metrics: they may be too simplistic and may not handle unbalanced data. Interpretation also poses challenges: aggregating results into a single measure can obscure important heterogeneity, such as subgroup effects. In this work, we do not include model-based resemblance metrics, although they remain valid options when carefully applied in specific use cases.

\textit{Latent space representations}, applied in \cite{Zhang2021,Achterberg2024OnRecords}, embed repeated measurements into a lower-dimensional space. These approaches can capture complex dependencies that are difficult to summarize with simple statistics of the original data. However, their interpretability is limited. Latent factors are often abstract and may not clearly reflect patterns in the original measurement space. In addition, some commonly used methods, such as principal component analysis, assume linearity and require balanced data. More flexible methods (e.g., non-linear embeddings) can relax these assumptions, but at the cost of further reducing transparency and complicating validation. In this work, we do not pursue resemblance metrics based on latent representations, as their abstraction makes them difficult to interpret in the context of LPD.

\textit{Measurement time reconstruction error} proposed by \cite{Pang2024CEHR-GPT:Timelines} assesses how well chronological structure is preserved by comparing the original measurement times with their model reconstructions. Unlike marginal summary statistics derived from the measurement time distribution, this comparison specifically captures the model’s information loss when mapping continuous time into discrete classes, such as tokens in language models. Consequently, this metric is restricted to a particular SDG model and is not considered here, albeit relevant when generating synthetic data with language models.

Although we identified several valid approaches for measuring temporal preservation, our key concern in the literature is, besides the above limitations, that many studies rely on a single metric. This can prevent obtaining a clear understanding of whether the temporal structure is genuinely preserved. As we show in Section~\ref{sec:illustr}, a variable may retain, for example, marginal data patterns yet completely losing covariance structures. Therefore, when assessing temporal preservation, we emphasize the importance of a comprehensive evaluation, for which we present a range of metrics in the following section.

\section{Metrics for assessing temporal preservation} \label{sec:evalm}

Our approach to evaluating temporal preservation in synthetic longitudinal patient data is grounded in explanatory data analysis (EDA) \cite{Diggle2002ExploringData}. In general, EDA aims to reveal important patterns in observed data that can then guide the development of more formal analysis models. Similar methods can be used to assess how closely synthetic data resembles the original data. In our approach, we organize the key characteristics of longitudinal data into four principal categories. The first three categories are concerned with preserving temporal structure, while the fourth describes the distribution of measurement times:

\begin{itemize}
    \item[I.] \textit{Marginal structure} describes population-level behavior, representing how the overall mean and quantiles of a continuous variable or the class proportions of a discrete variable change over time. This includes broad trends (linear or nonlinear) and potential periodicity (recurring patterns).
    \item[II.] \textit{Covariance structure} reflects how repeated measures are related over time, capturing both the between-subject variability and temporal within-subject correlation patterns. For continuous variables, covariance structure includes variance, autocorrelation, and the stability of individuals’ relative rankings (rank-order stability). For discrete variables, it covers transition probabilities between states, their consistency across time (time homogeneity), and persistence within the same state (class stability).
\item[III.] \textit{Individual structure} focuses on subject-specific trajectories, which complement population-level summaries.
\item[IV.] \textit{Measurement structure} concerns the timing and frequency of measurements, including missing data, dropouts or loss to follow-up.
\end{itemize}

\begin{figure}
    \centering
    \includegraphics[width=0.9\textwidth]{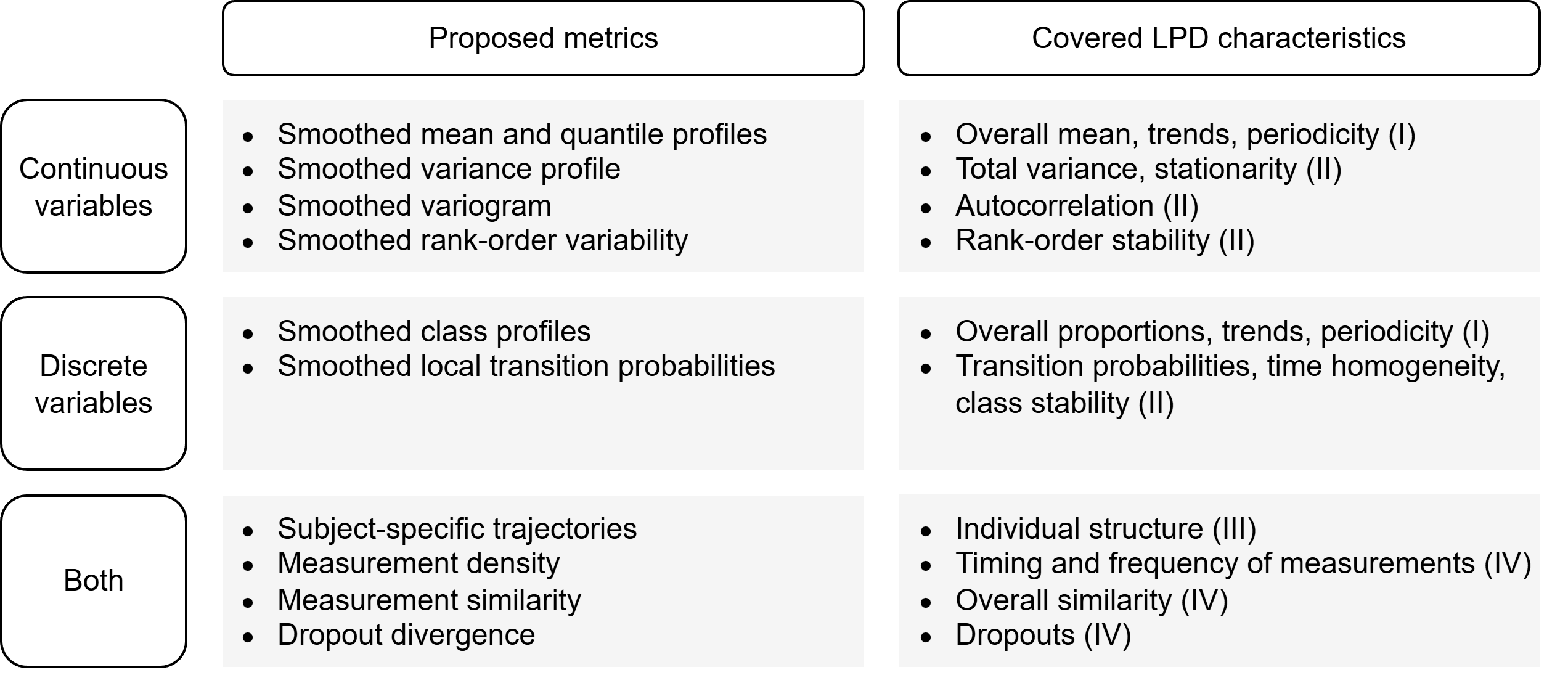}
    \caption{Evaluation metrics proposed in this work for assessing univariate temporal preservation in synthetic LPD. Some metrics are only applicable to either continuous or discrete variables.}
    \label{fig:metrics}
\end{figure}

Given the inherent complexity and variability of LPD, where subject-specific trajectories may exhibit highly diverse and irregular forms, visual representations play a central role in evaluation. Unlike scalar summary measures, visualizations enable a more intuitive assessment of shapes and temporal patterns, facilitating easy detection of trends, anomalies, and structures that might otherwise remain obscured. This justifies the emphasis on graphical methods in our evaluation framework, where visual tools are prioritized over numerical metrics to better capture the resemblance of synthetic and original data. Figure \ref{fig:metrics} presents the metrics proposed in this study, which are further discussed in detail in the following subsections and illustrated in Section \ref{sec:illustr}. All metrics presented in this work can also be computed separately within appropriate subsets (strata) of the data.

\subsection{Metrics to assess the marginal structure} \label{sec:mean_struc}

Different summary statistics offer practical tools to assess temporal preservation and resemblance between synthetic and original LPD. For each variable, we begin by examining its \textit{temporal profile}, defined as the evolution of cross-sectional distributions over time. For a numerical variable, this metric involves tracking the mean and quantiles, while for a categorical variable we consider class probabilities.

The \textit{kernel-smoothed mean profile} provides a basic summary by describing the average value of a variable at each time point. This smoothed mean curve not only captures the overall trend but also reveals underlying periodic patterns at the population level.

\begin{oframed}
\vspace{-0.4cm}
\begin{metric}[] \label{metric:mean_profile}
The kernel-smoothed mean profile of variable $j$ at time $t \in T_j^\ast$ is 
\begin{align*}
\mu_j^{(K)}(t) &= \sum_{i = 1}^n \sum_{k = 1}^{n_{ij}} \bar{\omega}_{ijk}(t)  x_{ijk}, \quad t \in T_j^\ast.
\end{align*}    
\end{metric}
\vspace{-0.4cm}
\end{oframed}

While the smoothed mean profile effectively captures the central tendency, it does not provide information about the variability or other distributional characteristics. \textit{Kernel-smoothed quantile profiles} overcome this by capturing the distribution’s shape across quantiles, providing a fuller view of the variable’s temporal evolution.

\begin{oframed}
\begin{metric}[]\label{metric:quant_prof}
The kernel-smoothed quantile profile of  variable $j$ at time $t \in T_j^\ast$ is
$$
Q_{jq}^{(K)}(t) := \left\{F_{jq}^{(K)}\right\}^{-1}(t), \quad t \in T_j^\ast,
$$
where $F_{jq}^{(K)}$ is the weighted empirical CDF and $q$ is the quantile level of interest. 
\end{metric}
\end{oframed}

Mean and quantile profiles can be visualized using line charts (see Section \ref{sec:illustr_moder}). These plots can be enhanced by overlaying observations outside the defined quantile range. Such visualizations are particularly useful since the mean profile and outer quantiles are sensitive to outliers, and plotting them can help explain deviations in the observed metrics and assess how well synthetic data capture outlier behavior. 

For discrete variables, the \textit{kernel-smoothed class profile} represents the proportion of each class $c_{j\ell} \in \mathcal{C}_j$, capturing how class distributions evolve over time. These profiles can be visualized, for example, with area charts (see Section~\ref{sec:illustr_categ}).

\begin{oframed}
\begin{metric}[]\label{metric:class}
The kernel-smoothed class profile at time $t \in T_j^\ast$ is the $L$-dimensional vector $\bigl(p_{j1}^{(K)}(t), \dots, p_{jL}^{(K)}(t)\bigr)$ where
$$
p_{j\ell}^{(K)}(t) = \sum_{i = 1}^n \sum_{k = 1}^{n_{ij}} 
    \bar{\omega}_{ijk}(t) \, \mathbf{1}(x_{ijk} = c_{j\ell}), 
    \quad \ell = 1,\dots, L.
$$
\end{metric}
\end{oframed}

\subsection{Metrics to assess the covariance structure} \label{sec:cov_stuct}

Assessing the marginal structure is a fundamental step in evaluating synthetic data because it provides valuable insight into whether the data captures even the simplest underlying patterns. However, this alone is not enough, especially when dealing with longitudinal data as these metrics do not directly account for variability or time-dependent relationships within a subject. Additionally, in longitudinal data analysis, covariance structure is often of great importance as it plays a crucial role in guiding the selection of the appropriate model.

A natural entry point for examining the covariance structure is the total variance of each variable. To this end, we use the \textit{kernel-smoothed variance profile}, which depicts how the total variance evolves over time.

\begin{oframed}
\begin{metric}[]\label{metric:variance}
The kernel-smoothed variance profile of variable $j$ at time $t \in T_j^\ast$ is 
$$
\sigma_{j}^{2(K)}(t) = \sum_{i=1}^n \sum_{k=1}^{n_{ij}} \bar{\omega}_{ijk}(t) \left(r_{ijk}^{(K)}(t)\right)^2,
$$
where the time-specific residuals are $r_{ijk}^{(K)}(t) = x_{ijk} - \mu_j^{(K)}(t)$.
\end{metric}
\end{oframed}

The variance profile reveals two important aspects: how much individual observations deviate from the variable's mean profile (Metric \ref{metric:mean_profile}) and the temporal shape of the variance, including stationarity (variance remains constant over time). Similarly to the mean and quantile profiles, the variance profile can be visualized using line charts (see Section \ref{sec:illustr_moder}). 

In the context of LPD, repeated measurements introduce dependencies that necessitate a more detailed analysis of temporal structures, requiring tools capable of capturing autocorrelation. One such tool is the \textit{variogram} (sometimes referred to as the \textit{semivariogram}) \cite{Diggle2002ExploringData,Diggle1998NonparametricData,Gringarten2001TeachersModeling}, which we also estimate using kernel smoothing. The variogram measures variability through weighted squared differences between pairs of kernel-smoothed residuals separated by a time lag of $u$ time units. \clearpage

\begin{oframed}
\begin{metric}[] \label{metric:vario}
The kernel-smoothed variogram is calculated as

$$\gamma_j^{(K)}(u) = \frac{1}{2}\sum_{i=1}^n\sum_{k=1}^{n_{ij}-1}\sum_{l =k+1}^{n_{ij}}
{(r_{ijk}^{(K)}(t_{ijk})-r_{ijl}^{(K)}(t_{ijl}))}^2 \bar w_j^{(K)}(u,t_{ijl}-t_{ijk}),~u>0,$$
where the weights are
$$\bar w_j^{(K)}(u,t_{ijl}-t_{ijk}) = \frac{K_h(u,t_{ijl}-t_{ijk})}
{ \sum_{i^\prime=1}^{n}\sum_{k\prime=1}^{n_{i^\prime j}-1}\sum_{l^\prime=k^\prime+1}^{n_{i^\prime j}}K_h(u,t_{i^\prime jl^\prime}- t_{i^\prime j k^\prime}).
}$$
\end{metric}
\end{oframed}

\begin{figure}[h]
\centering
\begin{tikzpicture}
  \pgfmathsetmacro{\a}{3}            
  \pgfmathsetmacro{\tauTwo}{0.2}     
  \pgfmathsetmacro{\sigmaTwo}{0.8}   
  \pgfmathsetmacro{\nuTwo}{0.5}      
  \pgfmathsetmacro{\sillWith}{\tauTwo+\sigmaTwo+\nuTwo}

  \definecolor{expcol}{HTML}{FF7F0E}
  \definecolor{gausscol}{HTML}{2CA02C}
  \definecolor{totalcol}{HTML}{1F77B4}

  \pgfmathdeclarefunction{gammaexp}{1}{\pgfmathparse{\tauTwo + \sigmaTwo*(1 - exp(-(#1/\a)))}}%
  \pgfmathdeclarefunction{gammagauss}{1}{\pgfmathparse{\tauTwo + \sigmaTwo*(1 - exp(-(#1/\a)^2))}}%

  \def\braceX{9.3}  
  \def\braceShift{16pt}
  \def\labelX{10.6}     
  \def\labelShift{1pt}
  \def\braceAmp{5pt}

  \begin{axis}[
    width=0.7\textwidth, height=5.9cm,
    xmin=0, xmax=10,
    ymin=0, ymax=0.25+\sillWith,
    xtick={0,...,10},
    xtick pos=lower,
    xlabel={Lag (u)},
    ytick pos=left,                   
    ylabel={Variability},
    axis line style={black}, tick style={black},
    clip=false,
    legend style={draw = none, at={(0.5,1.05)}, anchor=south, legend columns=3, /tikz/every even column/.style={column sep=6pt}},
    legend cell align=left,
    font = \footnotesize
  ]
    \addplot[totalcol, thick] coordinates {(0,\sillWith) (10,\sillWith)};
    \addlegendentry{Total Variance}
    \addplot[expcol,  thick, domain=0:10, samples=200] {gammaexp(x)};
    \addlegendentry{Variogram (Exponential)}
    \addplot[gausscol, thick, domain=0:10, samples=200] {gammagauss(x)};
    \addlegendentry{Variogram (Gaussian)}

    \draw[decorate,decoration={brace,amplitude=\braceAmp}, xshift=\braceShift]
      (axis cs:\braceX,\tauTwo) -- (axis cs:\braceX,0);
    \node[rotate=-90, anchor=center, xshift=\labelShift]
      at (axis cs:\labelX,{0.5*\tauTwo}) {$\tau^2$};

    \draw[decorate,decoration={brace,amplitude=\braceAmp}, xshift=\braceShift]
      (axis cs:\braceX,{\tauTwo+\sigmaTwo}) -- (axis cs:\braceX,\tauTwo);
    \node[rotate=-90, anchor=center, xshift=\labelShift]
      at (axis cs:\labelX,{\tauTwo+0.5*\sigmaTwo}) {$\sigma^2$};

    \draw[decorate,decoration={brace,amplitude=\braceAmp}, xshift=\braceShift]
      (axis cs:\braceX,\sillWith) -- (axis cs:\braceX,{\tauTwo+\sigmaTwo});
    \node[rotate=-90, anchor=center, xshift=\labelShift]
      at (axis cs:\labelX,{\tauTwo+\sigmaTwo+0.5*\nuTwo}) {$\nu^2$};

  \end{axis}
\end{tikzpicture}

\caption{Schematic (theoretical) variograms with exponential and Gaussian correlation under stationarity. The variance decomposes into measurement error ($\tau^{2}$), a serially correlated (autocorrelated) component ($\sigma^{2}$), and, if present, the variance of a random intercept representing between-subject variability ($\nu^{2}$).}
\label{fig:variogram}
\end{figure}
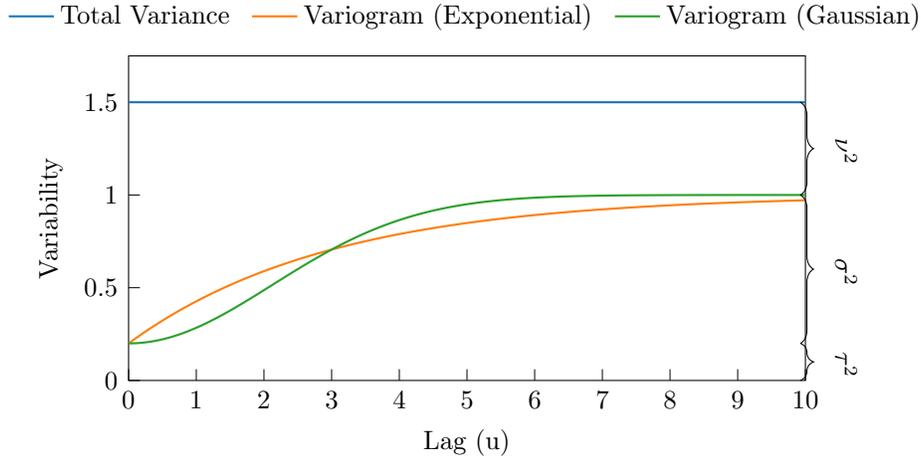

Assuming the total variance (Metric~\ref{metric:variance}) is stationary, the variogram can be used to separate different sources of variation (Figure~\ref{fig:variogram}), as it is directly related to the autocorrelation function \cite{Diggle2002ExploringData}. Under stationarity, the gap between the total variance and the variogram corresponds to \textit{between-subject variability}, the unexplained component at baseline ($t=0$) represents \textit{random measurement error}, also referred to as \textit{nugget}, and the remaining part reflects temporal within-subject variability (\textit{serial correlation}, i.e., autocorrelation).

The shape of the variogram conveys the strength of autocorrelation. A sharp increase near the origin indicates strong short-range dependence, whereas a more gradual rise suggests longer-range correlation. As the lag increases, the variogram typically approaches a plateau, known as the \textit{sill}, which corresponds to the total variance of the process when no between-subject variability is present. Under stationarity, a variogram that exceeds the total variance generally indicates the presence of a trend in the data.

For clarity, our illustrations in Section~\ref{sec:illustr} display both the kernel-smoothed variance profile and the variogram side by side as in Figure~\ref{fig:variogram}. When the variable is non-stationary, the precise analytical interpretation of the variogram is less relevant for our purpose as the primary goal is that the metrics resemble one another, even if the figure itself does not carry the same interpretative meaning.

In addition to the variance profile and the variogram, it is also useful to examine whether subjects preserve their relative ordering over time. This property, called \textit{rank-order stability}, reflects how consistently repeated measurements maintain their positions, with high stability linked to strong autocorrelation and low stability to volatile patterns. To measure this, we propose \textit{kernel-smoothed rank-order variability}.

\begin{oframed}
\begin{metric}[]\label{metric:rank_var}
Let $\mathcal{Q} = \{q_1,\dots,q_Q\}$ be a set of quantile levels, and let 
$Q_{jq}^{(K)}(t), \; q \in \mathcal{Q},$ denote the kernel-smoothed quantile 
profile of variable $j$ at time $t$ (see Metric~\ref{metric:quant_prof}). 
The kernel-smoothed rank-order variability for subject $i$ and variable $j$ is
$$
\mathcal{R}_{ij} = \frac{1}{Q} \cdot \frac{1}{n_{ij} - 1} 
\sum_{k=1}^{n_{ij}-1} 
\big( R_{ij(k+1)}^{(K)} - R_{ijk}^{(K)} \big),
$$
where $R_{ijk}^{(K)} \in \{1,\dots,Q+1\}$ is the quantile-bin index of
$x_{ijk}$ defined by
\[
R_{ijk}^{(K)}=\ell
\quad\text{if}\quad
Q_{jq_{\ell-1}}^{(K)}(t) < x_{ijk} \le
Q_{jq_\ell}^{(K)}(t),
\]
with the conventions
$Q_{jq_0}^{(K)}(t)=-\infty$ and
$Q_{jq_{Q+1}}^{(K)}(t)=+\infty$.
\end{metric}
\end{oframed}

In other words, rank-order variability $\mathcal{R}_{ij} \in [-1,1]$ captures the average directional drift in rank: positive values indicate upward trends, negative values downward trends, and values near zero little systematic change. Distributions of $\mathcal{R}_{ij}$ across individuals can be visualized with box or density plots (see Section~\ref{sec:illustr_moder}).

The metrics described above for evaluating covariance structures apply to continuous variables. For example, Metric~\ref{metric:variance} has no direct analogue in the discrete setting, since the variance of a finite discrete variable is typically determined directly by its expected value. To study covariance structures in discrete variables, especially for capturing autocorrelation, we use \textit{kernel-smoothed local transition probabilities}. These metrics quantify the likelihood of transitioning from one class (or state) to another over time.

\begin{oframed}
\begin{metric}[]\label{metric:trans}
The kernel-smoothed local transition probability from state $a \in \mathcal{C}$ to state $b \in \mathcal{C}$ is
$$
\pi_{jab}^{(K)}(t) = 
\sum_{i=1}^n \sum_{k=1}^{n_{ij}-1} 
\bar w_{ijk}^{(a)}(t)\,\mathbf{1}(x_{ijk}=a,\,x_{ij(k+1)}=b),
$$
where the weights are
$$
\bar w_{ijk}^{(a)}(t) = 
\frac{K_h\!\left(t,\,t-d_{ijk}(t)\right)\,\mathbf{1}(x_{ijk}=a)}
{\sum_{i'=1}^n \sum_{k'=1}^{n_{i'j}-1} 
K_h\!\left(t,\,t-d_{i'jk'}(t)\right)\,\mathbf{1}(x_{i'jk'}=a)},
$$ and $d_{ijk}(t) = \max\{|t-t_{ijk}|,\,|t-t_{ij(k+1)}|\}$.
\end{metric}    
\end{oframed}

The resulting $t$-indexed collection of transition matrices $\Pi_j^{(K)}(t) = [\pi_{jab}^{(K)}(t)]_{a,b\in\mathcal{C}}$ captures the locally smoothed probabilities. These transition probabilities reflect both \textit{class stability}, i.e., self-transitions ($a \rightarrow a$) and change across states and are particularly useful for assessing temporal preservation in discrete variables. A variable is said to be \textit{time-homogeneous}, analogous to stationarity in continuous settings, if its local transition probabilities remain approximately constant across time. We visualize local transitions using area charts (see Section~\ref{sec:illustr_categ}).

\subsection{Metrics to assess the individual structure}\label{sec:ind_struc}

Aside from the variogram (Metric \ref{metric:vario}) and rank-order variability (Metric~\ref{metric:rank_var}), the previously discussed metrics offer limited insight into individual structure, i.e., the subject-level data. Additionally, phenomena observed in other metrics can often be explained more clearly by examining the individual trajectories. However, analyzing subject-level data numerically can be tedious and challenging to interpret. Therefore, we focus on visual inspection of \textit{subject-specific trajectories} through line charts (see Section~\ref{sec:illustr_moder}
). 

When dealing with large sample sizes, plotting all trajectories in a single plot becomes impractical as it can be difficult to follow each trajectory. Thus, stratification combined with subsampling is recommended. Stratification can be based on known strata in the data, such as treatment and placebo groups, or, as adopted in this work, on subjects’ baseline (i.e., first observed) smoothed quantile ranks. This approach facilitates examining whether individuals in different parts of the baseline distribution, such as those in the lower and upper percentiles, exhibit different temporal patterns compared to those near the median.

\subsection{Metrics to assess the measurement structure} \label{sec:metrics_struc}

To evaluate how well synthetic data reproduces the measurement structure of the original data, we use three complementary metrics: measurement density, measurement similarity, and dropout divergence. The \textit{measurement density} characterizes how measurements of a given variable are distributed over time. At each time point $t$, it represents the proportion of all measurements of variable $j$ occurring at $t$, relative to the total number of measurements across the common time grid $T_j^\ast$. Using this common grid allows accounting for both missing observations at original time points and additional observations that appear only in the synthetic data.

\begin{oframed}
\begin{metric}[]\label{metric:meas_dens}
For variable $j$, the measurement density at time point $t \in T_j^\ast$ is
$$
\mathcal{MD}_{jt} =
\frac{\sum_{i=1}^{n} I_{ijt}}
     {\sum_{s \in T_j^\ast} \sum_{i=1}^{n} I_{ijs}}.
$$
\end{metric}
\end{oframed}

Here, the numerator counts the number of subjects observed for variable $j$ at time $t$, while the denominator normalizes by the total number of observed measurements across all time points. As a result, $\mathcal{MD}_{jt}$ reveals which time points dominate the measurement schedule. A uniform density implies evenly spaced data collection, while peaks or troughs indicate periods of concentrated or scarce measurements. Owing to its intuitive interpretation, measurement density is well-suited for visualization through line charts (see Section~\ref{sec:illustr}).

To assess how closely the synthetic measurement patterns resemble the original, we compare their respective measurement indicator matrices (Section~\ref{sec:notations}) using \textit{measurement similarity}. Because subjects in the synthetic data do not directly correspond to those in the original, the rows of the synthetic matrix must be reordered. We obtain this optimal alignment with the Hungarian algorithm \cite{Kuhn1955TheProblem}, which finds a row permutation that minimizes the Euclidean distance between the rows of the real and the synthetic data in complexity $\mathcal{O}(n^3)$ \cite{Tomizawa1971OnProblems,Edmonds1972TheoreticalProblems}. The resulting minimal distance also serves as a secondary measure of resemblance between the two matrices.

\begin{oframed}
\begin{metric}[]\label{metric:sim} The measurement similarity of variable $j$ is the normalized absolute distance between the original and synthetic measurement indicator matrices after optimal row alignment: 
$$
\mathcal{MS}_j \;=\; 1 - \frac{1}{n \cdot |T_j^\ast|} \sum_{i=1}^{n} \sum_{t \in T_j^\ast} \big| I_{ijt} - \tilde{I}_{ijt} \big|. 
$$
\end{metric}    
\end{oframed}

The similarity score is normalized by the total number of entries in the measurement indicator matrix, ensuring that $\mathcal{MS}_j \in [0, 1]$. A value of $\mathcal{MS}_j = 1$ indicates perfect agreement between the original and synthetic measurement patterns, while $\mathcal{MS}_j = 0$ reflects complete dissimilarity. 

In LPD, dropouts and loss to follow-up are common. To evaluate how well synthetic data reproduces these patterns,  we introduce the \textit{dropout divergence}.

\begin{oframed}
\begin{metric}[]\label{metric:drop}
Let $P_j(t)$ and $\tilde{P}_j(t)$ denote the empirical distributions\footnotemark of dropout points $\{d_{ij}\}_{i=1}^n$ in the original data and synthetic data, respectively. The dropout divergence is the Kullback--Leibler divergence between $P_j(t)$ and $\tilde{P}_j(t)$:
$$
\mathcal{D}_j(P_j \,\|\, \tilde{P}_j) = \sum_{t \in T_j^\ast} P_j \log \left( \frac{P_j(t)}{\tilde{P}_j(t)} \right).
$$
\end{metric}    
\end{oframed}
\footnotetext{In practice, positivity of the empirical probabilities is ensured by adding a small constant $\varepsilon>0$ to all probability masses before normalization.}

Both measurement-structure metrics rely on finding an optimal permutation and therefore require the corresponding measurement matrices to have identical dimensions, in contrast to other metrics that do not impose this requirement. Accordingly, if the number of synthetic subjects differs from that of the original data, random subsampling of an equal number of rows from both matrices can be used to ensure dimensional compatibility. This subsampling approach is also useful when the Hungarian algorithm becomes computationally expensive for large sample sizes. The procedure should be repeated multiple times, with the resulting scores averaged, to reduce variability introduced by random sampling, as described in Section~\ref{sec:illustr}.

Furthermore, the numerical scores obtained from the original and synthetic data are seldom identical, raising the question of what defines a ``good" score. In practice, these metrics can be computed on a hold-out subset or a resampled version of the original data, as done in Section~\ref{sec:illustr}, enabling comparisons between original-to-synthetic and original-to-original samples, with the aim that the former closely matches the latter.

\section{Empirical illustration} \label{sec:illustr}

This section illustrates the evaluation metrics introduced in Section~\ref{sec:evalm} using synthetic longitudinal patient data generated by two distinct SDG methods, HALO \cite{Theodorou2023SynthesizeModel} and Health Gym GAN (HGG) \cite{Kuo2022TheAlgorithms}. HALO, based on the GPT-2 language model, represents a relatively novel approach, whereas HGG employs a generative adversarial network (GAN), a more established SDG technique. Both methods support unbalanced longitudinal data with mixed continuous and categorical variables and are available through publicly released implementations \cite{Perkonoja2025MethodsReview}. Their handling of unbalanced data differs, however: HALO can generate such data directly, whereas HGG requires prior imputation and the use of indicator variables, with unbalancedness reintroduced after generation. Both HALO and HGG have been earlier evaluated using autocorrelation-based metrics. The author's of HALO reported that the method outperformed approaches such as EVA \cite{Biswal2021EVA:Autoencoders} and SynTEG \cite{Zhang2021}, whereas HGG demonstrated strong performance in a recent comparative study \cite{Isasa2024ComparativeSynthesis}. 

We generated synthetic data following the codebases released by the respective authors of HALO and HGG, including dataset selection (MIMIC-III \cite{Johnson2016MIMIC-III1.4,Johnson2016MIMIC-IIIDatabase}), cohort definitions (HALO: 32,060 subjects, 17 variables; HGG: 3,910 subjects, 13 clinical variables and 7 missingness indicator variables), and preprocessing procedures. The results presented here are intended to illustrate the proposed metrics rather than to enable a direct comparison between the methods, as such a comparison would require, at a minimum, training on identical datasets.

To harmonize the illustrative examples between the methods, we applied additional post-processing to the synthetic data generated by HALO by converting time from measurement intervals to a continuous hourly scale and restricting the analysis to the first 48 hours. This choice was also motivated by data sparsity: depending on the variable, only 25--50\% of subjects in the HALO implementation had measurements at 48 hours, with coverage declining rapidly thereafter, while the acute hypotension subset \cite{Gottesman2020InterpretableTransitions} used for HGG is inherently limited to a 48-hour window. Additionally, bandwidth $h=6$ was used in all experiments.

Metrics~\ref{metric:sim} and \ref{metric:drop} (measurement similarity and drop-out divergence), together with the Frobenius norms of the measurement indicator matrices \eqref{def:measurement_indicator_matrix}, were computed using random samples of 2,000 subjects and repeated over 100 iterations to accommodate the computational complexity of the Hungarian algorithm. Reported values correspond to averages across iterations, with reference values shown in parentheses and obtained from separate random samples of the original measurement matrix.

To aid interpretation, Table~\ref{tab:illustr_overview} summarizes the illustrative examples presented in this section. Although the analyses were conducted for all variables, the examples shown were selected to highlight different observed phenomena, while variables not included here exhibited results similar to those presented. The following subsections provide a detailed discussion of each case.

\begin{table}[ht]
\footnotesize
\centering
\caption{Overview of illustrative examples used to demonstrate the proposed evaluation metrics.}
\label{tab:illustr_overview}
\begin{tabularx}{\textwidth}{  
  P{0.22\textwidth}
  P{0.09\textwidth}
  P{0.28\textwidth}
  Y}
\toprule
Variable & Method & Illustration focus & Main observation \\
\midrule
Systolic blood pressure & HALO & Moderate temporal preservation &
Autocorrelation and profiles preserved; mild outlier inflation \\
Weight & HALO & Covariance structure failure &
Loss of between-subject heterogeneity despite correct marginals \\
Respiratory rate & HALO & Covariance structure failure &
Autocorrelation preserved but variance and outliers exaggerated \\
Aspartate aminotransferase & HGG & Covariance structure failure &
Suppressed extremes and reduced variance \\
GCS score & HGG & Transition probability preservation &
Minority-class transitions unreliable \\
GCS Eye opening assessment & HALO & Class proportion distortion &
Extra class introduced in the synthetic data \\
FiO\textsubscript{2} & HGG & Class proportion distortion &
One class not reproduced in the synthetic data \\
\bottomrule
\end{tabularx}
\end{table}

\subsection{Moderate temporal preservation} \label{sec:illustr_moder}

\begin{figure}[H]
\centering
     \includegraphics[width=\linewidth]{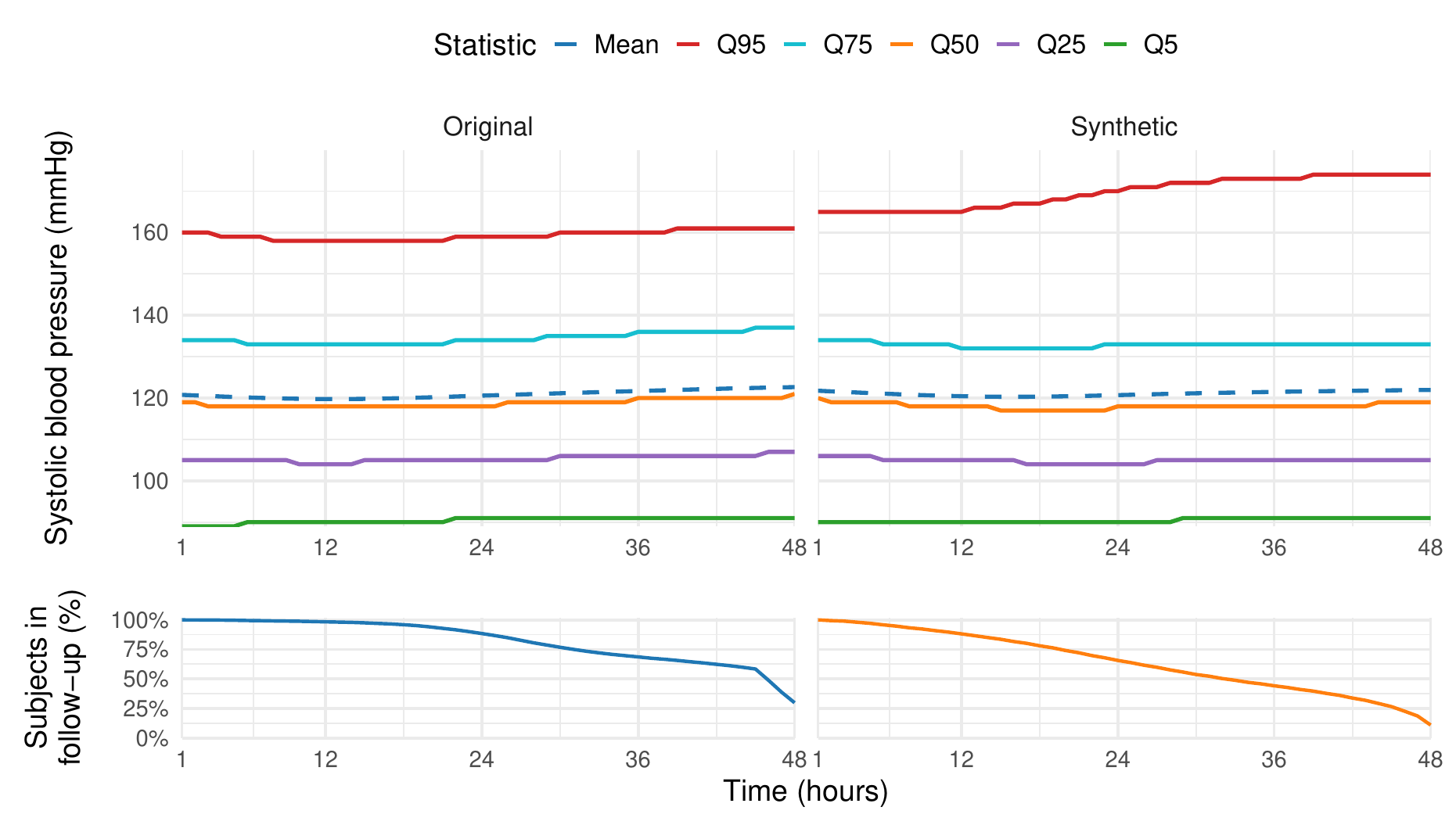}
    \caption{Kernel-smoothed mean (Metric~\ref{metric:mean_profile}) and quantile profiles (Metric~\ref{metric:quant_prof}) of systolic blood pressure for the original data (left) and HALO-generated synthetic data (right).}
    \label{fig:mean_quant_syst}
\end{figure}

We begin by examining \texttt{systolic blood pressure} generated by HALO, which demonstrates moderate temporal preservation according to our evaluation metrics. 
Figure~\ref{fig:mean_quant_syst} shows that the kernel-smoothed mean profile (Metric~\ref{metric:mean_profile}) and quantile profiles (Metric~\ref{metric:quant_prof}) of synthetic data closely match those of the original data. The main deviation occurs at the 95th percentile, which is higher in the synthetic data, indicating slightly deviating outlier patterns. No clear trends or periodic patterns emerge. As data sparsity can influence synthetic data generation, the bottom panels of Figure~\ref{fig:mean_quant_syst} also show the proportions of patients under follow-up over time. The original data shows a sharp drop in the number of patients just before 48 hours, whereas the synthetic data exhibit a more gradual decline. 

\begin{figure}[H]
    \centering
    \includegraphics[width=\linewidth]{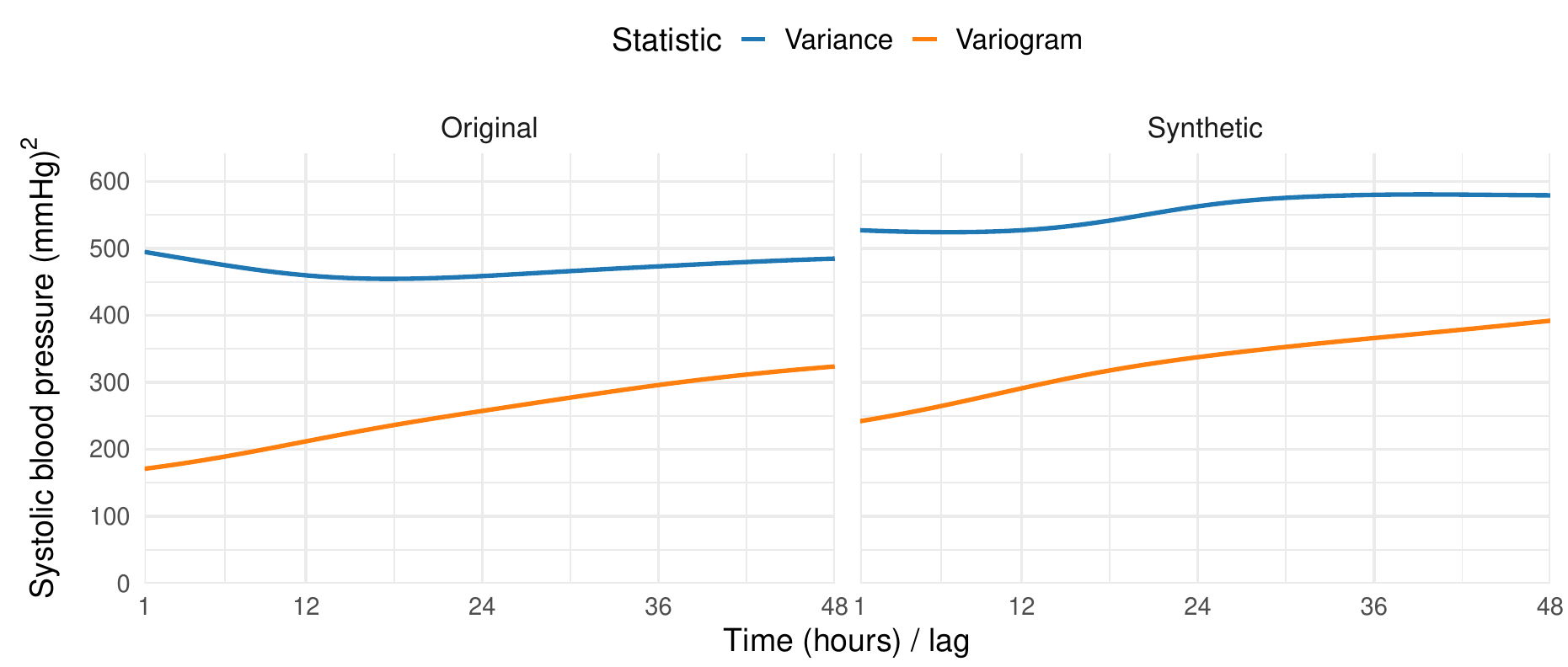}
    \caption{Kernel-smoothed variance profile (Metric~\ref{metric:variance}, blue) and variogram (Metric~\ref{metric:vario}, orange) of systolic blood pressure for the original data (left) and synthetic data generated by HALO (right).}  
    \label{fig:variogram_syst}
\end{figure}

The relatively stable variance profile (Metric~\ref{metric:variance}, Figure~\ref{fig:variogram_syst}, upper curve) suggests a stationary process that is reasonably preserved in the synthetic data. The variogram (Metric~\ref{metric:vario}, Figure~\ref{fig:variogram_syst}, lower curve) further indicates gradually declining autocorrelation and, because it does not reach the variance profile, the presence of between-subject variability, both of which are well preserved in the synthetic data.

\begin{figure}[H]
    \centering
    \includegraphics[width=\linewidth]{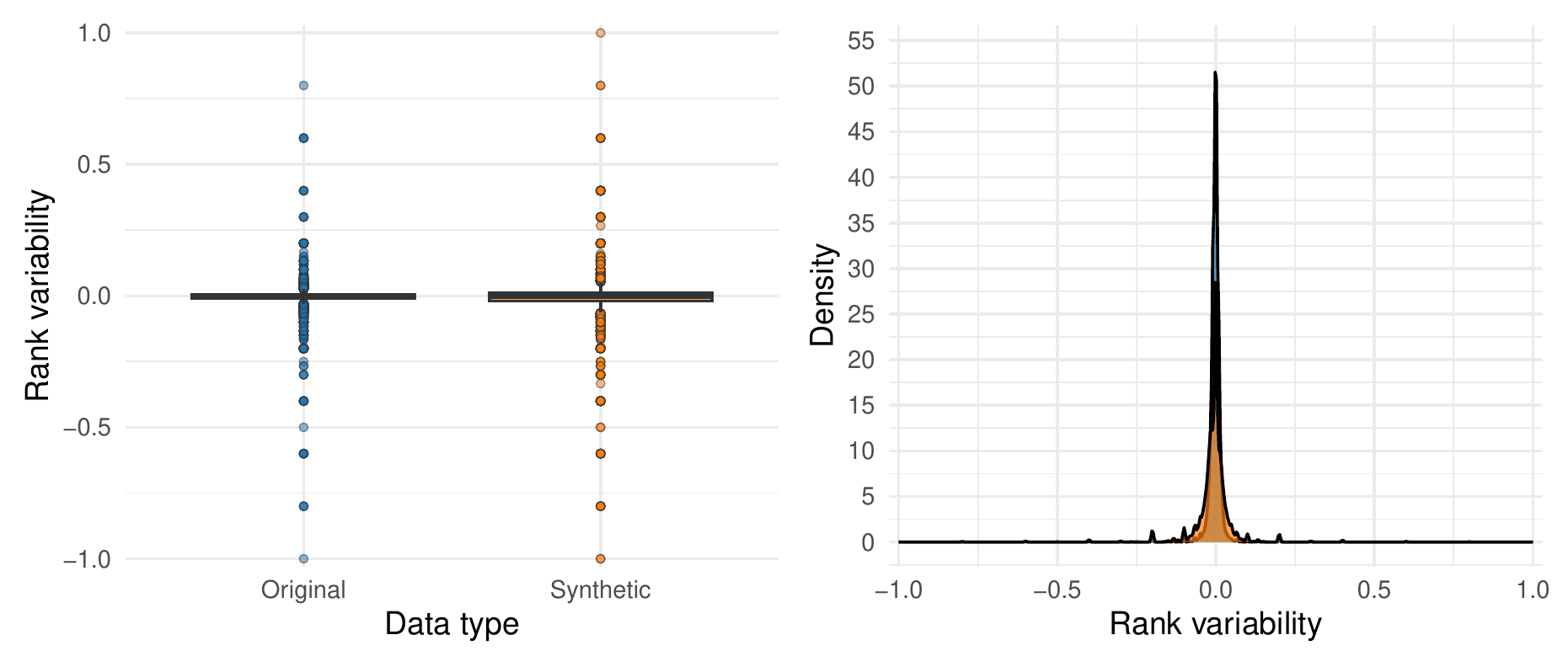}
    \caption{Boxplot (left) and density plot (right) showing the kernel-smoothed rank-order variability (Metric~\ref{metric:rank_var}) of systolic blood pressure for original (blue) and HALO-generated synthetic data (orange)}
    \label{fig:rank_order_syst}
\end{figure}

To further examine the covariance structure of systolic blood pressure, we analyzed the kernel-smoothed rank-order variability (Metric~\ref{metric:rank_var}, Figure~\ref{fig:rank_order_syst}). In both datasets, profiles are centered near zero, indicating that individuals generally maintain stable rankings relative to baseline, consistent with the autocorrelation observed in the variogram. 

\begin{figure}[H]
    \centering
    \includegraphics[width=\textwidth]{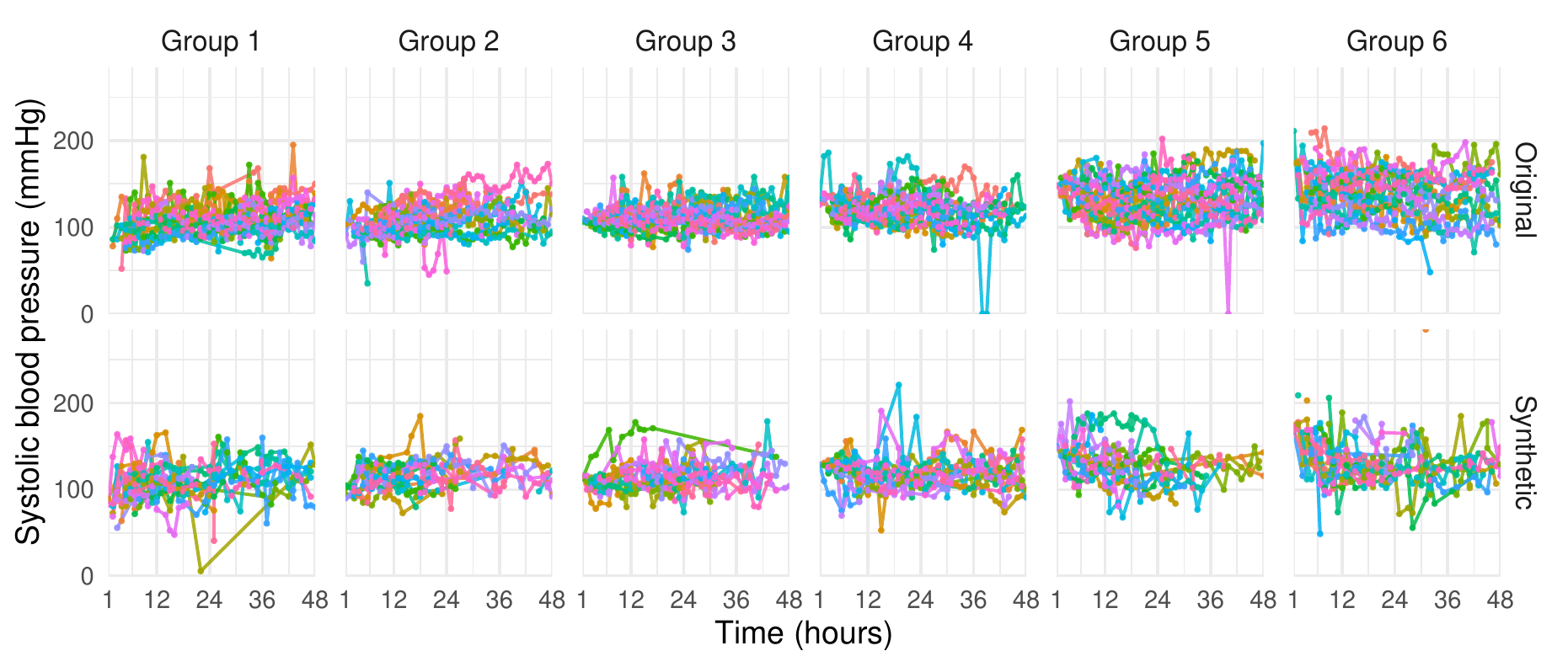}
    \caption{Subject-specific systolic blood pressure trajectories by baseline group for the original (top) and HALO-generated synthetic data (bottom).}
    \label{fig:indiv_traj_syst}
    \end{figure}

Subject-level trajectories (Section~\ref{sec:ind_struc}) can offer deeper insights into population-level metrics. For HALO, the raw unsmoothed trajectories (Figure~\ref{fig:indiv_traj_syst}), presented in six baseline quantile-defined percentile classes $[0, 5]$, $(5, 25]$, $(25, 50]$, $(50, 75]$, $(75, 95]$, $(95, 100]$, resemble the original data overall, with slightly more variation in groups 1 and 6. These patterns are directly connected to the increased variance in Figure \ref{fig:variogram_syst} and the elevated 95th percentile in Figure \ref{fig:mean_quant_syst}.

\begin{figure}[t!]
    \centering
    \includegraphics[width=\textwidth]{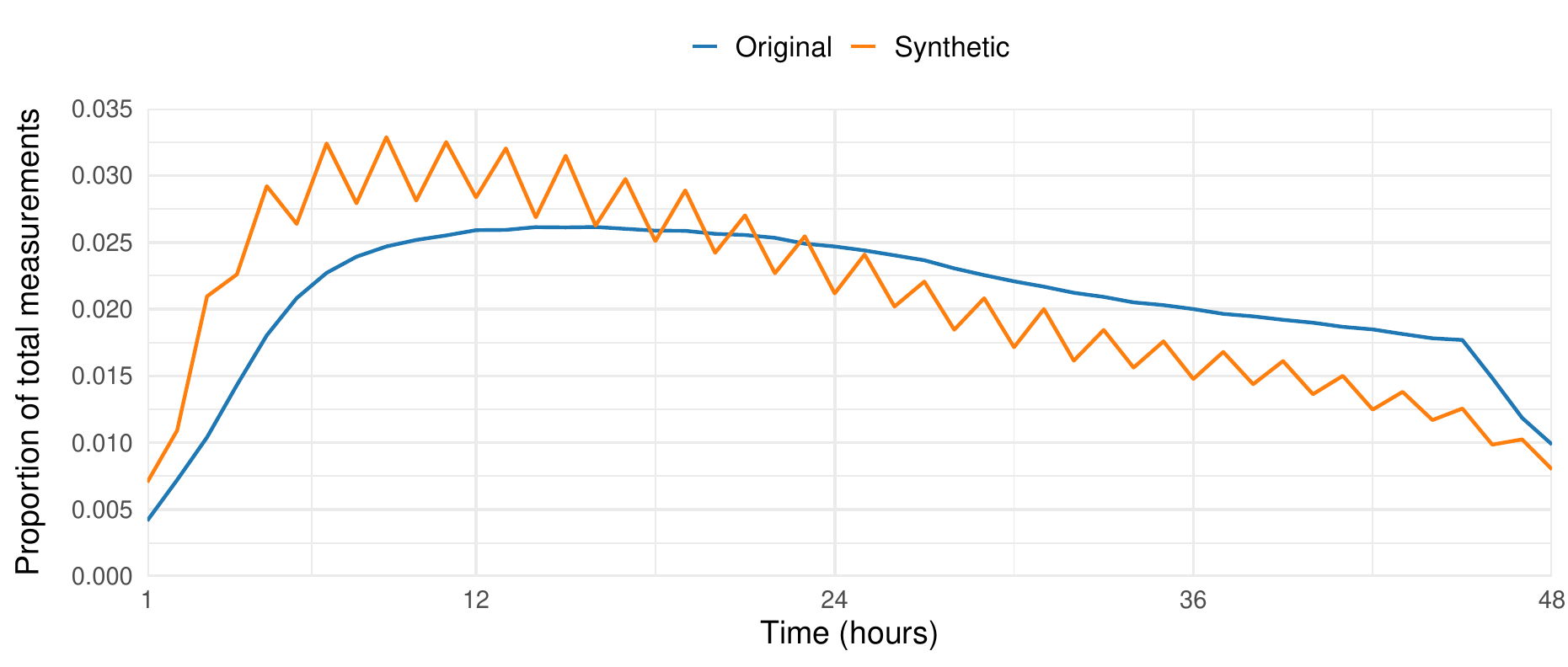}
    \caption{Measurement density (Metric~\ref{metric:meas_dens}) of systolic blood pressure for the original data (blue) and HALO-generated synthetic data (orange).}
    \label{fig:meas_dens_syst}
\end{figure}

Figure~\ref{fig:meas_dens_syst} shows the measurement density (Metric~\ref{metric:meas_dens}) for both datasets. In the synthetic data, observations are more frequent at earlier time points but become more sparse after about 18 hours, whereas in the original data, measurements increase rapidly up to 6 hours and then gradually decrease, with a sharp drop after 42 hours. The synthetic series also displays strong fluctuations between adjacent time points, forming a zigzag pattern that suggests a periodic artifact of the data generating mechanism.

The mean Frobenius norm (smaller is better) between the original and synthetic measurement matrices was 173.42 (reference 67.09). The mean measurement similarity score (Metric~\ref{metric:sim}, larger is better) was 0.69 (reference 0.95), and the mean drop-out divergence (Metric~\ref{metric:drop}, smaller is better) was 0.31 (reference 0.03). Compared to the natural variation within the real data, the synthetic data shows noticeably larger differences in measurement structure, as indicated by the higher Frobenius norm and drop-out divergence. Nevertheless, the measurement similarity remains comparatively high, aligning with the patterns identified through the other metrics.

\subsection{Problems in preserving the covariance structure}

In this section, we illustrate different failure modes of covariance structure preservation using \texttt{weight} and \texttt{respiratory rate} generated by HALO and \texttt{aspartate aminotransferase} (AST) generated by HGG as examples. The results illustrate that closely matching numerical scores and marginal behavior can still coincide with substantial differences in covariance structure in the synthetic data.

\begin{figure}[p]
    \centering
    \begin{subfigure}{0.78\linewidth}
        \includegraphics[width=\linewidth]{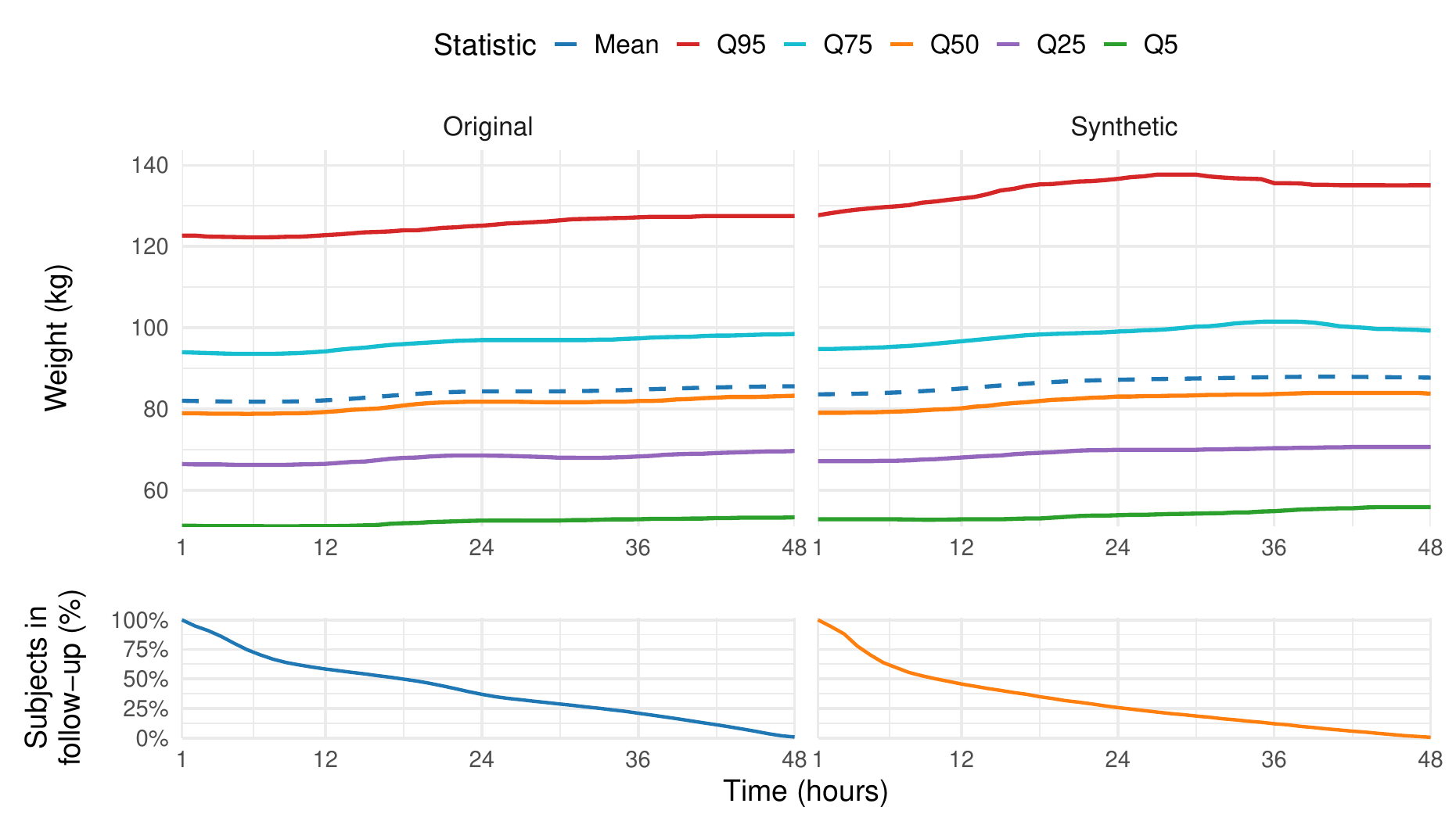}
        \caption{}
         \label{fig:mean_quant_weight}
    \end{subfigure}

    \begin{subfigure}{0.78\linewidth}
      \includegraphics[width=\linewidth]{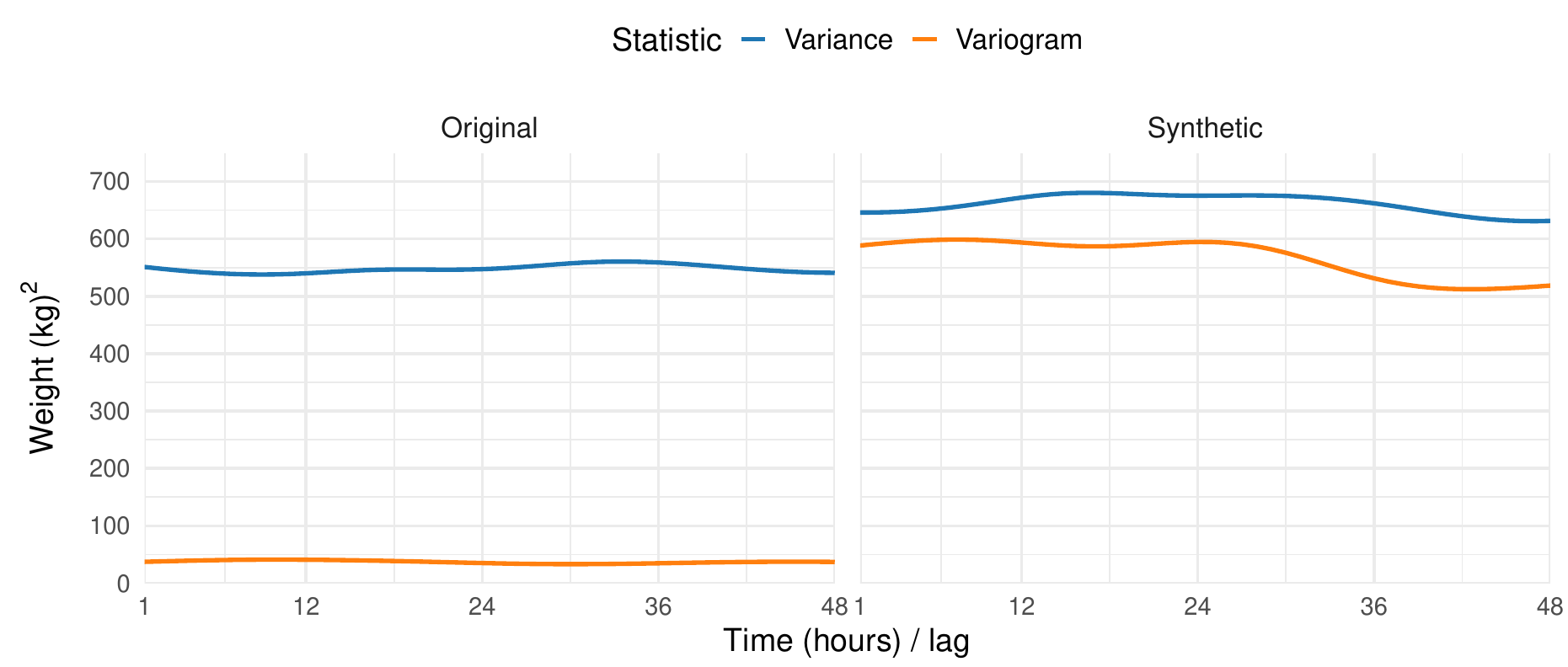}
        \caption{}  
        \label{fig:variogram_weight}
    \end{subfigure}

    \begin{subfigure}{0.78\linewidth}
        \includegraphics[width=\linewidth]{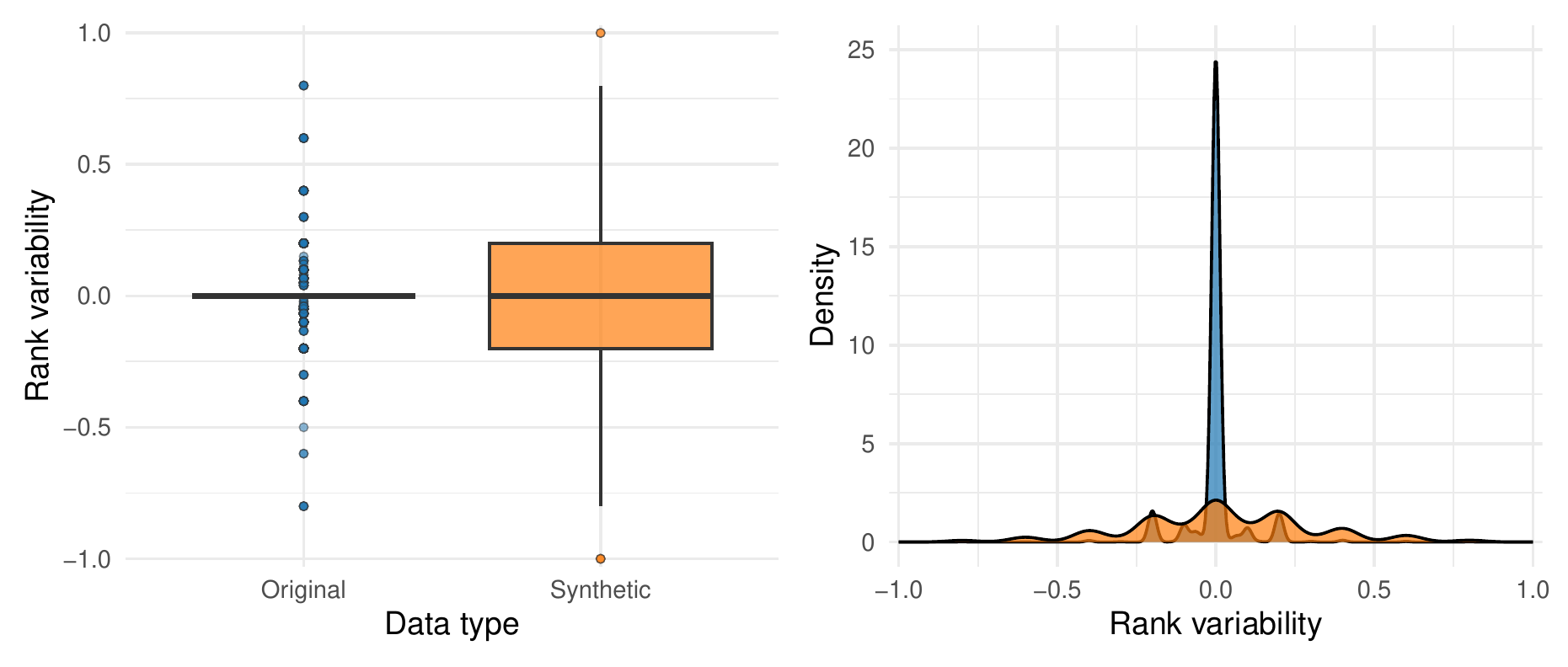}
        \caption{}
        \label{fig:rank_order_weight}
    \end{subfigure}
    \caption{Panel (a) shows the kernel-smoothed mean and quantile profiles, panel (b) the variance profile and variogram (in squared units), and panel (c) the distribution of rank-order variability of weight for the original data (left) and HALO-generated synthetic data (right).}
\end{figure}

\paragraph{Changes in variability decomposition} Figure \ref{fig:mean_quant_weight} shows the mean and quantile profiles of the variable weight generated by HALO, which appear remarkably similar between the original and synthetic datasets. However, the variance profile and variogram in Figure \ref{fig:variogram_weight} reveal critical differences. While the variance profiles largely agree, the variograms differ substantially.

The low and flat variogram in the left panel appropriately captures the fact that individual weight remains stable during the 48-hour follow-up, with almost all variability in weight following from between-subject variability. In contrast, the synthetic data show a markedly different decomposition of variability, with reduced between-subject heterogeneity and a large contribution from measurement error. This change in covariance structure is also evident in the failure to reproduce rank-order variability (Figure \ref{fig:rank_order_weight}).

\begin{figure}[H]
    \centering
    \includegraphics[width=\linewidth]{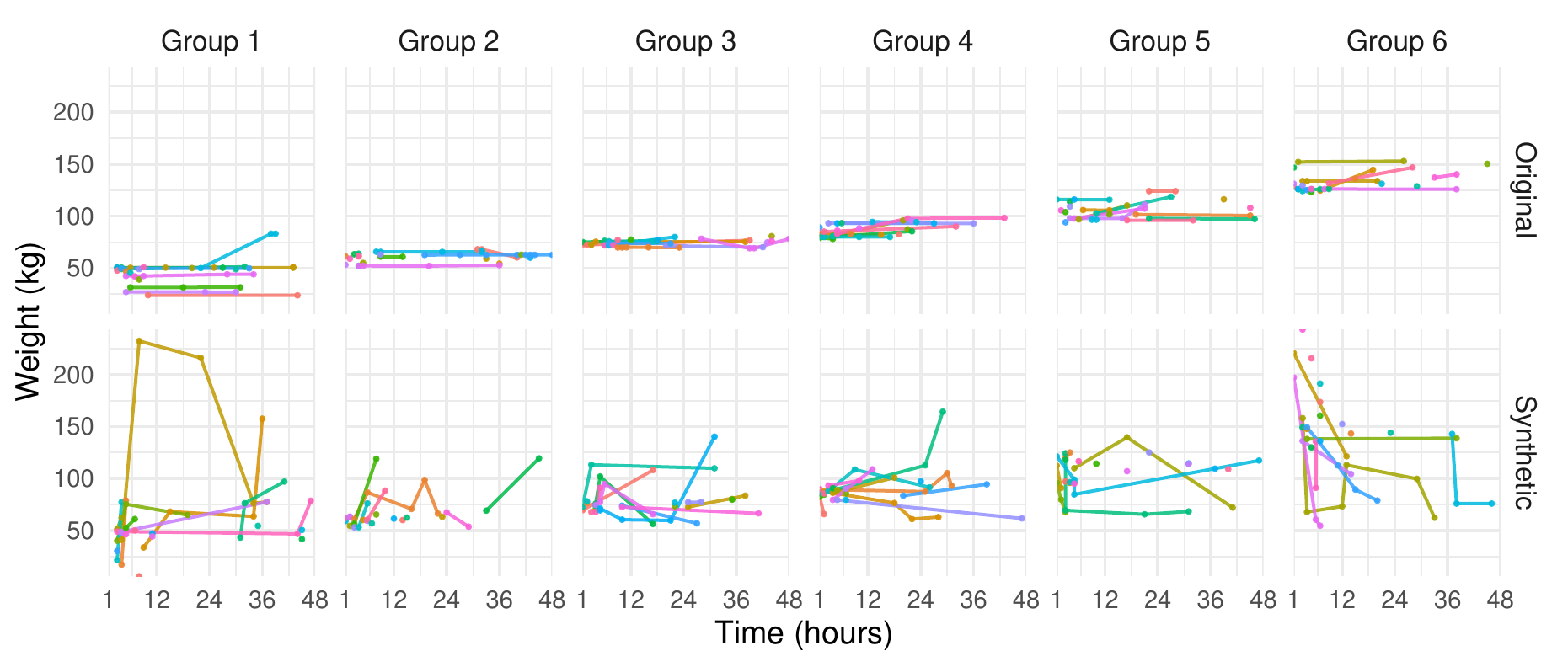}
    \caption{Subject-specific weight (kg) trajectories by baseline group for the original (top) and HALO-generated synthetic data (bottom).}
    \label{fig:indiv_traj_weight}
\end{figure}

\newpage
The subject-specific trajectories shown in Figure~\ref{fig:indiv_traj_weight} provide further insight into the differences in covariance structure. In the synthetic data, measurements exhibit pronounced short-term fluctuations and appear largely as random noise, with little evidence of consistent temporal patterns within individuals. The synthetic data also contains a larger number of subjects with only a single observation, visible as isolated points in Figure~\ref{fig:indiv_traj_weight}. Importantly, these differences are not driven by outliers as the overall distribution of values is well preserved (Supplemental Figure~\ref{SM:fig_mean_quantiles_weight_out}).

The measurement density closely matches the original distribution (Supplemental Figure~\ref{SM:fig_meas_dens_weight}), although the synthetic data show a tendency toward more frequent measurements early on. The mean Frobenius norm between the original and synthetic measurement matrices was 35.21 (reference 30.63), the mean measurement similarity score was 0.99 (reference 0.99), and the mean drop-out divergence was 0.14 (reference 0.02). Overall, aside from a slightly higher dropout divergence, these numerical scores closely match the reference values.

\begin{figure}[H]
    \centering
    \begin{subfigure}{0.78\linewidth}
       \includegraphics[width=\textwidth]{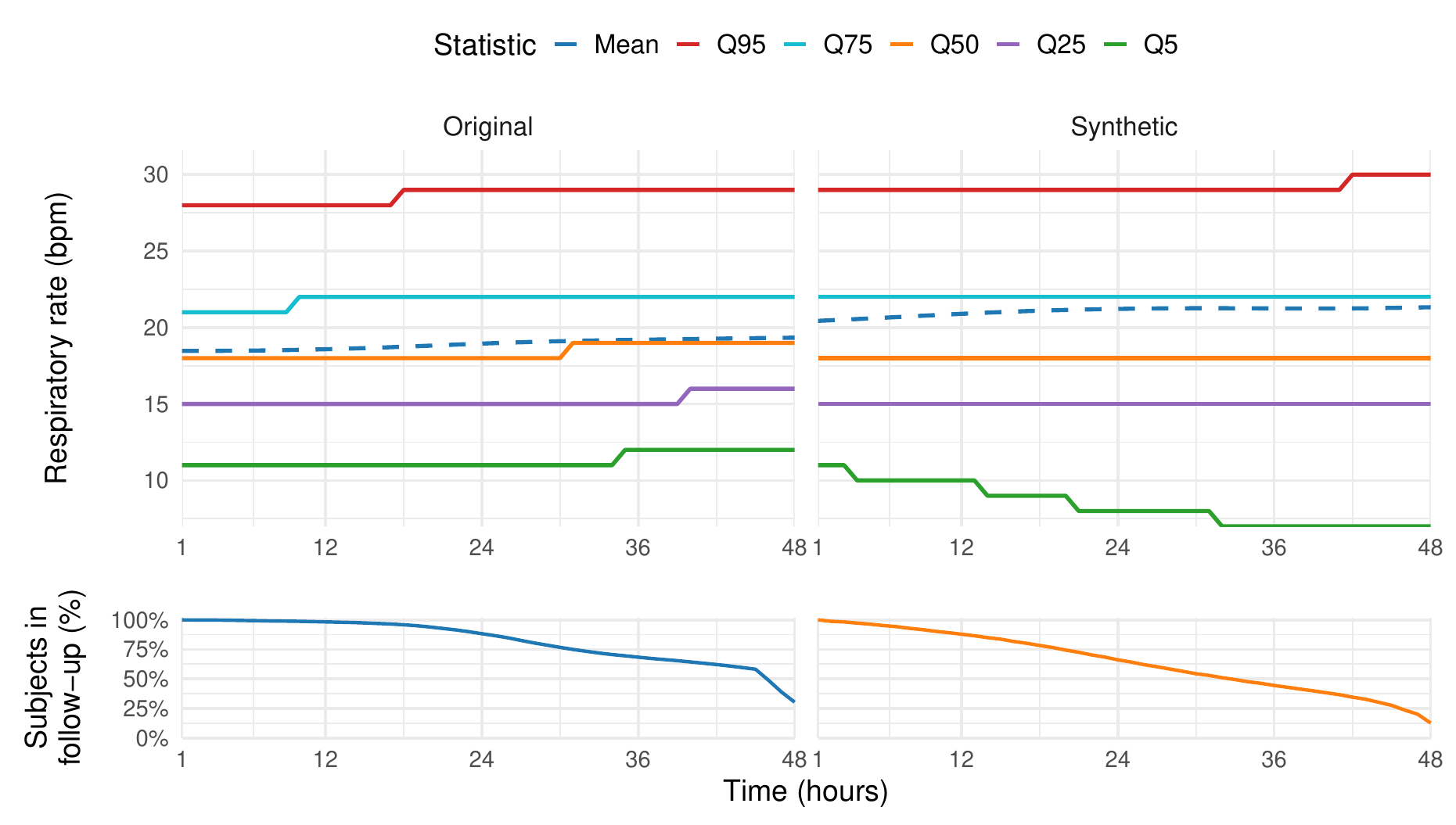}
    \caption{}
    \label{fig:mean_quant_resp}
    \end{subfigure}

    \begin{subfigure}{0.78\linewidth}
          \includegraphics[width=\textwidth]{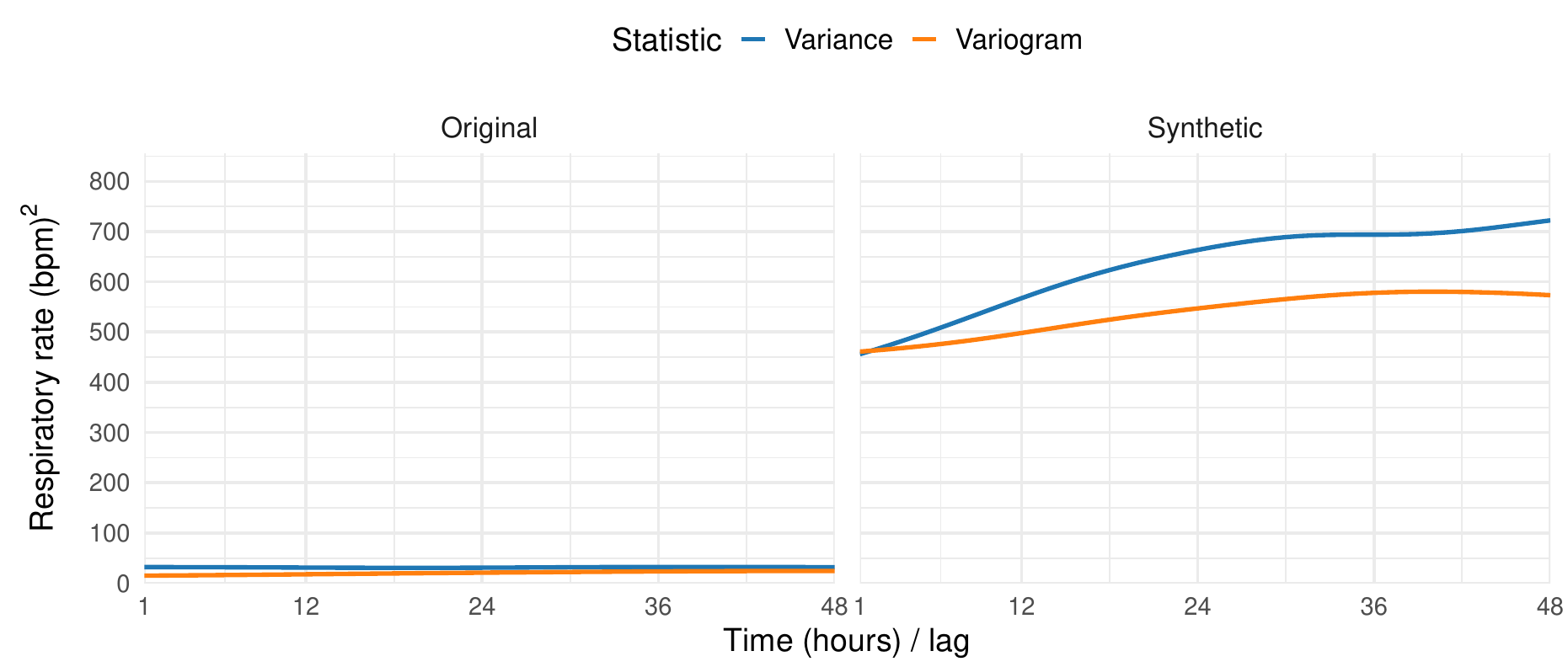}
    \caption{}
    \label{fig:variogram_resp}
    \end{subfigure}

    \begin{subfigure}{0.78\linewidth}
        \includegraphics[width=\textwidth]{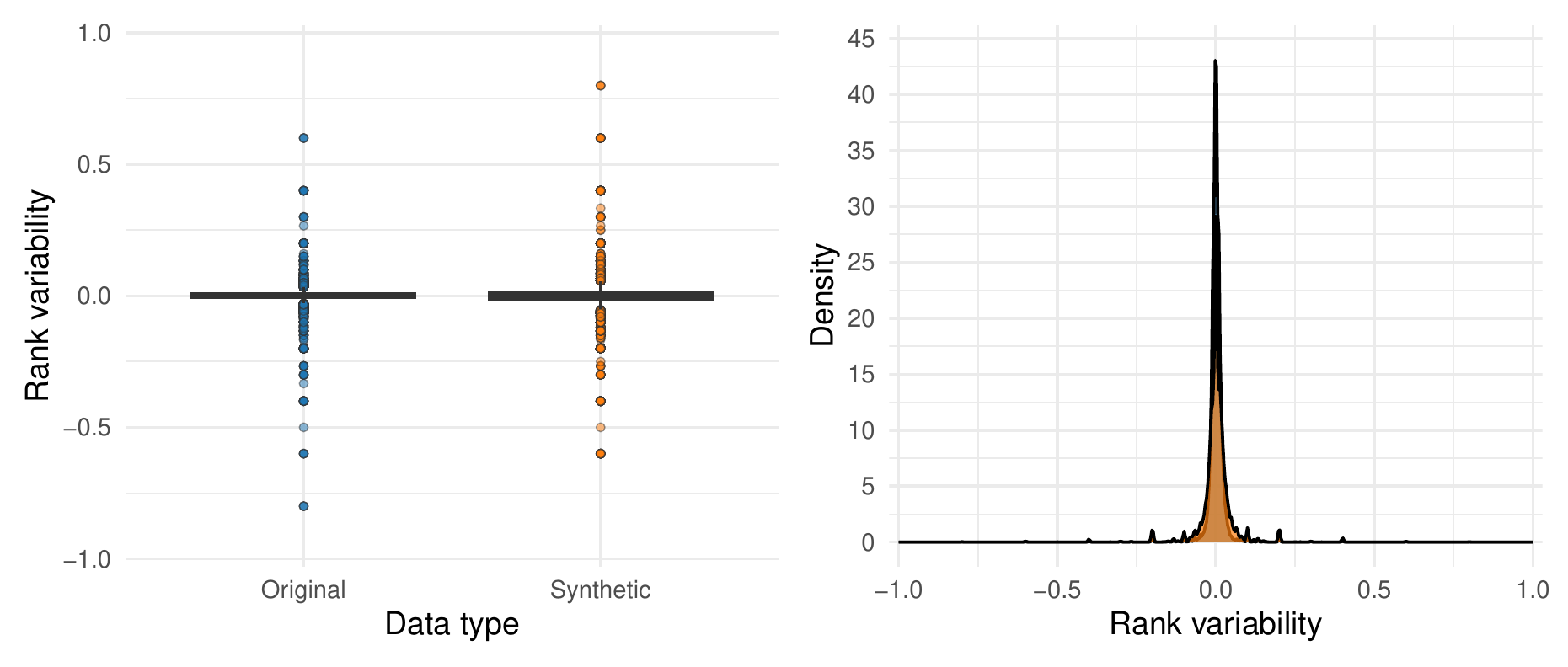}
    \caption{}
    \label{fig:rank_order_resp}
    \end{subfigure}
    \caption{Panel (a) shows the kernel-smoothed mean and quantile profiles, panel (b) the variance profile and variogram (in squared units), and panel (c) the distribution of rank-order variability of respiratory rate for the original data and HALO-generated synthetic data.}
\end{figure}

\paragraph{Inflated variance due to outliers} Figure~\ref{fig:mean_quant_resp} shows the mean-quantile profiles of the respiratory rate generated by HALO. The relative positions of quantiles appear largely consistent between the original and synthetic datasets. However, the synthetic data displays a slightly higher mean and a noticeably lower 5th percentile, which deviates from the stable trend seen in the original data.

\clearpage Figure~\ref{fig:variogram_resp} reveals that the variance profile in the HALO-generated synthetic data is substantially inflated. Yet, the rank-order variability (Figure~\ref{fig:rank_order_resp}) suggest that individual rankings are largely preserved, i.e., autocorrelation structure is reasonably maintained. Indeed, the shape of the variogram relative to the variance scale is adequately preserved in the synthetic data (Supplemental Figure~\ref{SM:fig_resp}).

\begin{figure}[H]
\centering
    \begin{subfigure}{0.88\linewidth}
    \centering
    \includegraphics[width=\textwidth]{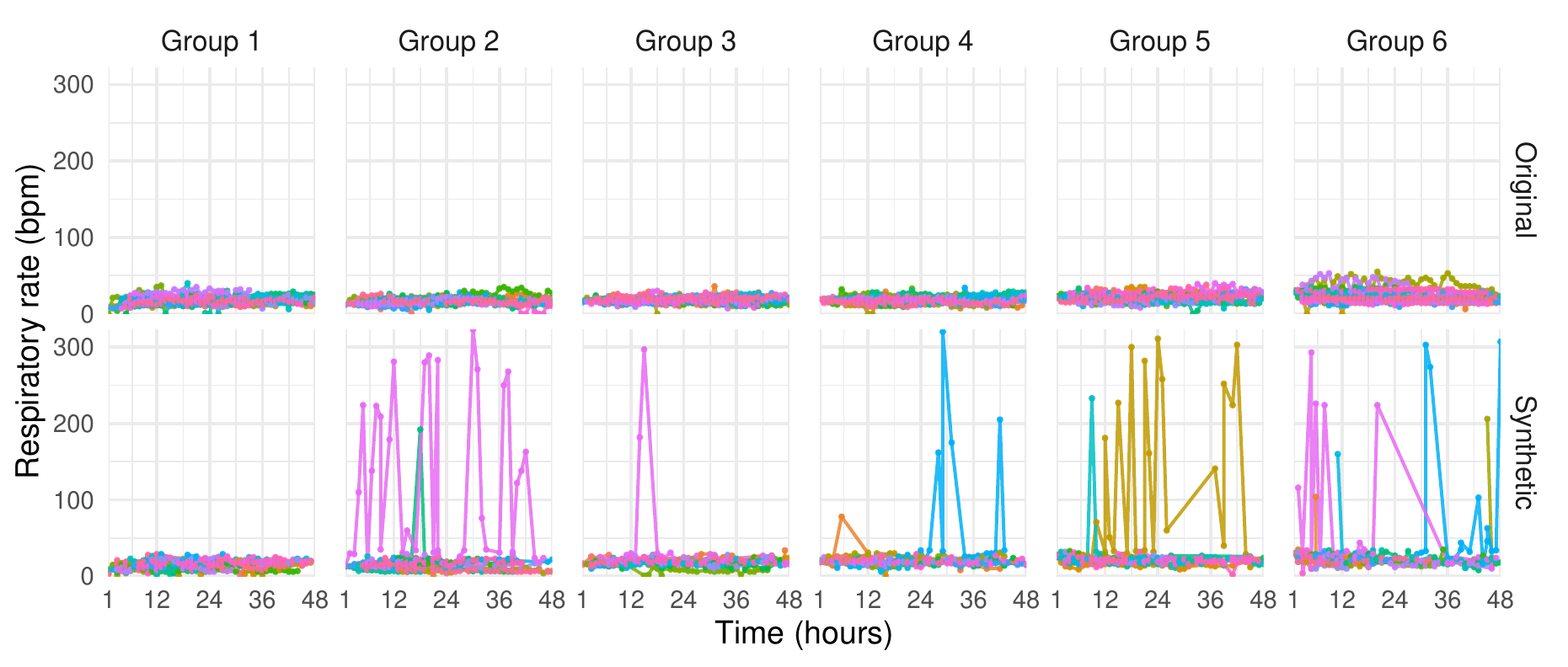}
    \caption{}
    \label{fig:indiv_traj_resp}
    \end{subfigure}

    \begin{subfigure}{0.88\linewidth}
        \centering
    \includegraphics[width=\textwidth]{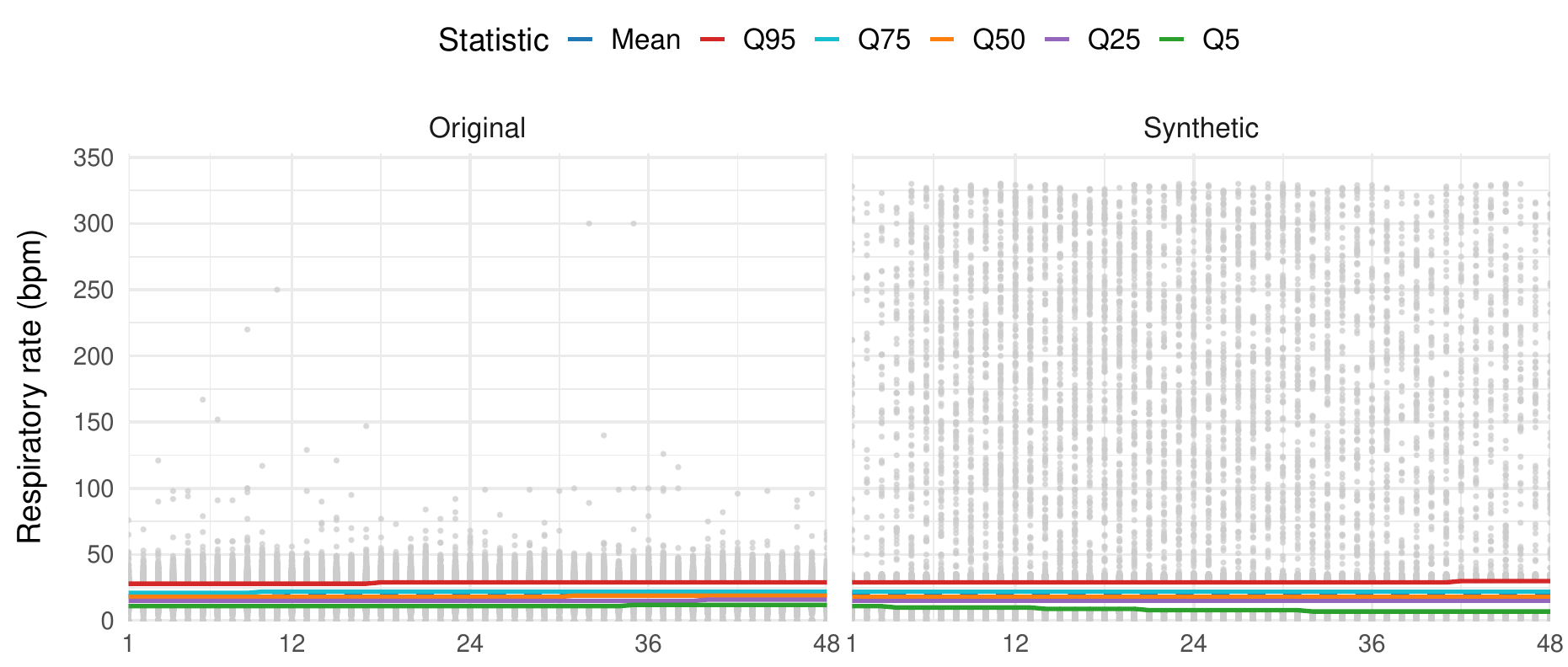}
    \caption{}
    \label{fig:mean_quant_resp_out}
    \end{subfigure}
    \caption{Panel (a) shows subject-specific respiratory rate trajectories by baseline group for the original and HALO-generated synthetic data, while panel (b) presents the mean–quantile profiles from Figure~\ref{fig:mean_quant_resp} augmented with observations below the 5th percentile and above the 95th percentile.}
    \label{fig:indiv_outlier_resp}
\end{figure}

An examination of subject-specific trajectories (Figure \ref{fig:indiv_traj_resp}) shows that while some synthetic subjects follow patterns resembling those in the original data, many exhibit abrupt or erratic changes. This supports the earlier conclusion that the SDG method has preserved some degree of autocorrelation.

When the mean-quantile profiles presented in Figure \ref{fig:mean_quant_resp} are supplemented by plotting observations below the 5th and above the 95th percentiles (Figure \ref{fig:mean_quant_resp_out}), we see that the synthetic data contains outliers across a much wider range. This helps explain the exaggerated variance profile and abrupt changes in the subject-specific trajectories. Moreover, the distribution of these outliers appears almost uniformly random, in contrast to the behavior seen in the original data.

The measurement density for respiratory rate generated by HALO (not shown) closely matches that of systolic blood pressure depicted in Figure \ref{fig:meas_dens_syst}. The mean Frobenius norm between the original and synthetic measurement matrices was 166.87 (reference 61.32), the mean measurement similarity score was 0.71 (reference 0.96), and the mean drop-out divergence was 0.28 (reference 0.03). These indicate non-negligible deviations in the measurement structure relative to the reference.

\begin{figure}[H]
       \includegraphics[width=\textwidth]{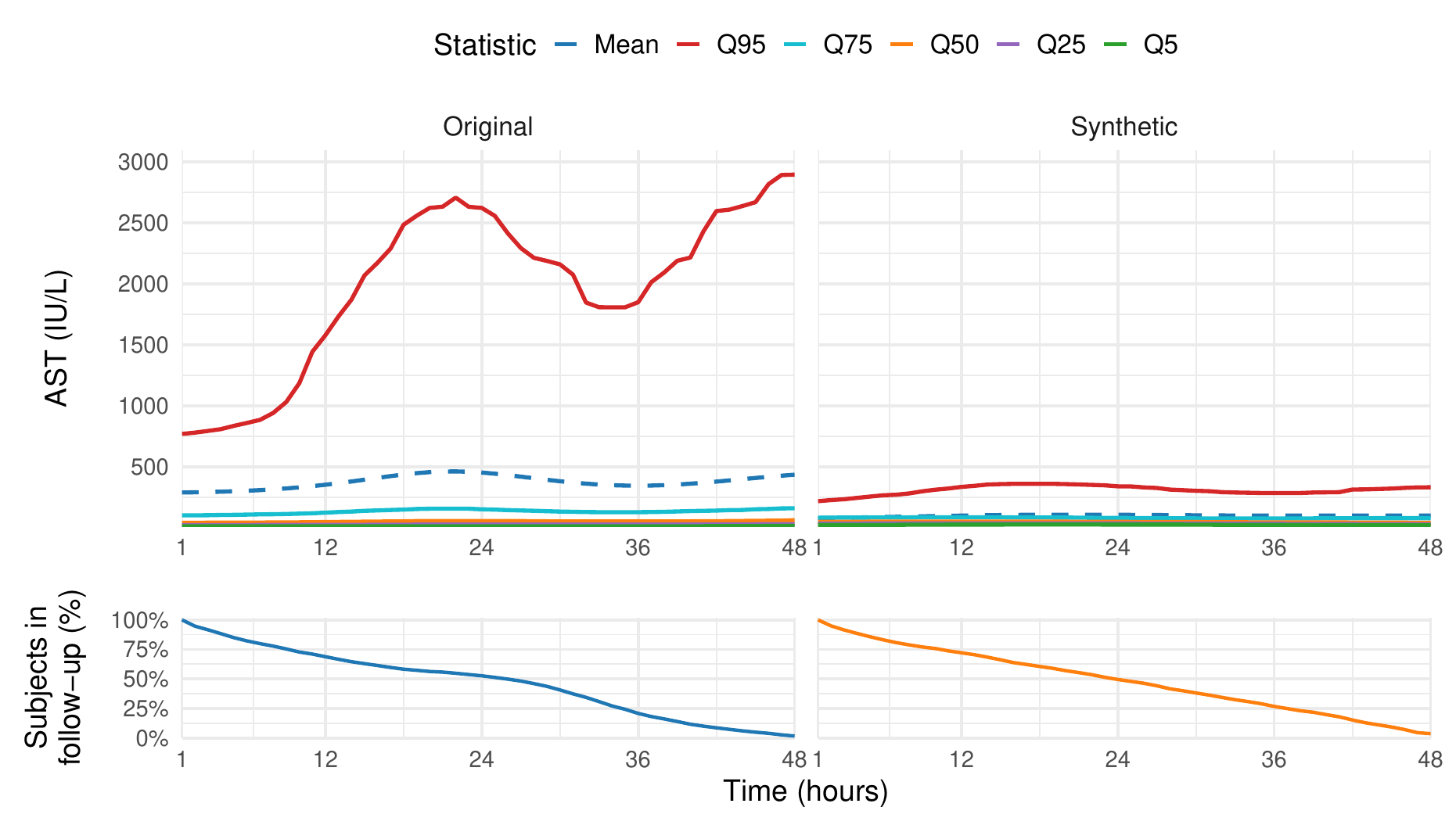}
       \caption{Kernel-smoothed mean and quantile profiles of aspartate aminotransferase for the original data and HGG-generated synthetic data.}
    \label{fig:mean_quant_ast}
\end{figure}

\paragraph{Deflated variance due to data sparsity} Figure~\ref{fig:mean_quant_ast} shows that the 95th percentile of aspartate aminotransferase generated by HGG is strongly suppressed and lacks the increasing trend present in the original data. 

The variance profile in Figure~\ref{fig:variogram_ast} is substantially deflated (see Supplemental Figure~\ref{SM:fig_ast} for a free y-axis scale). The variogram of the original data increases rapidly with lag and exceeds the total variance, indicating the presence of a temporal trend. This is not reproduced by the synthetic data. Moreover, the rank-order variability in Figure~\ref{fig:rank_order_ast} shows opposing behavior: values in the original data tend to decrease over time (variability skewed toward negative values), whereas those in the synthetic data increase.

\begin{figure}[t!]
\centering
\begin{subfigure}{\linewidth}
      \includegraphics[width=\textwidth]{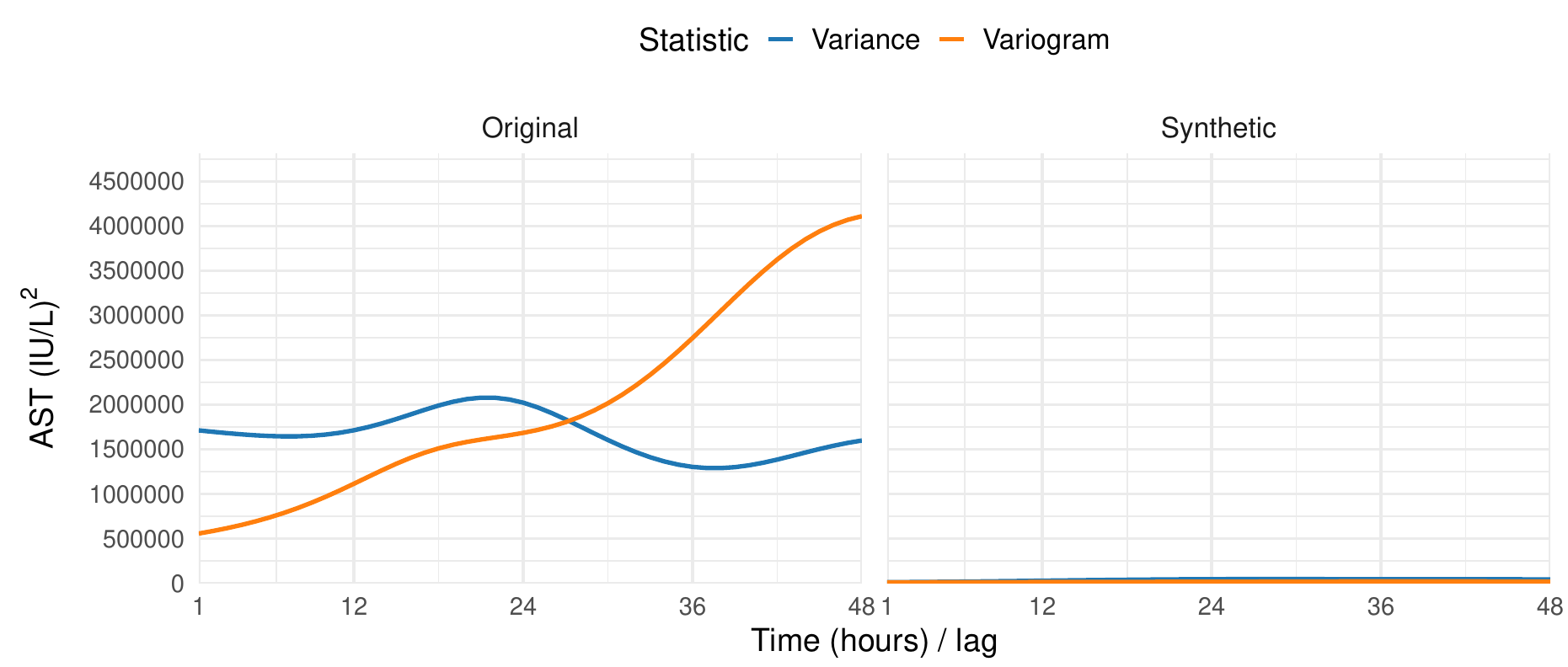}
      \caption{}
\label{fig:variogram_ast}
\end{subfigure}
\begin{subfigure}{\linewidth}
    \includegraphics[width=\textwidth]{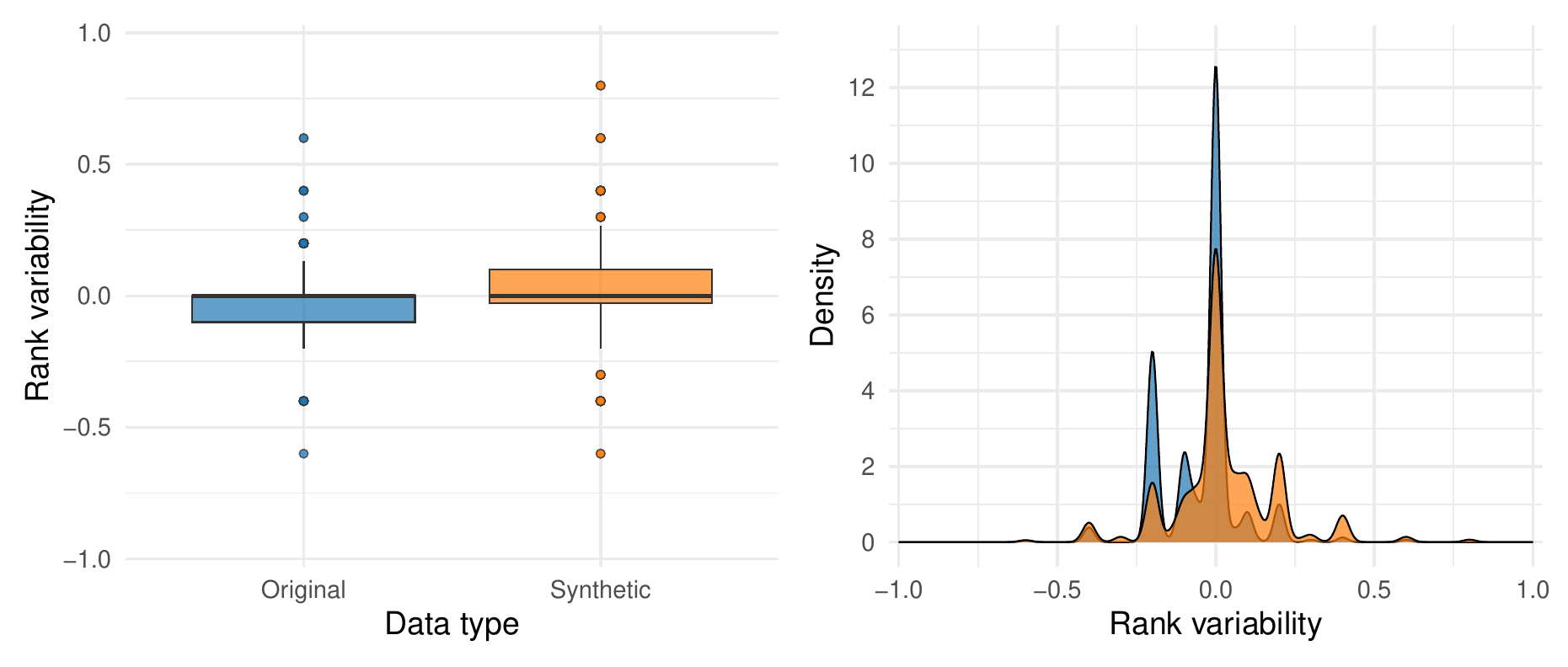}
    \caption{}
    \label{fig:rank_order_ast}
\end{subfigure}
\caption{Panel (a) shows the variance profile and variogram (in squared units), and panel (b) the distribution of rank-order variability of aspartate aminotrasferase for the original data and HGG-generated synthetic data.}
\end{figure}

\begin{figure}[H]
\centering
    \begin{subfigure}{0.88\textwidth}
    \centering
    \includegraphics[width=\textwidth]{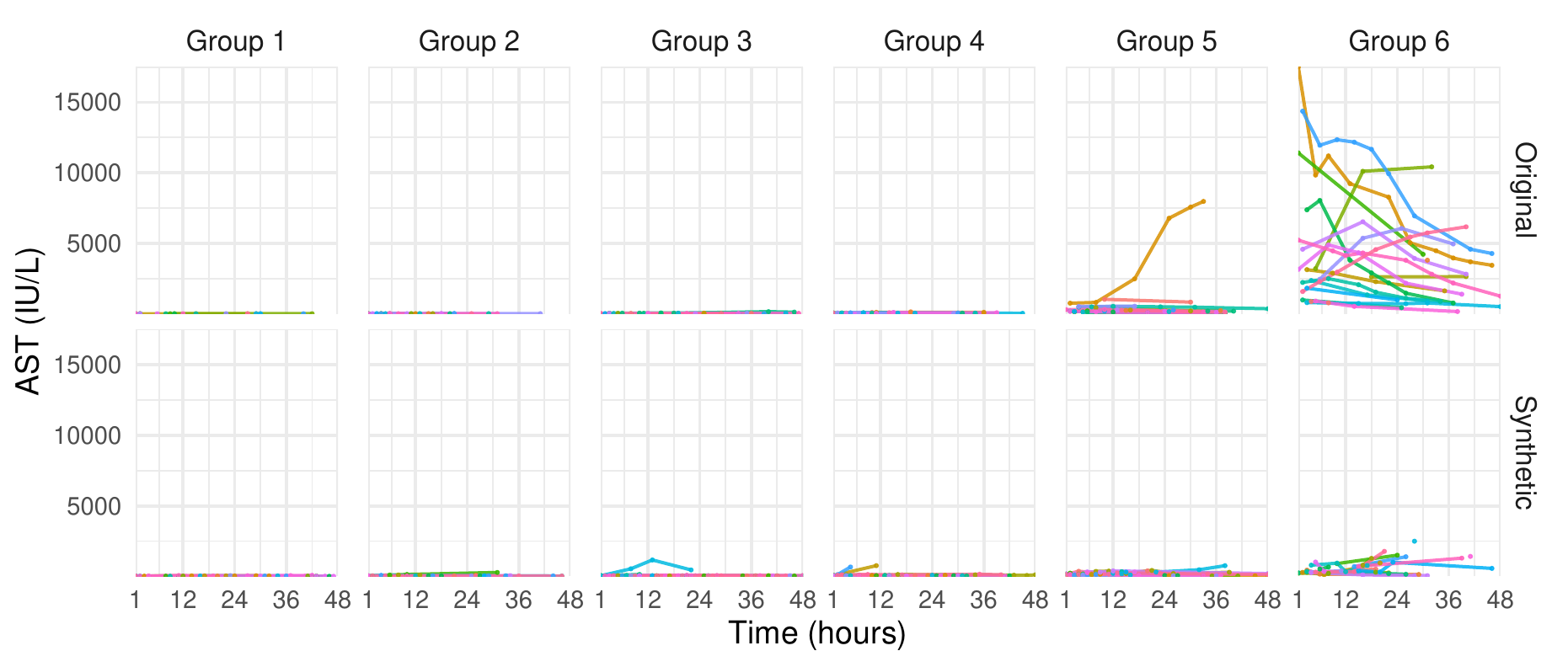}
    \caption{}
    \label{fig:indiv_traj_ast}
    \end{subfigure}

    \begin{subfigure}{0.88\textwidth}
        \centering
    \includegraphics[width=\textwidth]{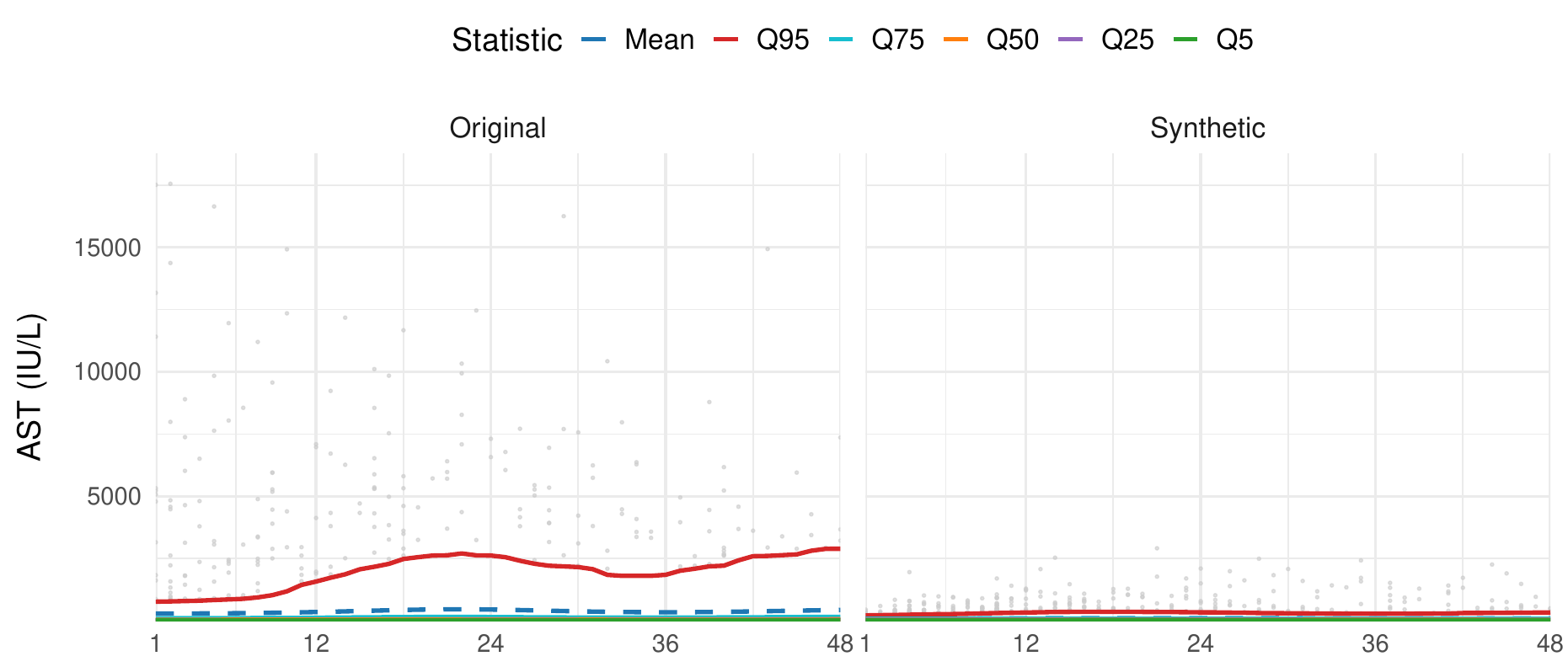}
    \caption{}
    \label{fig:mean_quant_ast_out}
    \end{subfigure}
    \caption{Panel (a) shows subject-specific aspartate aminotrasferase trajectories by baseline group for the original and HGG-generated synthetic data, while panel (b) presents the mean–quantile profiles from Figure~\ref{fig:mean_quant_ast} augmented with observations below the 5th percentile and above the 95th percentile.}
    \label{fig:indiv_outlier_ast}
\end{figure}

The subject-specific trajectories shown in Figure~\ref{fig:indiv_traj_ast} confirm greater variability among individuals above the 95th percentile (group 6) in the original data. Although trajectories are generally stable over time, individuals in this group often exhibit decreasing trends, consistent with the rank-order variability, where the median remains close to zero but the distribution is skewed toward negative values. This pattern is largely absent in the synthetic data.

The mean–quantile plot with outliers (Figure~\ref{fig:mean_quant_ast_out}) further shows that the synthetic data contains fewer extreme values than the original data, in contrast to the behavior observed for the HALO-generated respiratory rate (Figure~\ref{fig:mean_quant_resp_out}). Together with the suppressed 95th percentile, this reduction in outliers explains the deflation of total variance. 

The synthetic data reproduced similar measurement density (Supplemental Figure~\ref{fig:meas_dens_ast}). The mean Frobenius norm between the original and synthetic measurement matrices was 42.66 (reference 24.13), the mean measurement similarity score was 0.98 (reference 0.99), and the mean drop-out divergence was 0.07 (reference 0.01), suggesting minor deviations in terms of measurement structure.

\subsection{Discrepancies between class proportions and transition probabilities} \label{sec:illustr_categ}

Finally, we illustrate the proposed metrics for discrete variables using the \texttt{Glasgow Coma Scale} (GCS) score and \texttt{FiO\textsubscript{2}} generated by HGG, as well as the \texttt{Glasgow Coma Scale Eye Opening} assessment generated by HALO as examples. The results illustrate that synthetic data can exhibit discrepancies in both class proportions and transition probabilities, with rare classes being particularly difficult to reproduce reliably.

\begin{figure}[H]
    \centering 
    \includegraphics[width=\linewidth]{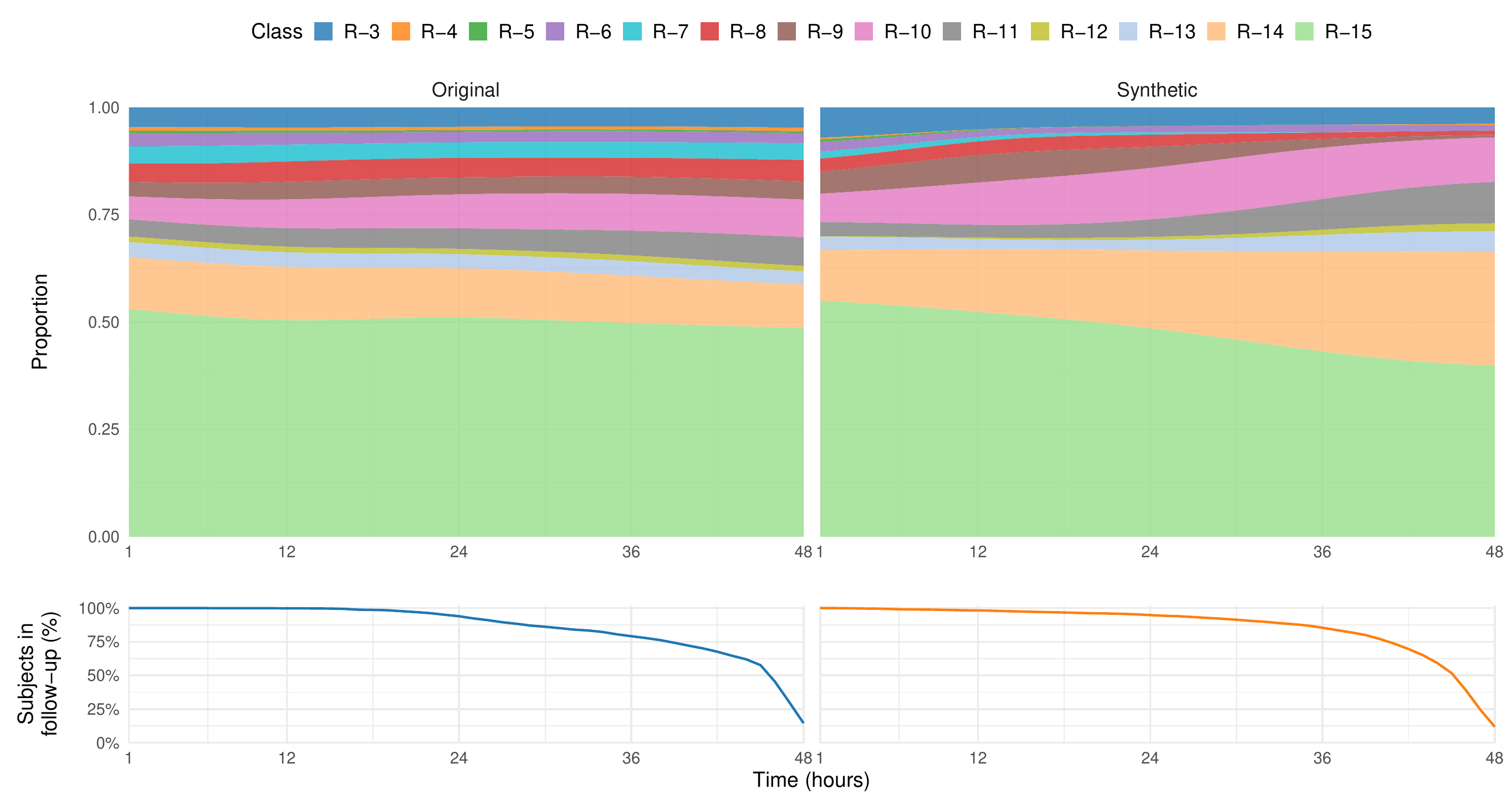} 
    \caption{Kernel-smoothed class profiles (Metric~\ref{metric:class}) of Glasgow Coma Scale score for the original data (left) and HGG-generated synthetic data (right).} 
    \label{fig:GCS_proportion} 
\end{figure}

The class profiles (Metric~\ref{metric:class}) of GCS score generated by HGG shown in Figure~\ref{fig:GCS_proportion} are largely preserved despite the presence of 13 distinct classes. Deviations are more pronounced for less frequent classes in the synthetic data, which reflects their limited representation in the training set.

\begin{figure}[t!]  
\centering 
\begin{subfigure}{0.9\textwidth} 
\centering \includegraphics[width=\linewidth]{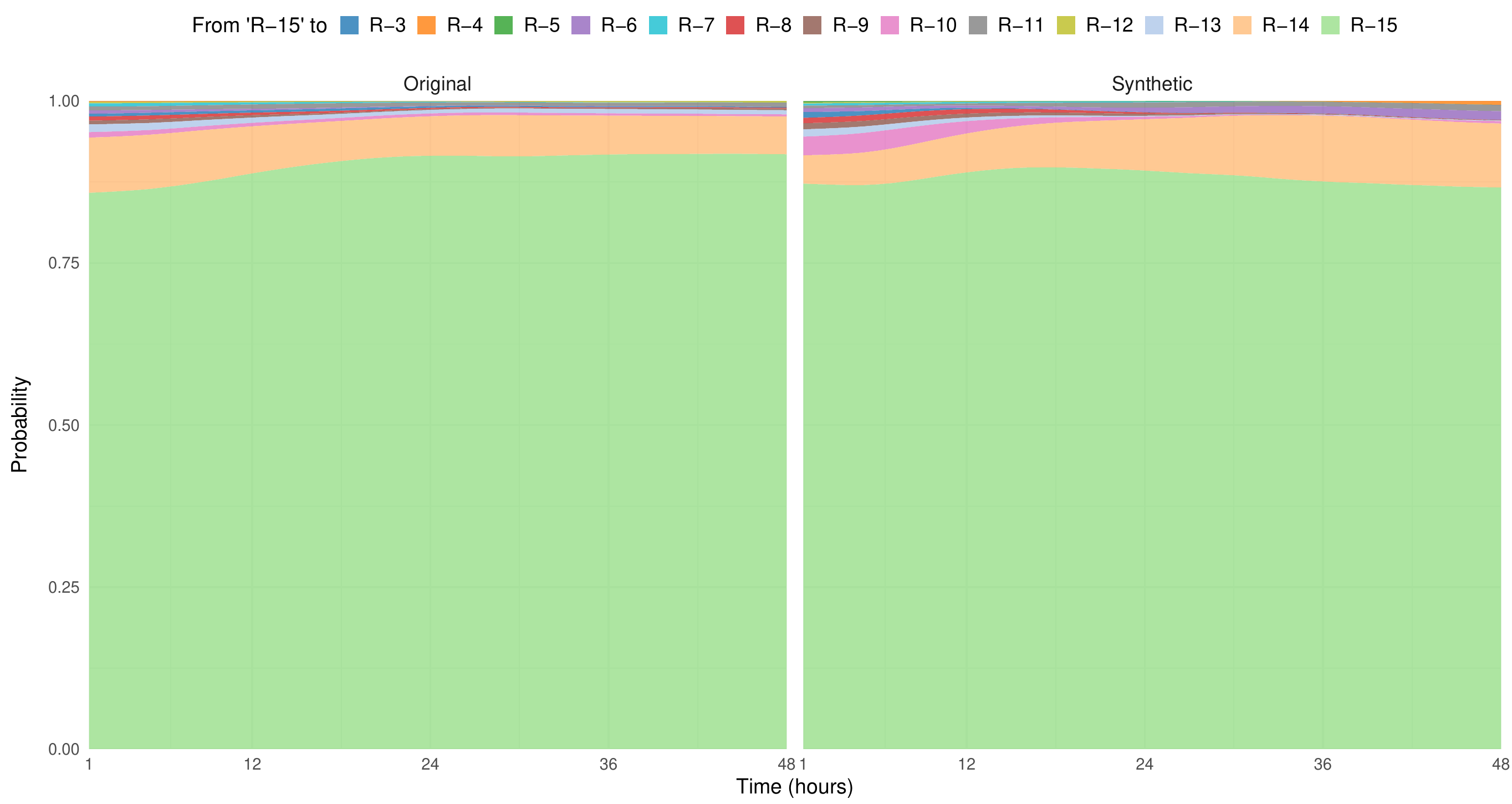} \caption{} 
\label{fig:GCS_R_15} 
\end{subfigure} \vspace{4pt} 
\begin{subfigure}{0.9\textwidth} 
\centering 
\includegraphics[width=\linewidth]{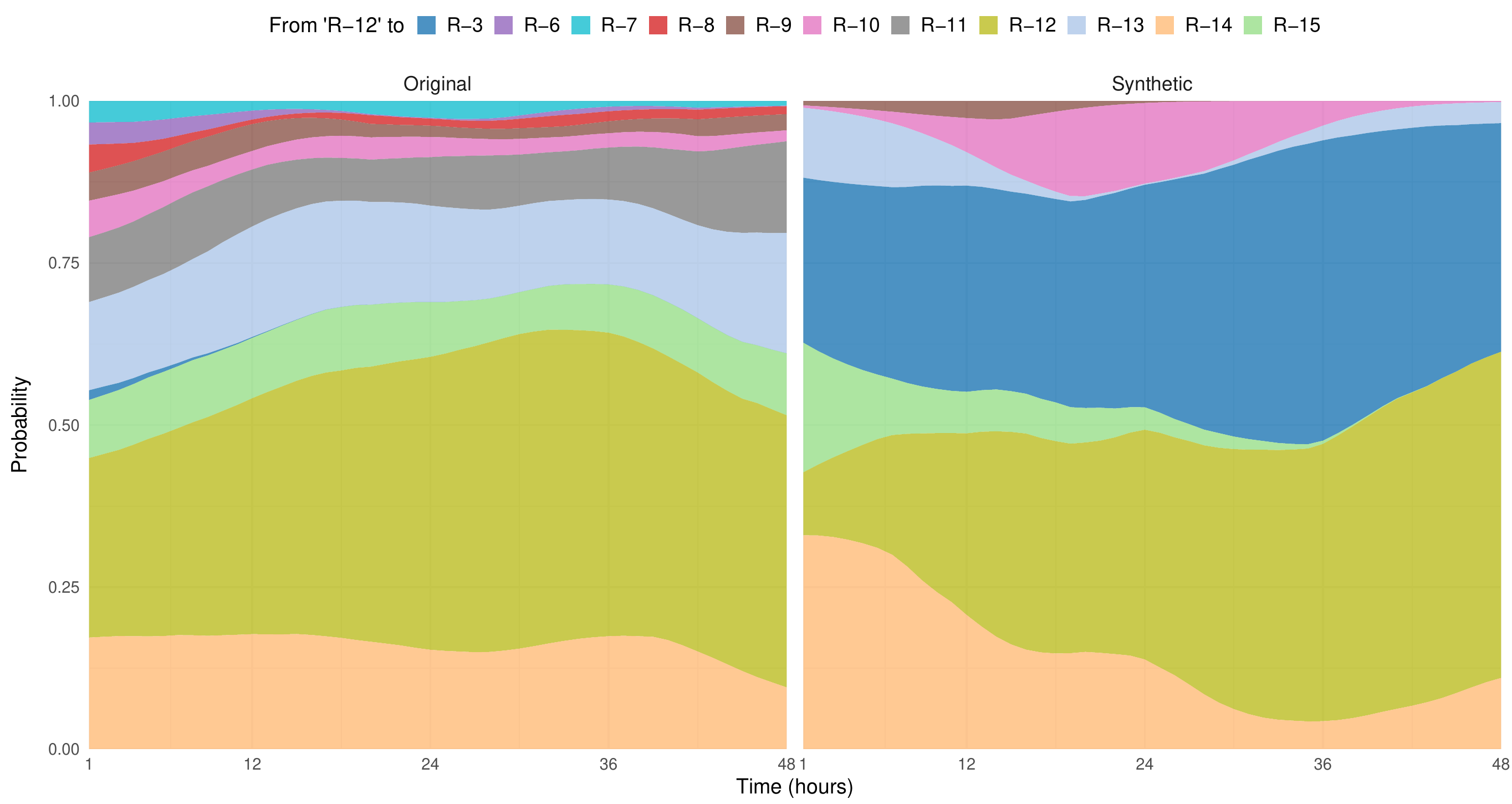} \caption{} \label{fig:GCS_R_12}
\end{subfigure}
\caption{Panel (a) displays the kernel-smoothed transition probabilities (Metric~\ref{metric:trans}) of the Glasgow Coma Scale score from class R-15, while panel (b) shows the transition probabilities from class R-12, for the original data (left) and the HGG-generated synthetic data (right).}
\label{fig:GCS_overview} \end{figure}

Transition probabilities (Metric~\ref{metric:trans}) are reasonably well captured for the majority class ``R-15'' (Figure~\ref{fig:GCS_R_15}), whereas the probabilities of minority classes are not learned reliably (Figure~\ref{fig:GCS_R_12}). This behavior was consistent across all discrete variables (see additional examples in Supplemental Figure~\ref{fig:GCS_R_3}). HALO exhibited similar limitations and showed greater difficulty in preserving overall class proportions (Supplemental Figure~\ref{fig:HALO_GCS}).

\begin{figure}[!tp]
    \centering
    \begin{subfigure}{0.89\linewidth}
    \includegraphics[width=\textwidth]{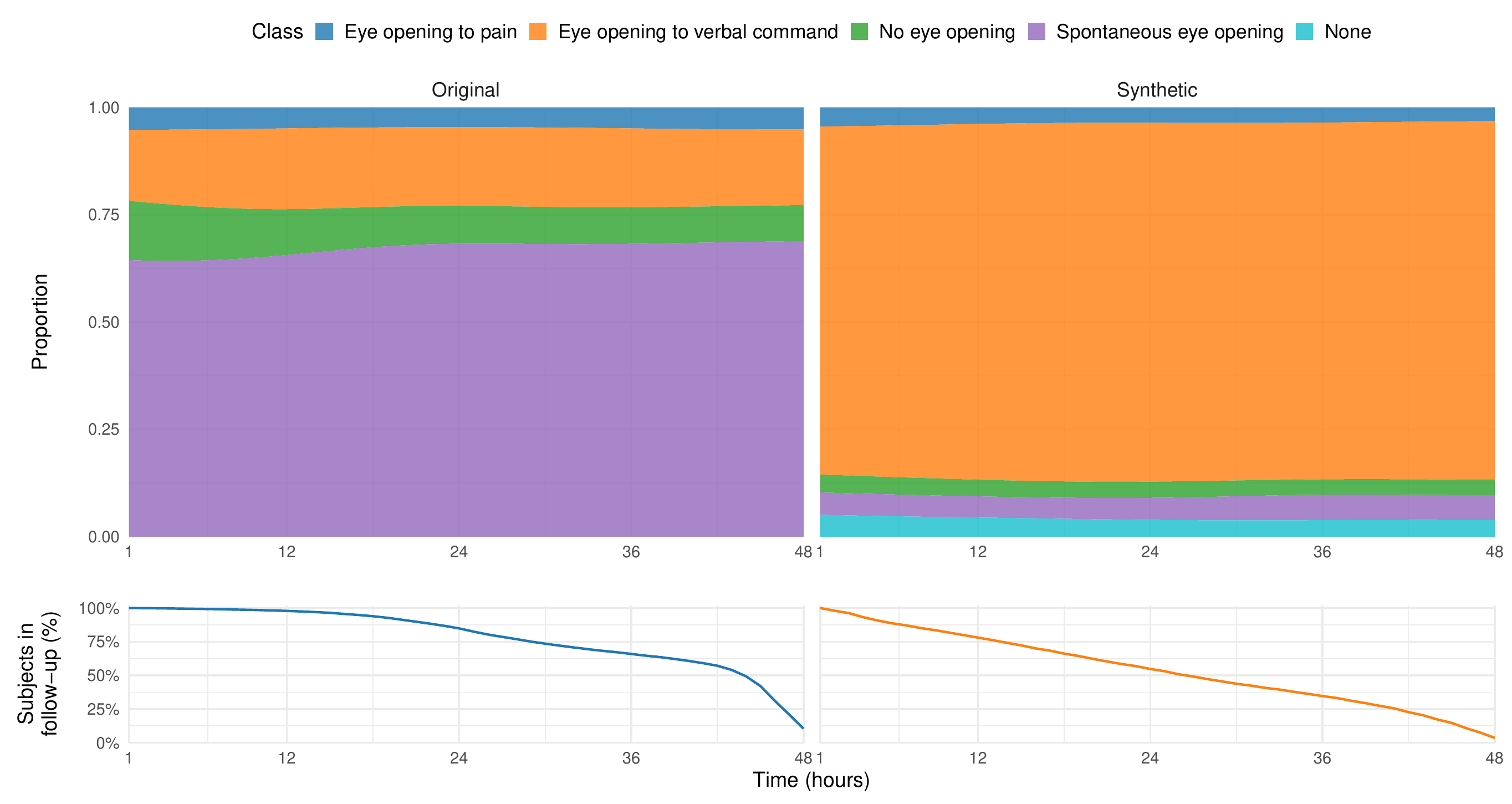}
    \caption{}
    \label{fig:props_eye_open}
    \end{subfigure}
    
    \begin{subfigure}{0.89\linewidth}
    \includegraphics[width=\textwidth]{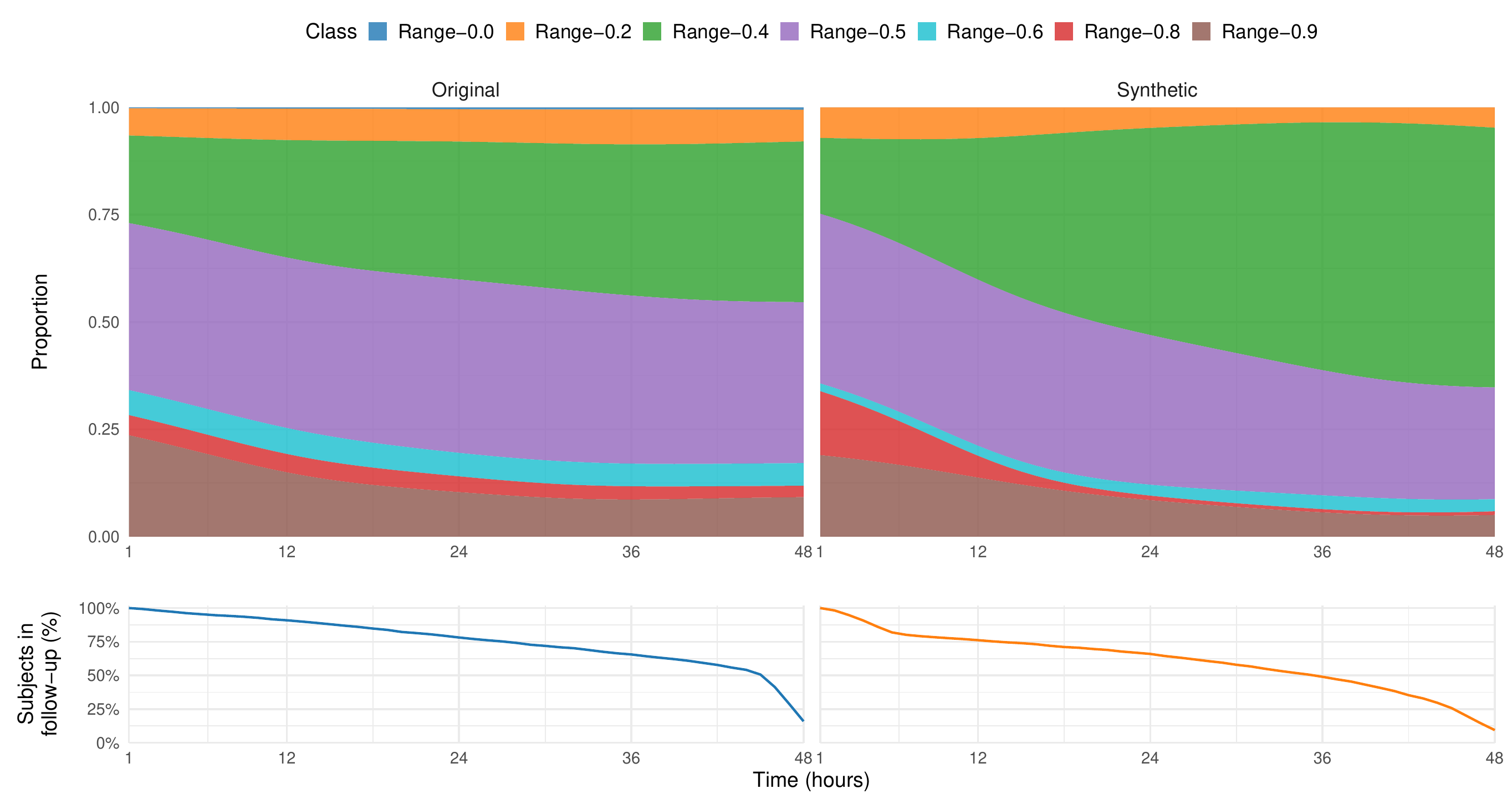}
    \caption{}
    \label{fig:FiO2_proportion}
    \end{subfigure}
    \caption{Kernel-smoothed class profiles of (a) the Glasgow Coma Scale Eye Opening assessment generated by HALO and (b) FiO\textsubscript{2} generated by HGG, shown for the original data and the synthetic data.}
\end{figure}

The mean Frobenius norm of HHG-generated GCS score between the original and synthetic measurement matrices was 121.64 (reference 62.82), the mean measurement similarity score was 0.85 (reference 0.96), and the mean drop-out divergence was 0.12 (reference 0.02). These results indicate moderate discrepancy in the measurement structure. 

 More pronounced discrepancies were observed in the treatment of rare categories. The GCS Eye Opening assessment generated by HALO included an additional class, ``none,'' that was absent from the training data (Figure~\ref{fig:props_eye_open}, right panel), whereas the FiO\textsubscript{2} variable generated by HGG failed to reproduce the class ``Range-0.0'' (Figure~\ref{fig:FiO2_proportion}, right panel).

\section{Discussion} \label{sec:discuss}

We introduced a comprehensive set of metrics designed for evaluating temporal preservation in synthetic longitudinal patient data. While recent advances in generative modeling enable increasingly realistic synthetic datasets, existing evaluation frameworks largely focus on cross-sectional similarity and overlook temporal dependencies essential to LPD. Our proposed metrics address this gap by providing tools to quantify how well synthetic data reproduces key longitudinal characteristics, including marginal, covariance and individual structures as well as measurement patterns. 

Our illustrations reveal that even when marginal summaries such as means or quantile profiles appear consistent with the original data, the covariance structure may be distorted or lost. For example, the HALO-generated weight variable appeared well aligned at the marginal level, yet the variogram analysis and subject-specific trajectories revealed disrupted covariance and irregular individual patterns. These findings highlight that marginal-level agreement alone can be misleading and underscore the need for multiple complementary metrics to comprehensively assess synthetic data quality. Evaluating results jointly across multiple domains offers a more complete picture of temporal preservation and helps identify why certain aspects of the underlying temporal structure may fail. An integrated approach is essential for understanding both how well a given SDG method retains the original longitudinal characteristics and where improvements are needed.

Applying a full range of metrics revealed several factors that critically influence the temporal fidelity of synthetic LPD. A key determinant of achieving high quality synthetic data is the quality of the original data itself, which highlights the familiar ``garbage in, garbage out'' principle. Several of the issues observed can be traced to how variables were represented and preprocessed for SDG. Outliers and implausible values, such as negative measurement times present in the original datasets, were reproduced either with exaggerated or diminished magnitude in synthetic data, thereby distorting the covariance structure. These results suggest that systematic data cleaning is essential before applying SDG methods, unless the downstream tasks specifically depend on extreme or atypical values. In most longitudinal applications, however, the primary goal is to model overall disease behavior or predict patient outcomes, rendering such outliers rather noise than signal.

Beyond data quality, modeling choices and preprocessing decisions further shape temporal preservation. Many existing SDG methods, including HALO, require discretizing continuous variables and reconstruct them by sampling uniformly within predefined bins. However, if the bins are excessively wide or if binning produces very small classes, these choices can introduce artifacts and amplify data imbalance. For instance, in our examples HALO generated outliers from an overly wide uniformly distributed bin (Figure~\ref{fig:mean_quant_resp_out}). 
Severe class imbalance may also lead to mode collapse \cite{Kuo2023GeneratingHIV}, whereby dominant classes are reproduced while the smallest minority classes fail to be generated, as observed in Figure~\ref{fig:FiO2_proportion}, and may further limit the ability of the methods to learn reliable transition probabilities (Figure~\ref{fig:GCS_R_12}). More adaptive and data-driven preprocessing strategies, such as density-based binning \cite{Aristodimou2022ADomain}, could mitigate these issues by adapting bin boundaries to the empirical data distribution, thus ensuring a more balanced representation of rare values while preserving important local variations.

Another major determinant of model performance was measurement density. Variables measured less frequently or inconsistently likely provided insufficient information for the SDG methods to learn the underlying distribution. This illustrates a key challenge in unbalanced LPD: while temporal patterns can be learned well for frequently measured variables, such as systolic blood pressure depicted in Figure~\ref{fig:mean_quant_syst}, those with sparse or irregular sampling (e.g., aspartate aminotransferase; Figure~~\ref{fig:mean_quant_ast_out}, observed at only 1.5\% of time points) or small class representation remain difficult to model accurately. Uneven measurement density and irregular measurement schedules can therefore lead to variable-specific discrepancies in temporal preservation, highlighting the need for SDG methods that explicitly account for unbalancedness in LPD. 

In addition to differences in measurement density across variables, discrepancies also arose from how temporal information itself was represented and generated. In the original HALO training data, 99.7\% of measurements occurred at full-hour intervals (integers), whereas in the corresponding synthetic data this proportion dropped to 13.1\%. Such discrepancies may arise when the method generates time as real-valued (floats) even though the underlying distribution is predominantly integer-based, or when severe unbalancedness in the training data biases the learning process. When these real-valued synthetic times are transformed back into the original scale, as was done in our illustrations, overlapping or alternating time points can appear, producing the zigzag pattern observed in the measurement density plots. This issue is less pronounced in HGG because the model uses a fixed 48-step time structure. These findings emphasize that preserving temporal structure in LPD requires careful alignment between how time is encoded and modeled.

Finally, model-specific assumptions introduce additional challenges. In language model–based generators such as HALO, defining a fixed vocabulary over the entire dataset can lead to artifacts when certain classes are so rare that they never appear in the training data; these unseen classes remain valid tokens and may still be generated, despite the model lacking evidence to calibrate their frequency or transitions (Figure~\ref{fig:props_eye_open}). A potential remedy is to adopt dynamic vocabulary construction \cite{Wu2018NeuralVocabularies}. In contrast, GAN-based architectures such as HGG are sensitive to missing data handling. Forward-fill or mean imputation combined with sparse observations may cause training trajectories to be dominated by imputed values, producing overly smooth patterns with reduced variability in the synthetic data, as observed for aspartate aminotransferase (Figure~\ref{fig:mean_quant_ast}). More advanced imputation strategies \cite{Huque2018AStudies} could better preserve the original data distribution and variability.

\subsection{Limitations} \label{sec:lim}

The present study has several limitations. First, it is important to note that our goal was not to compare specific SDG methods but to demonstrate the utility of the proposed resemblance metrics. Although we initially considered comparing different SDG approaches for generating LPD, a fair comparison proved infeasible because each method requires distinct preprocessing pipelines, preventing the use of a common training set for benchmarking. Furthermore, the continuous-variable version of HALO used here was only briefly described in the original publication \cite{Theodorou2023SynthesizeModel} and may still be under development, which could account for the coarse binning and limited data cleaning resulting to the outlier issues. For both HALO and HGG, only a single generator was trained, whereas a thorough performance assessment would have required multiple training runs across different datasets and hyperparameter settings, an approach beyond the scope of this study.

Second, we did not include metrics assessing logical consistency, such as monotonicity, where certain values are expected to change only in one direction over time, or non-recurrence, where specific events are expected to occur only once. This was primarily because our data did not contain variables for which such metrics could have been meaningfully illustrated. Nevertheless, these types of constraints represent important aspects of realistic temporal behavior and should be considered when assessing the plausibility of synthetic LPD.

Third, our evaluation focused exclusively on univariate resemblance metrics. Multivariate comparisons, while relevant for understanding relationships between variables, are often difficult to interpret and visualize in a temporal context. Moreover, we argue that if univariate resemblance metrics already reveal substantial deviations, assessing multivariate preservation may add limited additional insight. Developing interpretable and informative multivariate temporal preservation metrics remains an important direction for future research.

Fourth, we did not distinguish between true missingness and structural unbalancedness. Future research could build on established methods for characterizing and handling missing data to better disentangle the two. Descriptive diagnostics such as visualization of missing data patterns or Little’s MCAR test \cite{Little1988AValues,Li2013LittlesRandom} could help assess whether missingness is plausibly random, while logistic regression or random forest models predicting the probability of missingness from observed variables could aid in distinguishing missing at random from missing not at random. Extending and adapting these approaches to secondary real-world patient data, where the boundaries between missingness and unbalancedness are often blurred, represents a promising direction for methodological development.

Finally, we did not assess privacy protection or the analytical utility of synthetic data. While our metrics quantify structural differences between synthetic and original datasets, such differences may not directly correspond to performance gaps in downstream tasks such as prediction or classification. However, from a privacy perspective, larger deviations from the original data may indicate stronger protection against re-identification or attribute disclosure. Conversely, parametric modeling and statistical inference are often more sensitive to distributional deviations, which may have a greater impact on analytical utility. Future work should therefore complement structural resemblance assessments with joint evaluation of privacy and utility to better characterize the trade-offs between data protection and performance.

\subsection{Conclusion}

In this study, we introduced a comprehensive set of resemblance metrics for evaluating univariate temporal preservation in synthetic longitudinal patient data, addressing a key limitation of existing evaluation frameworks that largely focus on cross-sectional similarity. Through illustrative examples, we showed that agreement at the marginal level alone can be misleading as synthetic data may exhibit substantial distortions in covariance structure, individual structure, and measurement patterns. Our results indicate that temporal fidelity is strongly influenced by data quality, preprocessing choices, and modeling assumptions, with sparse measurements, imbalanced class distributions, and differences between generated time values and the original measurement schedule posing particular challenges across different generative paradigms. The proposed metrics provide a structured and interpretable framework for diagnosing these issues, offering insight into where and why temporal structure is preserved or lost. Together, these findings underscore the importance of rigorous temporal evaluation as a prerequisite for the responsible use of synthetic longitudinal patient data and motivate future work integrating temporal resemblance with assessments of multivariate structure, privacy protection and analytical utility.

\section{Acknowledgments}

\noindent We gratefully acknowledge Auria Clinical Informatics and Arho Virkki from Turku University Hospital for providing the computational environment that enabled this research.

\section{Author contributions}

\noindent \textbf{Katariina Perkonoja}: Conceptualization, Methodology, Formal analysis, Investigation, Resources, Data curation, Writing – original draft, Visualization, Project administration, Funding acquisition.
\textbf{Parisa Movahedi}: Conceptualization, Data curation, Formal analysis, Visualization, Writing – original draft.
\textbf{Antti Airola}: Conceptualization, Writing – review \& editing, Supervision, Funding acquisition.
\textbf{Kari Auranen}: Conceptualization, Methodology, Supervision, Writing – review \& editing. \textbf{Joni Virta}: Conceptualization, Methodology, Investigation, Writing – review \& editing, Supervision, Funding acquisition.

\section{Conflicts of interest}

\noindent The authors declared no potential conflicts of interest with respect to the research, authorship, and / or publication of this article.

\section{Funding}

\noindent This work has received funding from European Union’s Horizon Europe research and innovation programme (grant number 101095384). Views and opinions expressed are those of the authors only and do not necessarily reflect those of the European Union or the European Health and Digital Executive Agency (HADEA). Neither the European Union nor the granting authority can be held responsible for them. The work of KP, JV and AA was supported by the Research Council of Finland (grants 347501, 353769, 358868, 368494). KP also received support from the Finnish Cultural Foundation (grant 00260122). The funding agencies had no involvement in the study.

\section{Use of generative AI}

\noindent In the preparation of this work, the authors used ChatGPT (OpenAI, Version GPT-5) to assist in improving the English language and conciseness of the text. After using this tool, the authors reviewed and edited the content as needed and take full responsibility for the content of the final manuscript.

\section{Data availability}

\noindent The data used in this study are derived from the Medical Information Mart for Intensive Care (MIMIC-III) database, which is publicly available through the PhysioNet repository after completion of the required credentialed data use process. 

The data-extraction scripts and Health Gym GAN source code are available at \url{https://github.com/Nic5472K/ScientificData2021_HealthGym}, while the HALO implementation is provided at \url{https://github.com/btheodorou99/HALO_Inpatient}.

The source code accompanying this work, available at \url{https://github.com/Katsk1/Temporal-preservation-metrics}, provides the implementation of the proposed evaluation metrics and the analysis pipeline used in this study.

\appendix

\setcounter{figure}{0}
\renewcommand{\thefigure}{S\arabic{figure}}

\setcounter{table}{0}
\renewcommand{\thetable}{S\arabic{table}}

\section{Supplementary Figures}

\begin{figure}[H]
\includegraphics[width=\textwidth]{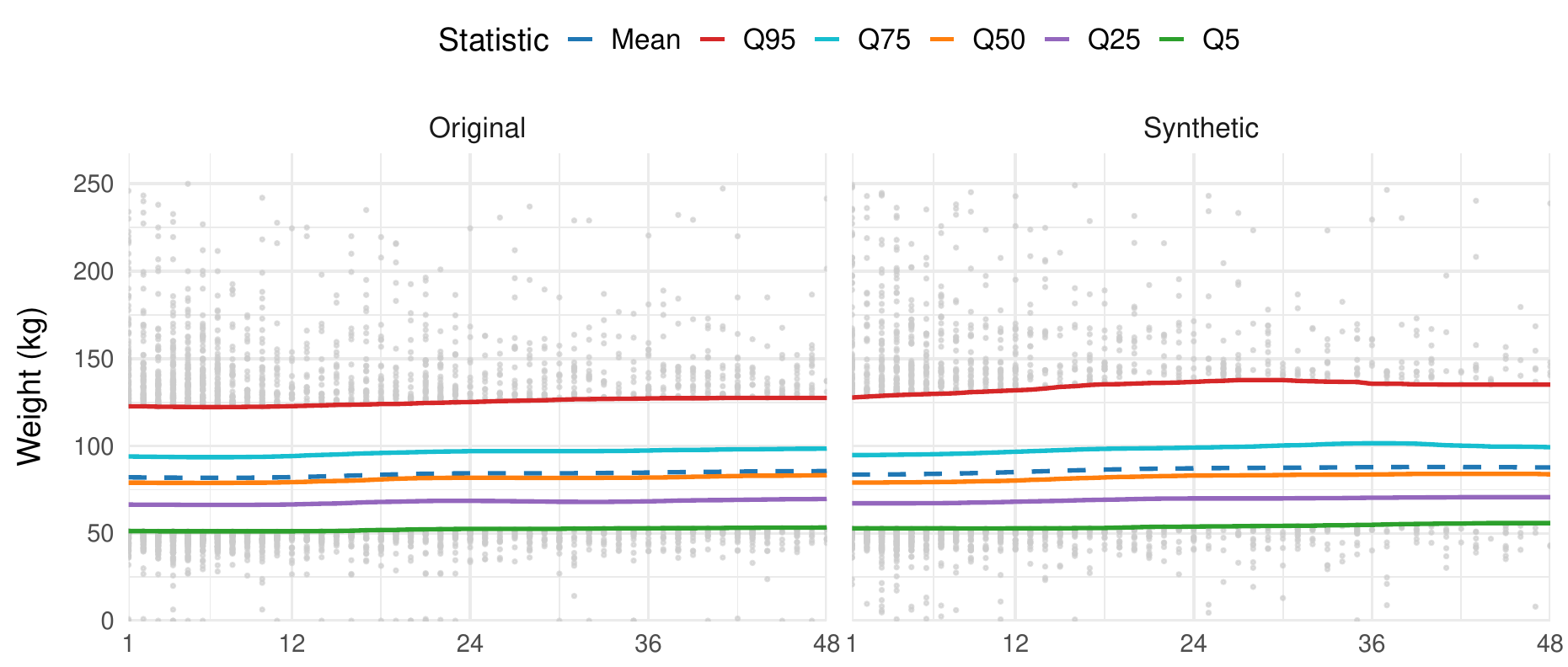}
\caption{The mean–quantile profiles of HALO-generated weight same as in Figure~\ref{fig:mean_quant_weight} augmented with observations below the 5th percentile and above the 95th percentile.}
\label{SM:fig_mean_quantiles_weight_out}
\end{figure}

\begin{figure}[H]
    \centering
    \includegraphics[width=0.8\textwidth]{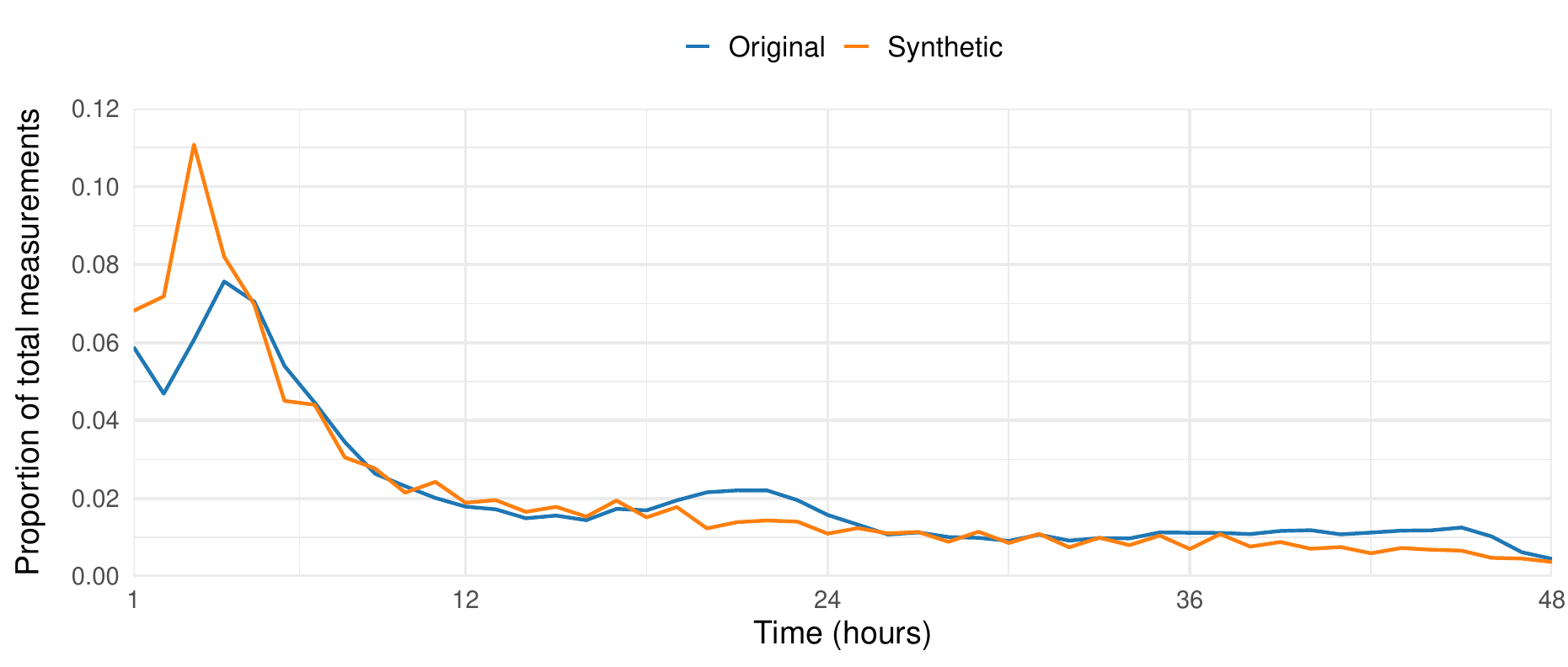}
    \caption{Measurement density of weight for the original data and HALO-generated synthetic data.}
    \label{SM:fig_meas_dens_weight}
\end{figure}

\begin{figure}[H]
    \centering
    \includegraphics[width=\linewidth]{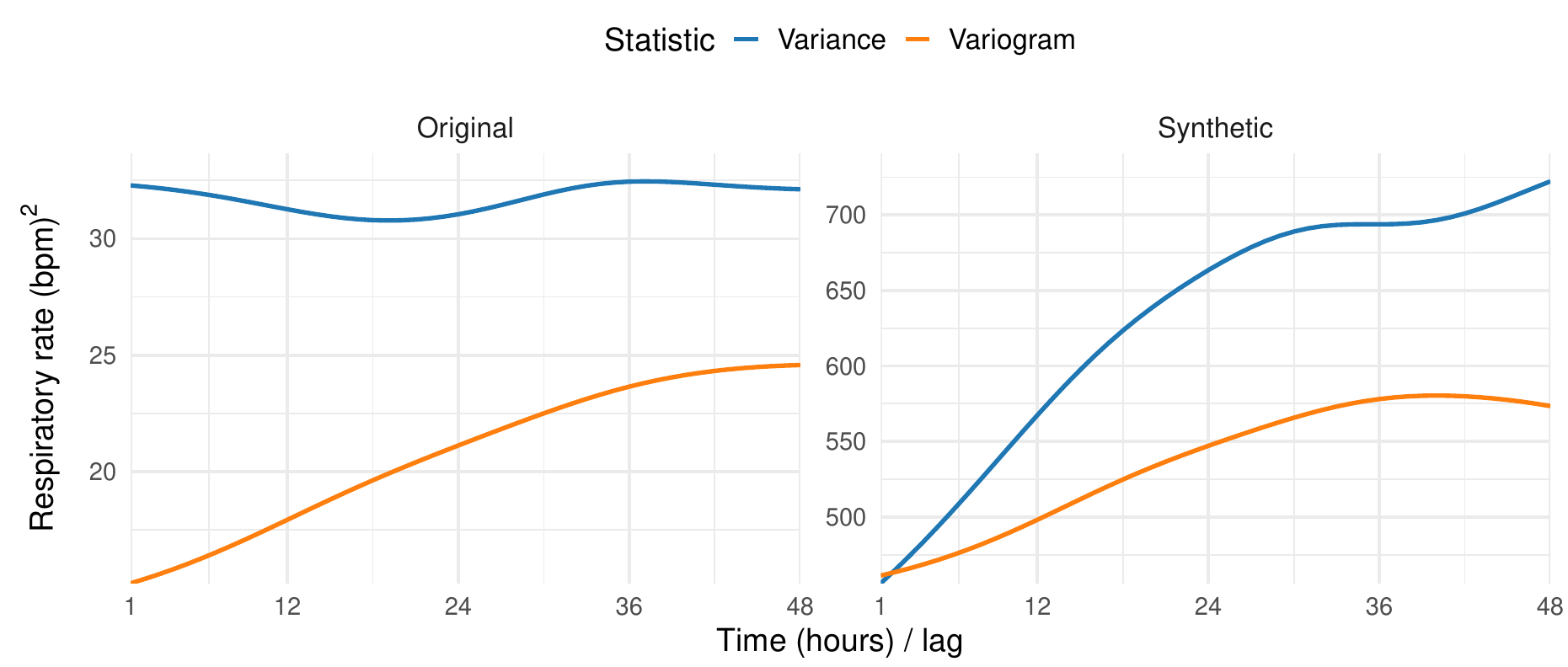}
    \caption{The figure replicates Figure~\ref{fig:variogram_resp} from the main text, but here the original and synthetic data are shown with separate y-axes. While the variogram exhibits an almost identical shape across the two datasets, the variance profile differs in both scale and form compared with the original data.}
    \label{SM:fig_resp}
\end{figure}

\begin{figure}[H]
    \centering
    \includegraphics[width=\linewidth]{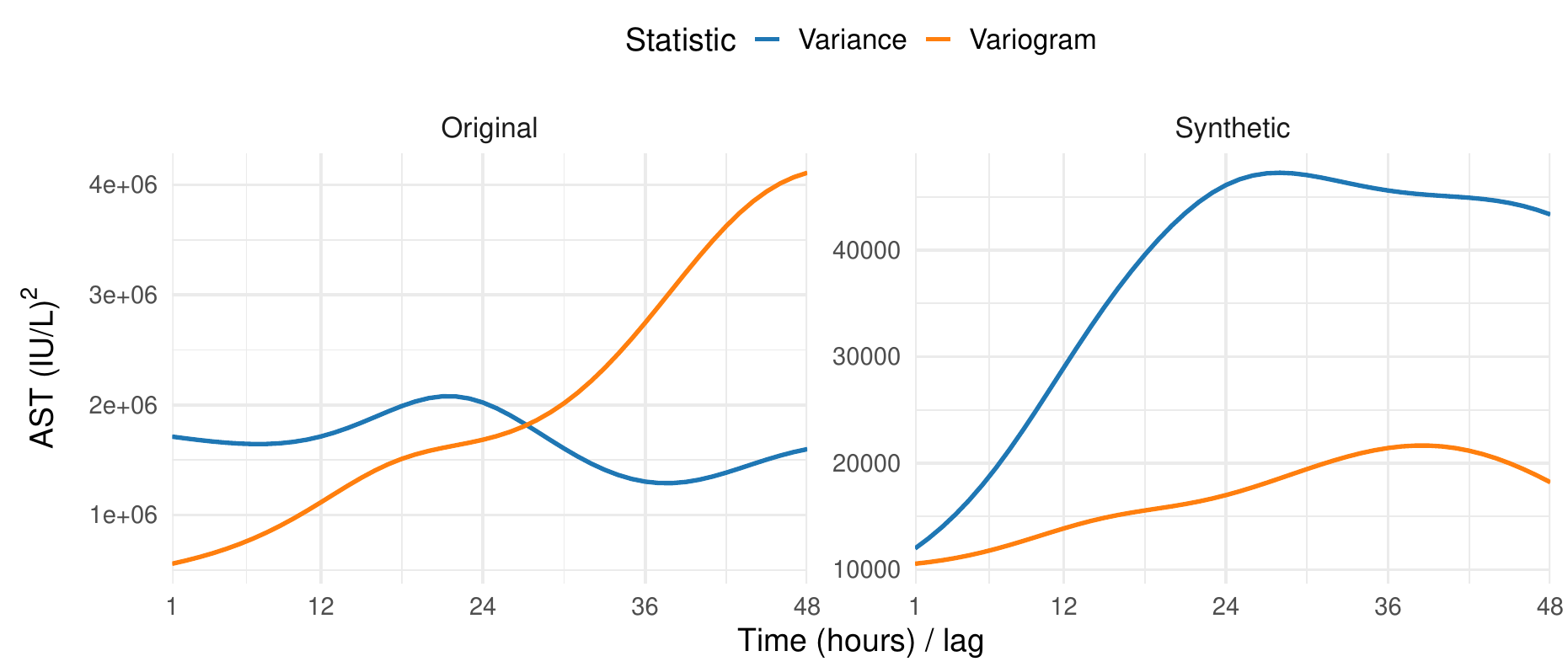}
    \caption{The figure replicates Figure~\ref{fig:variogram_ast} from the main text, but here the original and synthetic data are shown with separate y-axes.}
    \label{SM:fig_ast}
\end{figure}

\begin{figure}[H]
    \centering
    \includegraphics[width=0.9\textwidth]{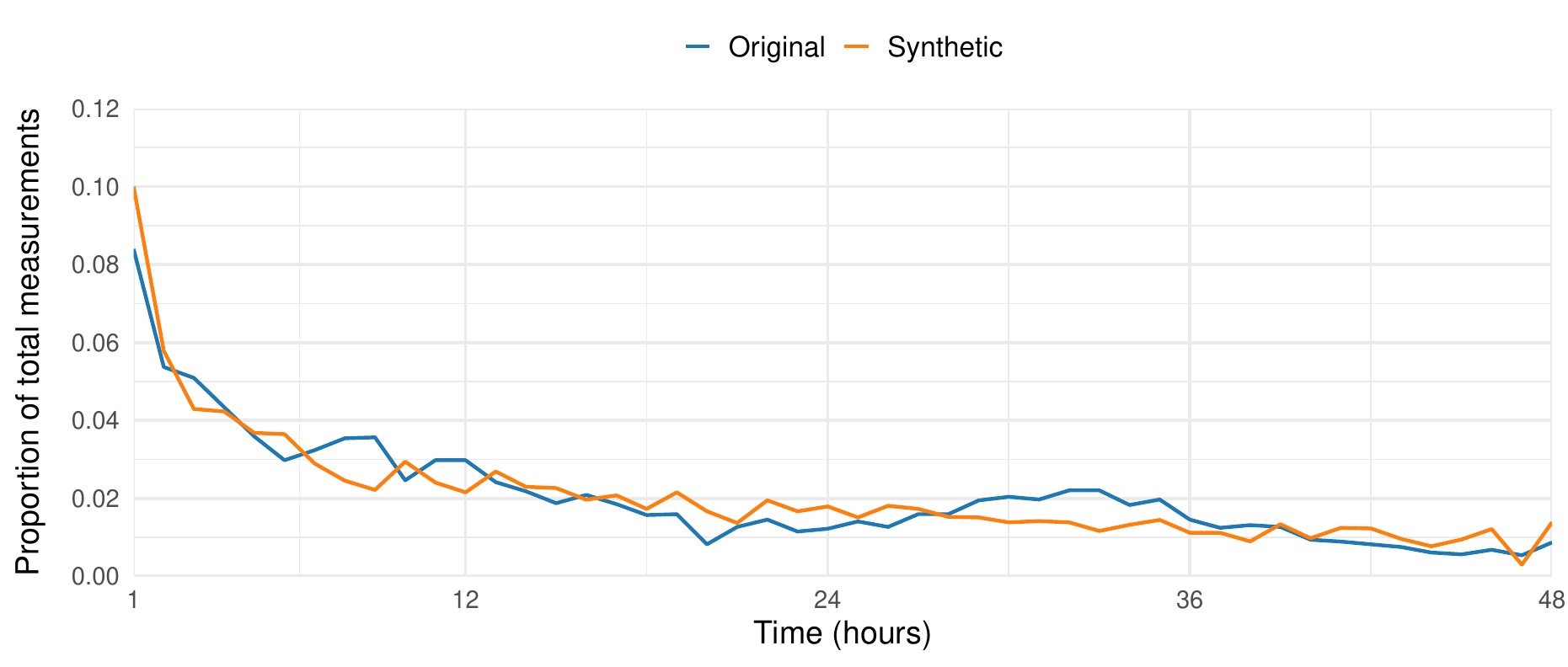}
    \caption{Measurement density of aspartate aminotransferase generated by HGG.}
    \label{fig:meas_dens_ast}
\end{figure}

\begin{figure}[H]
    \centering
\includegraphics[width=0.9\textwidth]{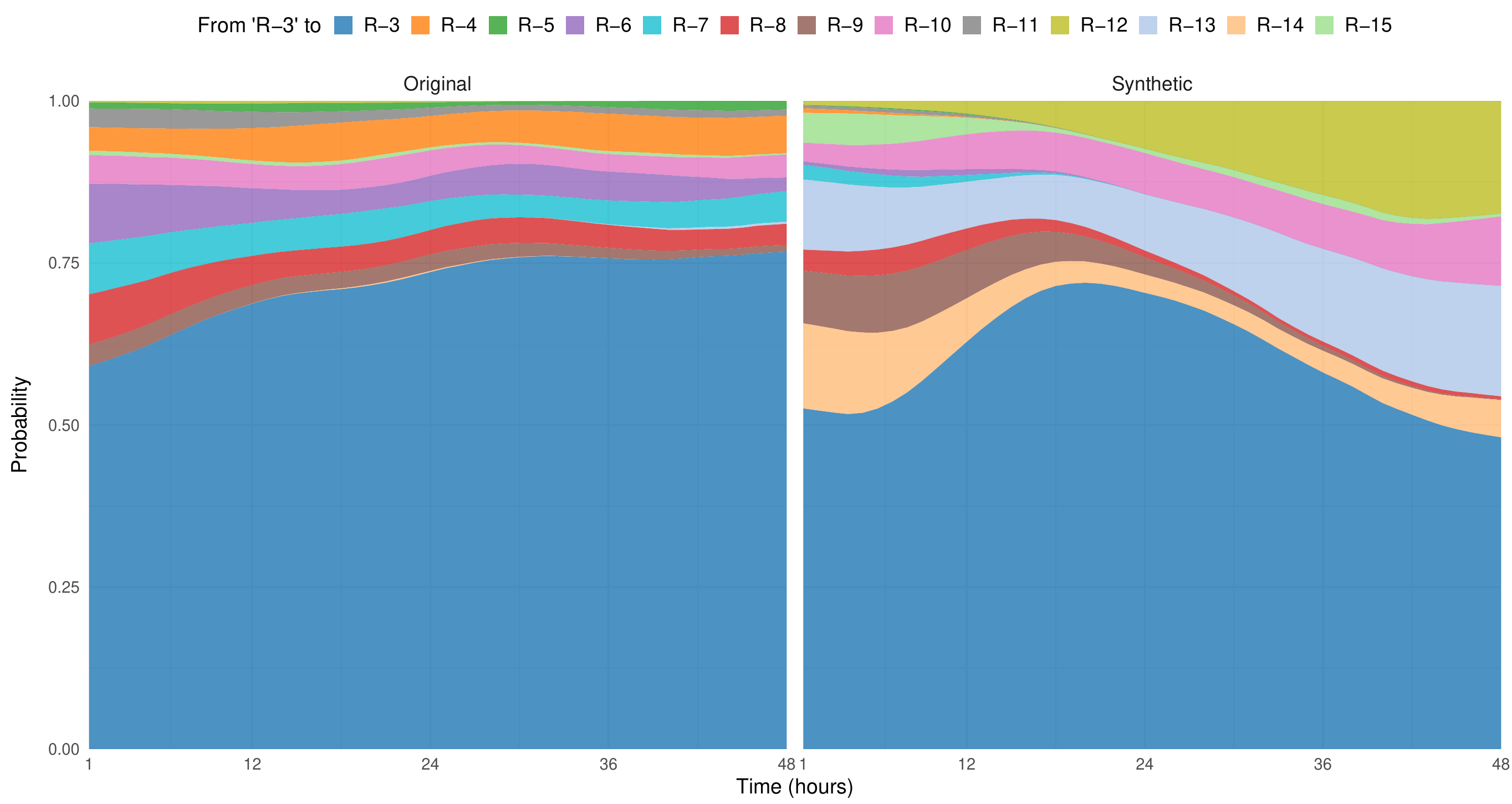}
    \caption{Transition probabilities of class R-3 of Glasgow Coma Scale score generated with HGG.}
    \label{fig:GCS_R_3}
\end{figure}

\begin{figure}[H]  
\centering 
\begin{subfigure}{0.7\textwidth} \centering \includegraphics[width=\linewidth]{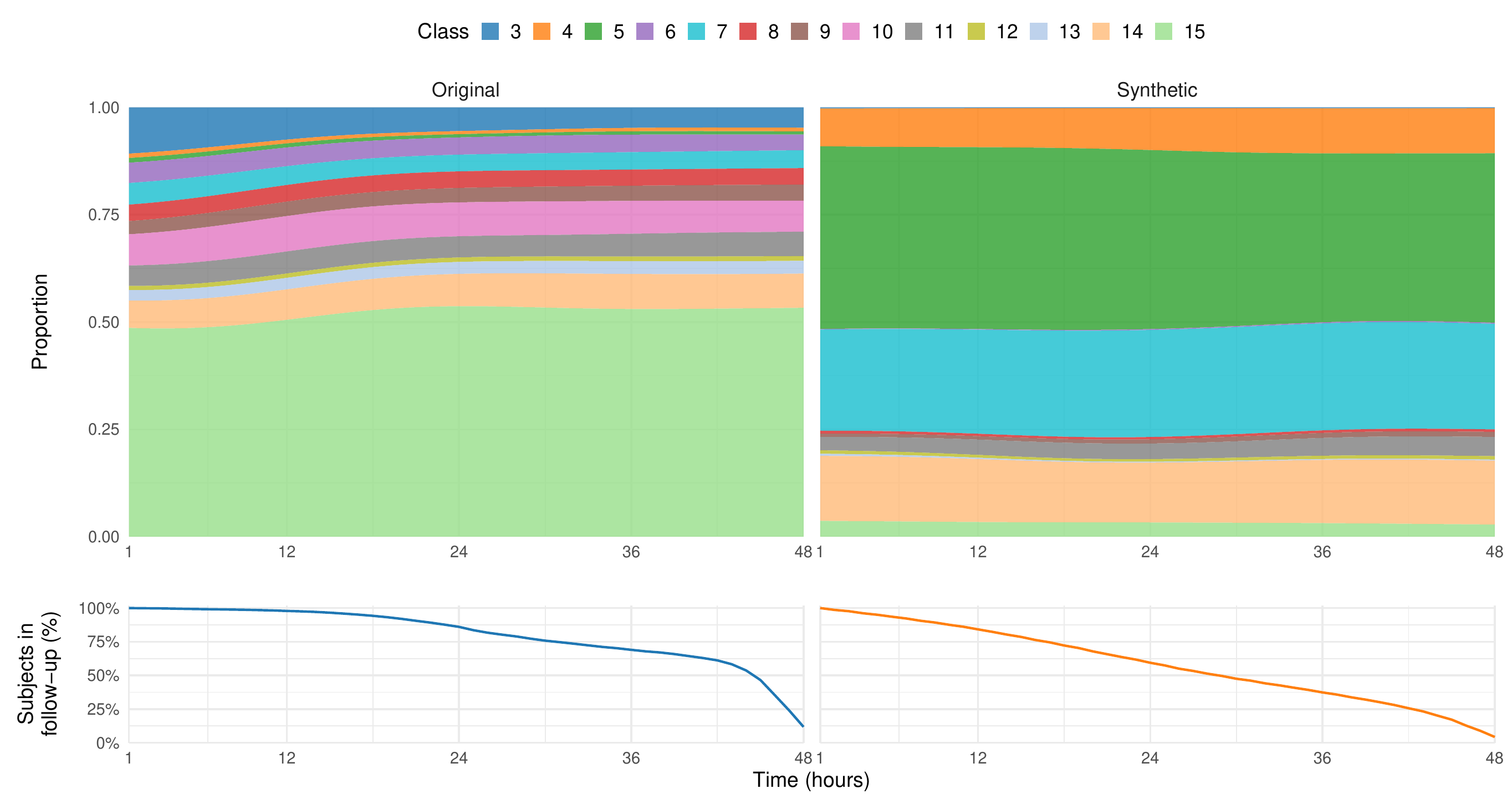} 
\caption{} 
\label{fig:HALO_GCS_prop} 
\end{subfigure} \vspace{4pt} \begin{subfigure}{0.7\textwidth} 
\centering \includegraphics[width=\linewidth]{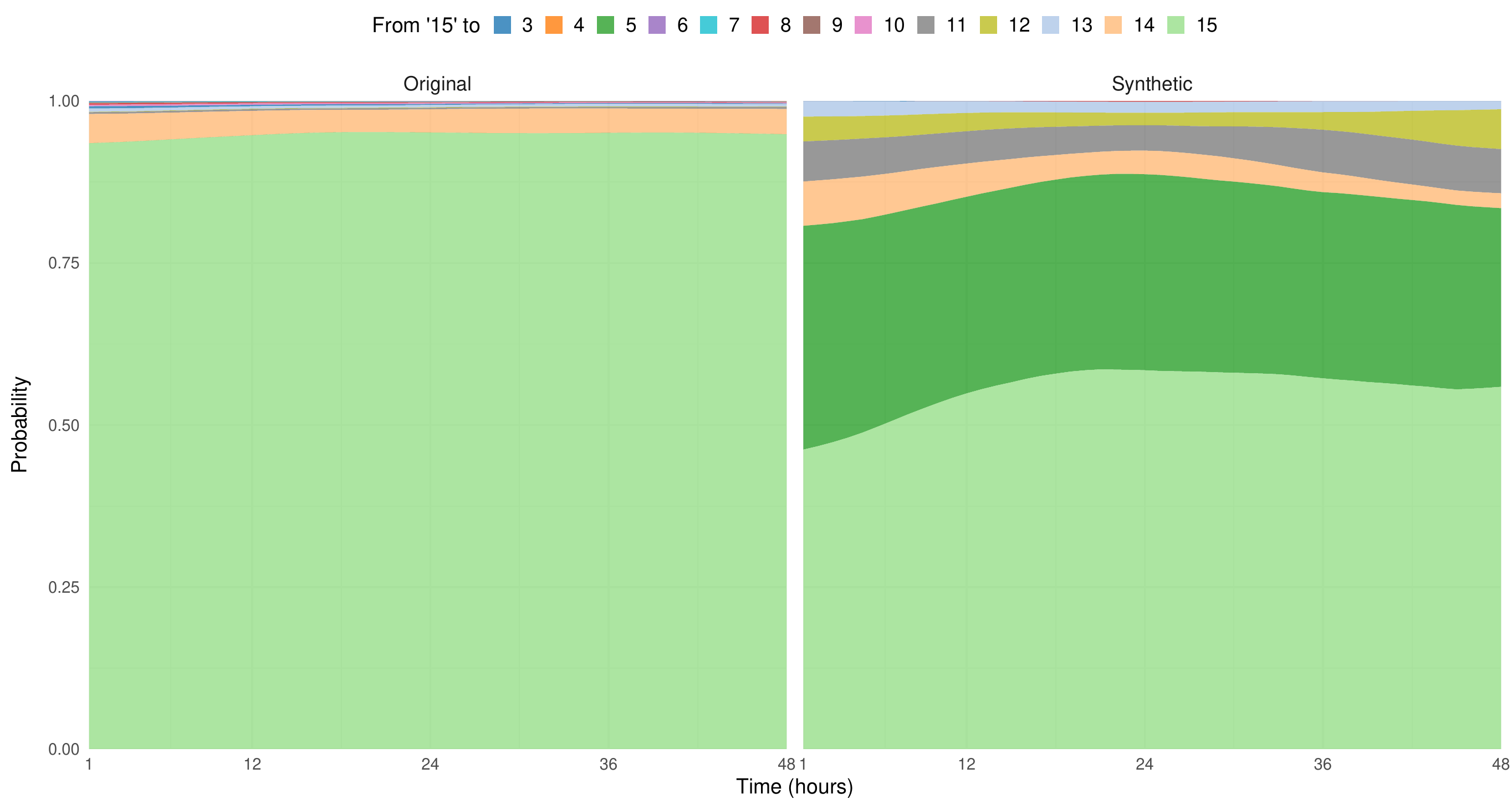} \caption{} 
\label{fig:HALO_GCS_trans15} 
\end{subfigure} \vspace{4pt} \begin{subfigure}{0.7\textwidth} 
\centering \includegraphics[width=\linewidth]{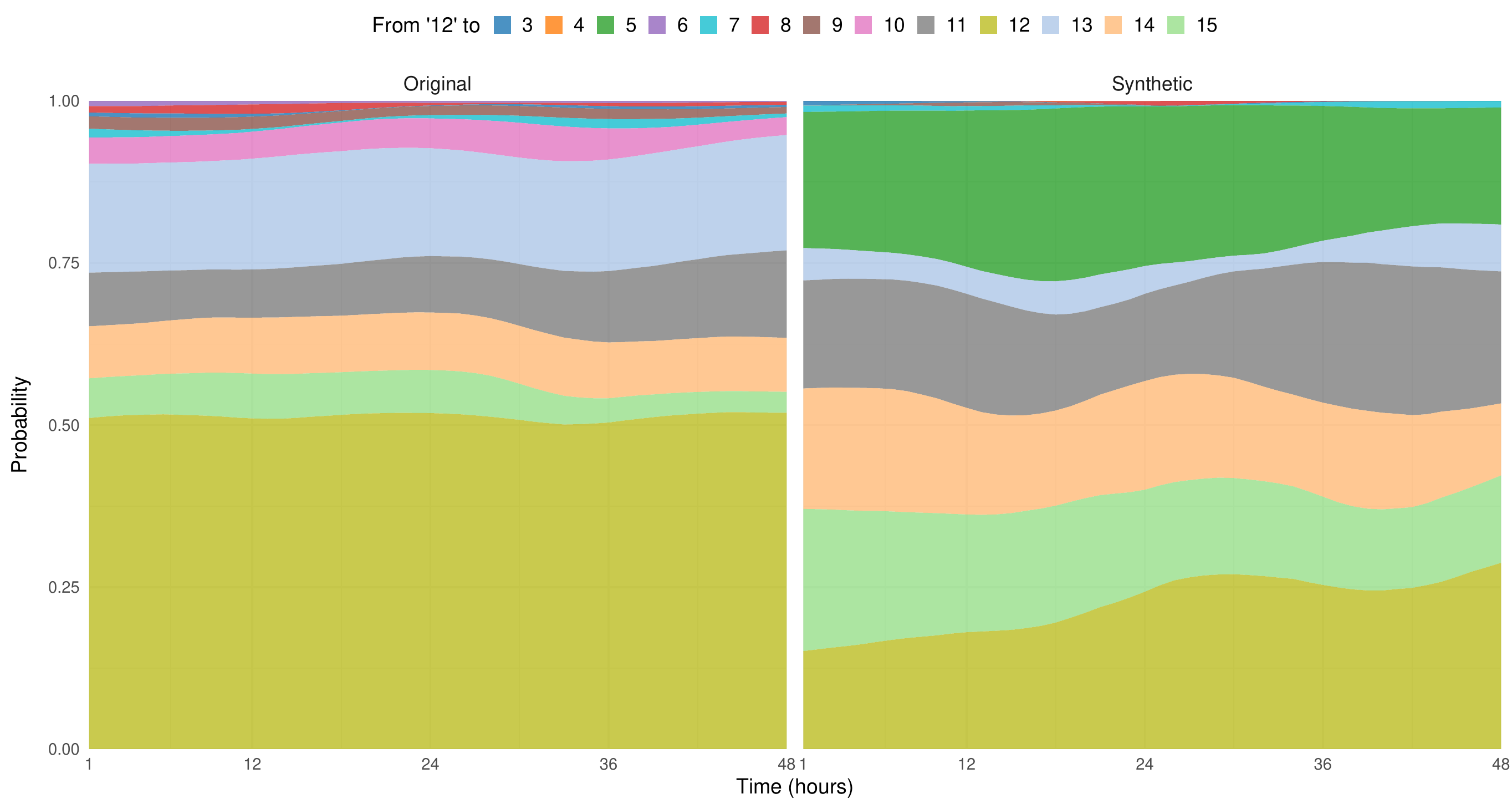} \caption{} 
\label{fig:HALO_GCS_trans12} \end{subfigure} \caption{Glasgow Coma Scale (GCS) scores generated by HALO, showing (a) the kernel-smoothed class profiles, (b) transition probabilities originating from class R-15, and (c) transition probabilities originating from class R-12.} \label{fig:HALO_GCS} \end{figure}

\bibliographystyle{elsarticle-num}
\bibliography{references}

@article{Theodorou2024,
    title = {{ConSequence: Synthesizing Logically Constrained Sequences for Electronic Health Record Generation}},
    year = {2024},
    journal = {Proceedings of the AAAI Conference on Artificial Intelligence},
    author = {Theodorou, Brandon and Jain, Shrusti and Xiao, Cao and Sun, Jimeng},
    number = {14},
    pages = {15355--15363},
    volume = {38},
    doi = {10.1609/aaai.v38i14.29460},
    issn = {2374-3468}
}

@article{Zhang2021,
    title = {{SynTEG: a framework for temporal structured electronic health data simulation}},
    year = {2021},
    journal = {Journal of the American Medical Informatics Association},
    author = {Zhang, Ziqi and Yan, Chao and Lasko, Thomas A and Sun, Jimeng and Malin, Bradley A},
    number = {3},
    month = {3},
    pages = {596--604},
    volume = {28},
    publisher = {Oxford University Press},
    doi = {10.1093/jamia/ocaa262},
    issn = {1527-974X}
}

@article{Hernandez2022,
    title = {{Synthetic data generation for tabular health records: A systematic review}},
    year = {2022},
    journal = {Neurocomputing},
    author = {Hernandez, Mikel and Epelde, Gorka and Alberdi, Ane and Cilla, Rodrigo and Rankin, Debbie},
    pages = {28--45},
    volume = {493},
    publisher = {Elsevier B.V.},
    doi = {10.1016/j.neucom.2022.04.053},
    issn = {09252312}
}

@article{Bhanot2021a,
    title = {{The Problem of Fairness in Synthetic Healthcare Data}},
    year = {2021},
    journal = {Entropy},
    author = {Bhanot, Karan and Qi, Miao and Erickson, John S. and Guyon, Isabelle and Bennett, Kristin P.},
    number = {9},
    month = {9},
    pages = {1165},
    volume = {23},
    publisher = {MDPI},
    doi = {10.3390/e23091165},
    issn = {1099-4300}
}

@article{Huque2018AStudies,
    title = {{A comparison of multiple imputation methods for missing data in longitudinal studies}},
    year = {2018},
    journal = {BMC Medical Research Methodology},
    author = {Huque, Md Hamidul and Carlin, John B. and Simpson, Julie A. and Lee, Katherine J.},
    number = {1},
    month = {12},
    pages = {168},
    volume = {18},
    doi = {10.1186/s12874-018-0615-6},
    issn = {1471-2288}
}

@article{Aristodimou2022ADomain,
    title = {{A fast supervised density-based discretization algorithm for classification tasks in the medical domain}},
    year = {2022},
    journal = {Health Informatics Journal},
    author = {Aristodimou, Aristos and Diavastos, Andreas and Pattichis, Constantinos S.},
    number = {1},
    volume = {28},
    doi = {10.1177/14604582211065397},
    issn = {17412811}
}

@article{Mosquera2023AData,
    title = {{A method for generating synthetic longitudinal health data}},
    year = {2023},
    journal = {BMC Medical Research Methodology},
    author = {Mosquera, Lucy and El Emam, Khaled and Ding, Lei and Sharma, Vishal and Zhang, Xue Hua and Kababji, Samer El and Carvalho, Chris and Hamilton, Brian and Palfrey, Dan and Kong, Linglong and Jiang, Bei and Eurich, Dean T.},
    number = {1},
    volume = {23},
    publisher = {BioMed Central Ltd},
    doi = {10.1186/s12874-023-01869-w},
    issn = {14712288},
    pmid = {36959532}
}

@article{Dankar2022AGenerators,
    title = {{A Multi-Dimensional Evaluation of Synthetic Data Generators}},
    year = {2022},
    journal = {IEEE Access},
    author = {Dankar, Fida K. and Ibrahim, Mahmoud K. and Ismail, Leila},
    pages = {11147--11158},
    volume = {10},
    publisher = {Institute of Electrical and Electronics Engineers Inc.},
    doi = {10.1109/ACCESS.2022.3144765},
    issn = {2169-3536}
}

@article{Yan2022AModels,
    title = {{A Multifaceted benchmarking of synthetic electronic health record generation models}},
    year = {2022},
    journal = {Nature Communications},
    author = {Yan, Chao and Yan, Yao and Wan, Zhiyu and Zhang, Ziqi and Omberg, Larsson and Guinney, Justin and Mooney, Sean D. and Malin, Bradley A.},
    number = {1},
    month = {12},
    pages = {7609},
    volume = {13},
    publisher = {Nature Research},
    doi = {10.1038/s41467-022-35295-1},
    issn = {2041-1723},
    pmid = {36494374},
    arxivId = {2208.01230}
}

@article{Ghosheh2024ARecords,
    title = {{A Survey of Generative Adversarial Networks for Synthesizing Structured Electronic Health Records}},
    year = {2024},
    journal = {ACM Computing Surveys},
    author = {Ghosheh, Ghadeer O. and Li, Jin and Zhu, Tingting},
    number = {6},
    month = {6},
    pages = {1--34},
    volume = {56},
    publisher = {Association for Computing Machinery},
    doi = {10.1145/3636424},
    issn = {0360-0300}
}

@article{Little1988AValues,
    title = {{A test of missing completely at random for multivariate data with missing values}},
    year = {1988},
    journal = {Journal of the American Statistical Association},
    author = {Little, Roderick J.A.},
    number = {404},
    volume = {83},
    doi = {10.1080/01621459.1988.10478722},
    issn = {1537274X}
}

@article{Pereira2024AssessmentData,
    title = {{Assessment of differentially private synthetic data for utility and fairness in end-to-end machine learning pipelines for tabular data}},
    year = {2024},
    journal = {PLOS ONE},
    author = {Pereira, Mayana and Kshirsagar, Meghana and Mukherjee, Sumit and Dodhia, Rahul and Lavista Ferres, Juan and de Sousa, Rafael},
    editor = {Cao, Tien-Dung},
    number = {2},
    pages = {e0297271},
    volume = {19},
    publisher = {Public Library of Science},
    doi = {10.1371/journal.pone.0297271},
    issn = {1932-6203},
    pmid = {38315667},
    arxivId = {2310.19250}
}

@article{Belgodere2024AuditingTrade-offs,
    title = {{Auditing and Generating Synthetic Data with Controllable Trust Trade-offs}},
    year = {2024},
    journal = {IEEE Journal on Emerging and Selected Topics in Circuits and Systems},
    author = {Belgodere, Brian and Dognin, Pierre and Ivankay, Adam and Melnyk, Igor and Mroueh, Youssef and Mojsilovic, Aleksandra and Navratil, Jiri and Nitsure, Apoorva and Padhi, Inkit and Rigotti, Mattia and Ross, Jerret and Schiff, Yair and Vedpathak, Radhika and Young, Richard A.},
    pages = {},
    doi = {10.1109/JETCAS.2024.3477976},
    issn = {2156-3357}
}

@article{Vallevik2024CanHealthcare,
    title = {{Can I trust my fake data – A comprehensive quality assessment framework for synthetic tabular data in healthcare}},
    year = {2024},
    journal = {International Journal of Medical Informatics},
    author = {Vallevik, Vibeke Binz and Babic, Aleksandar and Marshall, Serena E. and Elvatun, Severin and Br{\o}gger, Helga M.B. and Alagaratnam, Sharmini and Edwin, Bjørn and Veeraragavan, Narasimha R. and Befring, Anne Kjersti and Nyg{\aa}rd, Jan F.},
    month = {5},
    pages = {105413},
    volume = {185},
    publisher = {Elsevier Ireland Ltd},
    doi = {10.1016/j.ijmedinf.2024.105413},
    issn = {13865056},
    pmid = {38493547}
}

@article{Pang2024CEHR-GPT:Timelines,
    title = {{CEHR-GPT: Generating Electronic Health Records with Chronological Patient Timelines}},
    year = {2024},
    journal = {arXiv preprint},
    author = {Pang, Chao and Jiang, Xinzhuo and Pavinkurve, Nishanth Parameshwar and Kalluri, Krishna S. and Minto, Elise L. and Patterson, Jason and Zhang, Linying and Hripcsak, George and G{\"{u}}rsoy, Gamze and Elhadad, Noémie and Natarajan, Karthik},
    doi = {10.48550/arXiv.2402.04400},
    arxivId = {2402.04400}
}

@article{Sun2023CollaborativeInference,
    title = {{Collaborative Synthesis of Patient Records through Multi-Visit Health State Inference}},
    year = {2023},
    journal = {arXiv preprint},
    author = {Sun, Hongda and Lin, Hongzhan and Yan, Rui},
    doi = {10.48550/arXiv.2312.14646},
    arxivId = {2312.14646}
}

@article{Isasa2024ComparativeSynthesis,
    title = {{Comparative assessment of synthetic time series generation approaches in healthcare: leveraging patient metadata for accurate data synthesis}},
    year = {2024},
    journal = {BMC Medical Informatics and Decision Making},
    author = {Isasa, Imanol and Hernandez, Mikel and Epelde, Gorka and Londo{\~{n}}o, Francisco and Beristain, Andoni and Larrea, Xabat and Alberdi, Ane and Bamidis, Panagiotis and Konstantinidis, Evdokimos},
    number = {1},
    volume = {24},
    publisher = {BioMed Central Ltd},
    doi = {10.1186/s12911-024-02427-0},
    issn = {14726947},
    pmid = {38291386}
}

@article{Haleem2023Deep-Learning-DrivenSynthesis,
    title = {{Deep-Learning-Driven Techniques for Real-Time Multimodal Health and Physical Data Synthesis}},
    year = {2023},
    journal = {Electronics (Switzerland)},
    author = {Haleem, Muhammad Salman and Ekuban, Audrey and Antonini, Alessio and Pagliara, Silvio and Pecchia, Leandro and Allocca, Carlo},
    number = {9},
    volume = {12},
    publisher = {MDPI},
    doi = {10.3390/electronics12091989},
    issn = {20799292}
}

@techreport{TowardsEuropeanHealthDataSpaceTEHDAS2025DraftData,
    title = {{Draft guideline on data minimisation, pseudonymisation, anonymisation and synthetic data}},
    year = {2025},
    author = {{Towards European Health Data Space (TEHDAS)}},
    month = {9}
}

@article{Wang2023EnhancingWCGAN-GP,
    title = {{Enhancing Small Tabular Clinical Trial Dataset through Hybrid Data Augmentation: Combining SMOTE and WCGAN-GP}},
    year = {2023},
    journal = {Data},
    author = {Wang, Winston and Pai, Tun-Wen},
    number = {9},
    pages = {135},
    volume = {8},
    publisher = {Multidisciplinary Digital Publishing Institute (MDPI)},
    doi = {10.3390/data8090135},
    issn = {2306-5729}
}

@inproceedings{Biswal2021EVA:Autoencoders,
    title = {{EVA: Generating Longitudinal Electronic Health Records Using Conditional Variational Autoencoders}},
    year = {2021},
    booktitle = {Proceedings of the 6th Machine Learning for Healthcare Conference},
    author = {Biswal, Siddharth and Ghosh, Soumya and Duke, Jon and Malin, Bradley and Stewart, Walter and Sun, Jimeng},
    editor = {Jung, Ken and Yeung, Serena and Sendak, Mark and Sjoding, Michael and Ranganath, Rajesh},
    month = {8},
    pages = {260--282},
    publisher = {PMLR},
    arxivId = {2012.10020}
}

@incollection{Diggle2002ExploringData,
    title = {{Exploring longitudinal data}},
    year = {2002},
    booktitle = {Analysis of Longitudinal Data},
    author = {Diggle, Peter and Heagerty, Patrick and Liang, Kung-Yee and Zeger, Scott},
    chapter = {3},
    edition = {2nd},
    pages = {39--52},
    publisher = {Oxford University Press},
    isbn = {9780191664328},
    doi = {10.1093/oso/9780198524847.001.0001}
}

@article{Kuo2023GeneratingHIV,
    title = {{Generating synthetic clinical data that capture class imbalanced distributions with generative adversarial networks: Example using antiretroviral therapy for HIV}},
    year = {2023},
    journal = {Journal of Biomedical Informatics},
    author = {Kuo, Nicholas I-Hsien and Garcia, Federico and S{\"{o}}nnerborg, Anders and B{\"{o}}hm, Michael and Kaiser, Rolf and Zazzi, Maurizio and Polizzotto, Mark and Jorm, Louisa and Barbieri, Sebastiano},
    pages = {104436},
    volume = {144},
    publisher = {Academic Press Inc.},
    doi = {10.1016/j.jbi.2023.104436},
    issn = {15320464},
    pmid = {37451495},
    arxivId = {2208.08655}
}

@article{Li2023GeneratingApplications,
    title = {{Generating synthetic mixed-type longitudinal electronic health records for artificial intelligent applications}},
    year = {2023},
    journal = {npj Digital Medicine},
    author = {Li, Jin and Cairns, Benjamin J. and Li, Jingsong and Zhu, Tingting},
    number = {1},
    month = {5},
    pages = {98},
    volume = {6},
    doi = {10.1038/s41746-023-00834-7},
    issn = {2398-6352}
}

@article{Ibrahim2025GenerativeChallenges,
    title = {{Generative AI for synthetic data across multiple medical modalities: A systematic review of recent developments and challenges}},
    year = {2025},
    journal = {Computers in Biology and Medicine},
    author = {Ibrahim, Mahmoud and Khalil, Yasmina Al and Amirrajab, Sina and Sun, Chang and Breeuwer, Marcel and Pluim, Josien and Elen, Bart and Ertaylan, Gökhan and Dumontier, Michel},
    month = {5},
    pages = {109834},
    volume = {189},
    doi = {10.1016/j.compbiomed.2025.109834},
    issn = {00104825},
    arxivId = {2407.00116}
}

@inproceedings{Alaa2022HowModels,
    title = {{How Faithful is your Synthetic Data? Sample-level Metrics for Evaluating and Auditing Generative Models}},
    year = {2022},
    booktitle = {Proceedings of the 39th International Conference on Machine Learning},
    author = {Alaa, Ahmed M and Van Breugel, Boris and Saveliev, Evgeny and Van Der Schaar, Mihaela},
    editor = {Chaudhuri, Kamalika and Jegelka, Stefanie and Song, Le and Szepesvari, Csaba and Niu, Gang and Sabato, Sivan},
    pages = {290--306},
    publisher = {PMLR},
    address = {Baltimore, Maryland}
}

@inproceedings{Gottesman2020InterpretableTransitions,
    title = {{Interpretable Off-Policy Evaluation in Reinforcement Learning by Highlighting Influential Transitions}},
    year = {2020},
    booktitle = {Proceedings of the 37th International Conference on Machine Learning},
    author = {Gottesman, Omer and Futoma, Joseph and Liu, Yao and Parbhoo, Sonali and Celi, Leo Anthony and Brunskill, Emma and Doshi-Velez, Finale},
    editor = {III, Hal Daumé and Singh, Aarti},
    month = {7},
    pages = {3658--3667},
    publisher = {PMLR},
    doi = {https://dl.acm.org/doi/10.5555/3524938.3525281}
}

@book{Wand1995KernelSmoothing,
    title = {{Kernel Smoothing}},
    year = {1995},
    booktitle = {Kernel Smoothing},
    author = {Wand, M. P. and Jones, M. C.},
    doi = {10.1007/978-1-4899-4493-1}
}

@article{Li2013LittlesRandom,
    title = {{Little's test of missing completely at random}},
    year = {2013},
    journal = {Stata Journal},
    author = {Li, Cheng},
    number = {4},
    volume = {13},
    doi = {10.1177/1536867x1301300407},
    issn = {1536867X}
}

@article{Perkonoja2025MethodsReview,
    title = {{Methods for Generating and Evaluating Synthetic Longitudinal Patient Data: A Systematic Review}},
    year = {2025},
    journal = {Journal of Healthcare Informatics Research},
    author = {Perkonoja, Katariina and Auranen, Kari and Virta, Joni},
    doi = {10.1007/s41666-025-00223-7},
    issn = {2509-4971}
}

@article{Johnson2016MIMIC-III1.4,
    title = {{MIMIC-III clinical database (version 1.4)}},
    year = {2016},
    journal = {PhysioNet},
    author = {Johnson, Alistair and Pollard, Tom and Mark, Roger},
    number = {},
    volume = {},
    doi = {10.13026/C2XW26}
}

@article{Johnson2016MIMIC-IIIDatabase,
    title = {{MIMIC-III, a freely accessible critical care database}},
    year = {2016},
    journal = {Scientific Data},
    author = {Johnson, Alistair E.W. and Pollard, Tom J. and Shen, Lu and Lehman, Li Wei H. and Feng, Mengling and Ghassemi, Mohammad and Moody, Benjamin and Szolovits, Peter and Anthony Celi, Leo and Mark, Roger G.},
    volume = {3},
    publisher = {Nature Publishing Groups},
    doi = {10.1038/sdata.2016.35},
    issn = {20524463},
    pmid = {27219127}
}

@incollection{Diggle2002MissingData,
    title = {{Missing values in longitudinal data}},
    year = {2002},
    booktitle = {Analysis of Longitudinal Data},
    author = {Diggle, Peter and Heagerty, Patrick and Liang, Kung-Yee and Zeger, Scott L.},
    chapter = {13},
    edition = {2nd},
    pages = {282--318},
    publisher = {Oxford University Press},
    isbn = {978-0-19-852484-7},
    doi = {10.1093/oso/9780198524847.001.0001}
}

@inproceedings{Wu2018NeuralVocabularies,
    title = {{Neural response generation with dynamic vocabularies}},
    year = {2018},
    booktitle = {32nd AAAI Conference on Artificial Intelligence, AAAI 2018},
    author = {Wu, Yu and Wu, Wei and Yang, Dejian and Xu, Can and Li, Zhoujun},
    doi = {10.1609/aaai.v32i1.11943},
    issn = {2159-5399}
}

@article{Diggle1998NonparametricData,
    title = {{Nonparametric Estimation of Covariance Structure in Longitudinal Data}},
    year = {1998},
    journal = {Biometrics},
    author = {Diggle, Peter J. and Verbyla, Arunas P.},
    number = {2},
    month = {6},
    pages = {401},
    volume = {54},
    doi = {10.2307/3109751},
    issn = {0006341X}
}

@article{Tomizawa1971OnProblems,
    title = {{On some techniques useful for solution of transportation network problems}},
    year = {1971},
    journal = {Networks},
    author = {Tomizawa, N.},
    number = {2},
    volume = {1},
    doi = {10.1002/net.3230010206},
    issn = {10970037}
}

@article{Achterberg2024OnRecords,
    title = {{On the evaluation of synthetic longitudinal electronic health records}},
    year = {2024},
    journal = {BMC Medical Research Methodology},
    author = {Achterberg, Jim L. and Haas, Marcel R. and Spruit, Marco R.},
    number = {1},
    volume = {24},
    publisher = {BioMed Central Ltd},
    doi = {10.1186/s12874-024-02304-4},
    issn = {14712288},
    pmid = {39143466}
}

@article{Osorio-Marulanda2024PrivacyReview,
    title = {{Privacy Mechanisms and Evaluation Metrics for Synthetic Data Generation: A Systematic Review}},
    year = {2024},
    journal = {IEEE Access},
    author = {Osorio-Marulanda, Pablo A. and Epelde, Gorka and Hernandez, Mikel and Isasa, Imanol and Reyes, Nicolas Moreno and Iraola, Andoni Beristain},
    pages = {88048--88074},
    volume = {12},
    publisher = {Institute of Electrical and Electronics Engineers Inc.},
    doi = {10.1109/ACCESS.2024.3417608},
    issn = {2169-3536}
}

@inproceedings{Wang2022PromptEHR:Learning,
    title = {{PromptEHR: Conditional Electronic Healthcare Records Generation with Prompt Learning}},
    year = {2022},
    booktitle = {Proceedings of the 2022 Conference on Empirical Methods in Natural Language Processing},
    author = {Wang, Zifeng and Sun, Jimeng},
    pages = {2873--2885},
    publisher = {Association for Computational Linguistics},
    address = {Stroudsburg, PA, USA},
    doi = {10.18653/v1/2022.emnlp-main.185}
}

@article{Theodorou2023SynthesizeModel,
    title = {{Synthesize high-dimensional longitudinal electronic health records via hierarchical autoregressive language model}},
    year = {2023},
    journal = {Nature Communications},
    author = {Theodorou, Brandon and Xiao, Cao and Sun, Jimeng},
    number = {1},
    pages = {5305},
    volume = {14},
    publisher = {Nature Research},
    doi = {10.1038/s41467-023-41093-0},
    issn = {2041-1723}
}

@article{Murtaza2023SyntheticDomain,
    title = {{Synthetic data generation: State of the art in health care domain}},
    year = {2023},
    journal = {Computer Science Review},
    author = {Murtaza, Hajra and Ahmed, Musharif and Khan, Naurin Farooq and Murtaza, Ghulam and Zafar, Saad and Bano, Ambreen},
    month = {5},
    pages = {100546},
    volume = {48},
    publisher = {Elsevier Ireland Ltd},
    doi = {10.1016/j.cosrev.2023.100546},
    issn = {15740137}
}

@article{Nikolentzos2023SyntheticAutoencoders,
    title = {{Synthetic electronic health records generated with variational graph autoencoders}},
    year = {2023},
    journal = {npj Digital Medicine},
    author = {Nikolentzos, Giannis and Vazirgiannis, Michalis and Xypolopoulos, Christos and Lingman, Markus and Brandt, Erik G.},
    number = {1},
    volume = {6},
    publisher = {Nature Research},
    doi = {10.1038/s41746-023-00822-x},
    issn = {23986352}
}

@article{Hernadez2023SyntheticDimensions,
    title = {{Synthetic Tabular Data Evaluation in the Health Domain Covering Resemblance, Utility, and Privacy Dimensions}},
    year = {2023},
    journal = {Methods of Information in Medicine},
    author = {Hernadez, Mikel and Epelde, Gorka and Alberdi, Ane and Cilla, Rodrigo and Rankin, Debbie},
    number = {S 01},
    pages = {e19-e38},
    volume = {62},
    publisher = {Georg Thieme Verlag},
    doi = {10.1055/s-0042-1760247},
    issn = {0026-1270}
}

@article{Lautrup2024SystematicData,
    title = {{Systematic Review of Generative Modelling Tools and Utility Metrics for Fully Synthetic Tabular Data}},
    year = {2024},
    journal = {ACM Computing Surveys},
    author = {Lautrup, Anton Danholt and Hyrup, Tobias and Zimek, Arthur and Schneider-Kamp, Peter},
    doi = {10.1145/3704437},
    issn = {0360-0300}
}

@article{Gringarten2001TeachersModeling,
    title = {{Teacher's Aide Variogram Interpretation and Modeling}},
    year = {2001},
    journal = {Mathematical Geology},
    author = {Gringarten, Emmanuel and Deutsch, Clayton V.},
    number = {4},
    month = {5},
    pages = {507--534},
    volume = {33},
    doi = {10.1023/A:1011093014141},
    issn = {0882-8121}
}

@article{Kuo2022TheAlgorithms,
    title = {{The Health Gym: synthetic health-related datasets for the development of reinforcement learning algorithms}},
    year = {2022},
    journal = {Scientific Data},
    author = {Kuo, Nicholas I-Hsien and Polizzotto, Mark N. and Finfer, Simon and Garcia, Federico and S{\"{o}}nnerborg, Anders and Zazzi, Maurizio and B{\"{o}}hm, Michael and Kaiser, Rolf and Jorm, Louisa and Barbieri, Sebastiano},
    number = {1},
    month = {11},
    pages = {693},
    volume = {9},
    publisher = {Nature Research},
    doi = {10.1038/s41597-022-01784-7},
    issn = {2052-4463}
}

@article{Kuhn1955TheProblem,
    title = {{The Hungarian method for the assignment problem}},
    year = {1955},
    journal = {Naval Research Logistics Quarterly},
    author = {Kuhn, H. W.},
    number = {1-2},
    volume = {2},
    doi = {10.1002/nav.3800020109},
    issn = {0028-1441}
}

@article{Edmonds1972TheoreticalProblems,
    title = {{Theoretical Improvements in Algorithmic Efficiency for Network Flow Problems}},
    year = {1972},
    journal = {Journal of the ACM (JACM)},
    author = {Edmonds, Jack and Karp, Richard M.},
    number = {2},
    volume = {19},
    doi = {10.1145/321694.321699},
    issn = {1557735X}
}

@inproceedings{Hashemi2023Time-seriesNetwork,
    title = {{Time-series Anonymization of Tabular Health Data using Generative Adversarial Network}},
    year = {2023},
    booktitle = {2023 International Joint Conference on Neural Networks (IJCNN)},
    author = {Hashemi, Atiye Sadat and Etminani, Kobra and Soliman, Amira and Hamed, Omar and Lundstr{\"{o}}m, Jens},
    pages = {1--8},
    volume = {2023-June},
    publisher = {IEEE},
    isbn = {978-1-6654-8867-9},
    doi = {10.1109/IJCNN54540.2023.10191367}
}

\end{document}